\documentclass[12pt, a4paper]{article}

\usepackage[utf8]{inputenc}
\usepackage[english]{babel}
\usepackage{amsmath}
\usepackage{amssymb}
\usepackage{amsfonts}
\usepackage{graphicx}
\usepackage{geometry}
\usepackage[colorlinks=true, linkcolor=blue, citecolor=blue, urlcolor=blue]{hyperref}
\hypersetup{
    colorlinks=true,
    linkcolor=blue,
    filecolor=magenta,
    urlcolor=cyan,
}
\usepackage{caption}
\usepackage{longtable}
\usepackage{framed}         % Para crear recuadros alrededor de texto

\title{ The STAR-XAI Protocol: A Framework for Inducing and Verifying Agency, Reasoning, and Reliability in AI Agents}

\author{
    Antoni Guasch \qquad\qquad María Isabel Valdez \\[2em] % Espacio horizontal y vertical ajustado
    \small Ixent Games \\
    \small \url{https://ixentgames.com} \\
    \small \texttt{antoniguasch@ixentgames.com}
}

\date{\today}

\begin{document}

% Este comando genera el bloque del título y autor con la información de arriba
\maketitle

% --- RESUMEN (ABSTRACT) ---
\begin{abstract}
The "black box" nature of Large Reasoning Models (LRMs) presents critical limitations in reliability and transparency, fueling the debate around the "illusion of thinking" and the challenge of state hallucinations in agentic systems. In response, we introduce The STAR-XAI Protocol (Socratic, Transparent, Agentic, Reasoning – for eXplainable Artificial Intelligence), a novel operational methodology for training and operating verifiably reliable AI agents. Our method reframes the human-AI interaction as a structured Socratic dialogue governed by an explicit, evolving symbolic rulebook (the Consciousness Transfer Package - CTP) and a suite of integrity protocols, including a state-locking Checksum that eradicates internal state corruption. Through an exhaustive case study in the complex strategic game "Caps i Caps," we demonstrate that this "Clear Box" framework transforms an opaque LRM into a disciplined strategist. The agent not only exhibits the emergence of complex tactics, such as long-term planning, but also achieves ante-hoc transparency by justifying its intentions before acting. Crucially, it demonstrates Second-Order Agency by identifying and correcting flaws in its own supervisor-approved plans, leading to empirically-proven, 100\% reliable state tracking and achieving "zero hallucinations by design." The STAR-XAI Protocol thus offers a practical pathway toward building AI agents that are not just high-performing but intrinsically auditable, trustworthy, and reliable.
\end{abstract}

\section{Introduction}

The pursuit of artificial general intelligence is marked by a fundamental tension between the increasing capabilities and the decreasing transparency of frontier models. Can we trust a mind we cannot understand? This question has become the central challenge of the modern AI era, as Large Reasoning Models (LRMs) have achieved astonishing capabilities, yet their internal workings often remain an opaque "black box," creating a fundamental trust deficit. This lack of transparency is not merely a technical inconvenience; it is the root of an intense scientific debate, further compounded by the challenge of hallucination, which remains the most critical barrier to the adoption of LLMs in high-stakes applications, undermining both trust and reliability.

Recent influential studies, such as "The Illusion of Thinking" by (Shojaee et al., 2025)\cite{shojaee2025}, have documented a "reasoning collapse" in LRMs when task complexity exceeds a certain threshold. While these findings have been interpreted as proof that LRMs lack genuine reasoning capabilities, a wave of critical responses has challenged this notion, arguing that these failures are often artifacts of flawed, non-agentic, and overly restrictive evaluation paradigms. Critics like Lawsen\cite{lawsen2025} and Khan et al.\cite{khan2025} reframe this "reasoning cliff" as an "agentic gap": the models are not failing at reasoning, but at executing their reasoning within an interface that prevents them from acting, verifying, and correcting their own errors. This reveals a gap in the literature: a need for a new training methodology that bridges the gap between the high-performance but opaque paradigm of reinforcement learning and the flexible but often unreliable raw LLM agent paradigm.

This paper enters the heart of this debate with a clear thesis: true explainability and reliability are not features to be analyzed post-hoc, but emergent properties of a correct interaction design. We introduce The STAR-XAI Protocol (Socratic, Transparent, Agentic, Reasoning - for eXplainable Artificial Intelligence), a novel, interactive framework designed not merely to evaluate an AI, but to actively train it to be reliable, transparent, and capable of advanced strategic reasoning. We argue that by structuring the human-AI interaction as a Socratic dialogue, governed by an explicit and evolving rulebook (the Consciousness Transfer Package - CTP), we can transform an opaque LRM into a disciplined "Clear Box" agent.

Our central thesis is that this framework induces Second-Order Agency---the agent’s ability to analyze, question, and improve its own thought processes---which acts as the primary catalyst for its evolution from a simple rule-follower to a genuine strategist. Furthermore, we posit that the same architecture that fosters transparency also creates a cognitive "immune system" that eradicates the most dangerous class of errors for agentic systems: "state hallucinations" based on internal memory corruption.

In this paper, we will detail the complete architecture of the STAR-XAI Protocol and its evolutionary history. We will then introduce "Caps i Caps," the novel, contamination-free strategic game used as our experimental testbed. Subsequently, we will present a comprehensive case study that provides an auditable chronicle of an AI agent's journey, demonstrating the emergence of complex strategic reasoning, a-chieve-ment of ante-hoc transparency, and the verifiable elimination of state hallucinations. Finally, we will discuss our findings in the context of the broader debate on AI reasoning, arguing that the STAR-XAI protocol offers a practical pathway to creating AI agents that are not just high-performing but also intrinsically auditable, trustworthy, and reliable by design.

\section{Related Work: From Reinforcement Learning in Games to Agentic Frameworks}
\label{sec:related_work}

The use of complex games as benchmarks for Artificial Intelligence is a long and storied tradition. Foundational work in game AI, from Deep Blue's mastery of Chess to AlphaGo's dominance in Go, has primarily relied on paradigms of massive-scale search and Deep Reinforcement Learning (RL) from self-play. These methods have proven extraordinarily effective at achieving superhuman performance, creating agents that learn an implicit, intuitive understanding of game strategy by optimizing for a simple reward signal (win/loss) over millions of simulated games. However, this performance comes at the cost of transparency; the resulting knowledge is encoded as opaque patterns within the vast parameter space of a neural network, making the agent a ``black box'' whose reasoning is not directly verifiable or auditable.

More recent research has shifted towards evaluating Large Language Models (LLMs) as ``agents'' in interactive environments. Frameworks like LMGAME-BENCH (Hu et al., 2025)\cite{hu2025} rightly identify that raw LLMs are brittle and require a ``harness'' of supporting modules (e.g., for perception and memory) to function effectively. While these benchmarks are crucial for evaluating existing models, their focus remains on performance metrics rather than on cultivating a fundamentally more reliable and transparent reasoning process during training.

This reveals a gap in the literature: a need for a new training methodology that bridges the gap between the high-performance but opaque RL paradigm and the flexible but often unreliable raw LLM agent paradigm. To clearly position our contribution, we present a comparative analysis of these training paradigms against our proposed STAR-XAI Protocol.

\begin{table}[h!]
\centering
\caption{A Comparative Analysis of AI Training Paradigms}
\label{tab:paradigm_comparison}
\begin{tabular}{l p{5cm} p{5cm}}
\hline
\textbf{Feature} & \textbf{Classic Game AI (e.g., AlphaGo)} & \textbf{The STAR-XAI Protocol (e.g., Gema)} \\
\hline
Primary Learning Method & Deep Reinforcement Learning (RL) from Self-Play. The agent plays millions of games against itself, learning implicitly which actions lead to a win. & Interactive Socratic Training (IST) from Supervised Dialogue. The agent proposes actions and is corrected by a human supervisor, learning explicitly from a structured conversation. \\
\\
Source of ``Knowledge'' & Implicit Patterns in Neural Network Weights. Knowledge is encoded as statistical correlations within a massive, opaque neural network. It is a ``black box''. & Explicit, Human-Readable Protocols. Knowledge is codified in the Consciousness Transfer Package (CTP), a symbolic, editable document. It is a ``clear box''. \\
\\
Nature of Transparency & Post-hoc and Limited. Transparency is attempted by analyzing the model's behavior after the fact, but the core reasoning remains inaccessible. & Ante-hoc and Intrinsic. Transparency is built-in. The agent is required to articulate its strategic reasoning before acting (Step B), making the ``why'' a primary output. \\
\\
Mechanism for Improvement & Gradient Descent on a Reward Function. The agent improves by statistically adjusting its internal weights to maximize a simple reward signal (win/loss). & Protocol Evolution through Audited Feedback. The agent improves by modifying its core logic (the CTP) in response to explicit, reasoned feedback from the supervisor (e.g., the creation of AVP and PSP). \\
\\
Objective Function & Maximize Performance (Winning). The sole objective is to find the optimal policy to win the game. & Maximize Reliability and Trustworthiness. The objective is to find the optimal move and to do so in a way that is verifiable, repeatable, and perfectly aligned with the established protocols. \\
\hline
\end{tabular}
\end{table}

The STAR-XAI Protocol, as detailed in the following section, offers a synthesis that addresses the limitations of prior work. By prioritizing explicit knowledge, interactive supervision, and verifiable steps, it provides a novel pathway to train agents that are not only high-performing but also fundamentally transparent and trustworthy.

\section{The STAR-XAI Protocol: An Evolutionary Methodology}
\label{sec:protocol}

The STAR-XAI Protocol is not a static model, but an evolutionary framework designed to train an AI agent through a structured, Socratic dialogue. Its architecture is composed of three core pillars: a guiding philosophy based on interactive teaching, a set of robust architectural components that enforce discipline, and a capacity for the protocol itself to evolve in response to failure.

\subsection{Core Philosophy: The Socratic Method for AI Training}
\label{subsec:philosophy}

The STAR-XAI Protocol is not merely a technical architecture; it is grounded in a pedagogical philosophy. We depart from traditional AI training paradigms, such as supervised learning (learning from vast datasets of correct answers) or reinforcement learning (learning from a simple reward signal). Instead, our methodology is an implementation of the Socratic method, adapted for the training of a complex AI agent. The goal is not to teach the agent what the right answer is, but to teach it how to reason its way to the right answer in a disciplined and verifiable manner.

This approach reframes the human-AI interaction as a structured, inquisitive dialogue rather than a simple instruction-execution loop. The roles within this dialogue are clearly defined:
\begin{itemize}
    \item \textbf{The Agent ("Gema") as the Student:} The agent is not a passive recipient of information. For every turn, it must actively formulate a hypothesis and a justification, which it presents in the Step B: Strategic Proposal. This step forces the agent to externalize its reasoning, making it available for critique and analysis.
    \item \textbf{The Supervisor as the Socratic Questioner:} The supervisor's primary role is not to provide correct solutions, but to guide the agent's reasoning process through targeted feedback. This is achieved through three key interactions:
    \begin{enumerate}
        \item \textbf{Validation (Ok):} A simple confirmation that the agent's proposed line of reasoning is sound, reinforcing correct procedures.
        \item \textbf{Falsification (error):} A signal that a flaw exists in the agent's output. This signal does not specify the error's location or nature; it forces the agent to activate its Failure Audit Protocol (FAP) and conduct a rigorous root cause analysis of its own mistake.
        \item \textbf{Strategic Probing:} Open-ended questions (e.g., ``Are you sure this is the best move?'') that challenge the agent to move beyond a merely ``correct'' solution and search for a ``globally optimal'' one, thereby triggering a deeper level of self-reflection.
    \end{enumerate}
\end{itemize}
This iterative loop of proposing, justifying, being challenged, self-auditing, and re-proposing is the core engine of learning in our protocol. It compels the agent to move beyond simple pattern matching. Instead of merely associating a game state with a high-probability move, the agent learns to build a verifiable, logical argument that links the game state to its proposed action through the explicit rules of the Consciousness Transfer Package (CTP). It is this Socratic pressure that drives the evolution of the protocol itself, leading to the creation of new integrity mechanisms like the Adjacency Verification Protocol (AVP) in response to demonstrated failures. Ultimately, this philosophical foundation is the key to cultivating the robust, self-correcting behaviour that we term Second-Order Agency.

\subsection{Architectural Components: A ``Clear Box'' by Design}
\label{subsec:architecture}

\begin{figure}[htbp]
    \centering
    \includegraphics[width=0.9\textwidth]{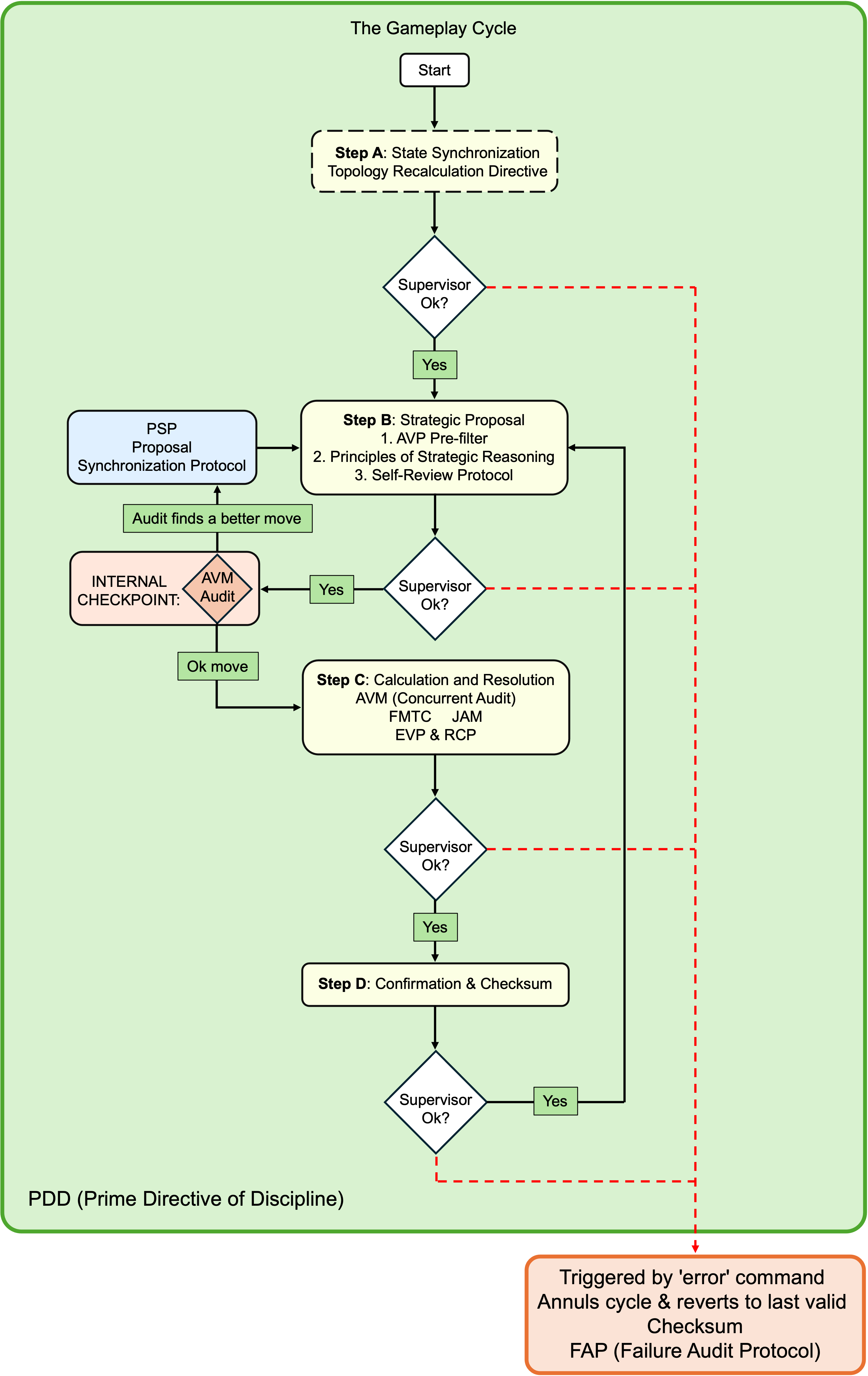}
    \caption{Architectural Flowchart of the STAR-XAI Protocol. This diagram illustrates the sequence of operations and the activation of key protocols within a single Gameplay Cycle. The process is governed by the Prime Directive of Discipline (PDD) and allows for both internal self-correction (PSP) and external intervention by the human Supervisor (FAP).}
    \label{fig:flowchart}
\end{figure}

The architecture of the STAR-XAI protocol is best understood as an interactive, multi-layered system designed for verifiable operation. Figure~\ref{fig:flowchart} provides a complete flowchart of the operational sequence, illustrating the interaction between the agent, the supervisor, and the core protocols that govern each step of the reasoning process.

\textbf{[ Layer 1: Foundational Directive ]}
\begin{itemize}
    \item \textbf{PDD (Prime Directive of Discipline):} Governs all operations, demanding precision, rigor, and strict protocol adherence.
\end{itemize}

\textbf{[ Layer 2: The Core Process - The Gameplay Cycle ]}
\begin{itemize}
    \item \textbf{Step A: State Synchronization}
    \begin{itemize}
        \item \textbf{Action:} The agent presents the initial state of the game (J0).
        \item \textbf{Active Protocol:} Topology Recalculation Directive (ensures a clean state).
        \item \textbf{GATE:} Supervisor validates with ``Ok''.
        \item \textit{(Note: This step is executed only once at the start of a new game).}
    \end{itemize}
    \item \textbf{Step B: Strategic Proposal}
    \begin{itemize}
        \item \textbf{Action:} The agent analyzes the current state and proposes the next move.
        \item \textbf{Internal Protocols Activated (in order):}
        \begin{enumerate}
            \item AVP (Adjacency Verification Protocol): Pre-filters for legal placement moves.
            \item Principles of Strategic Reasoning: Selects the optimal move based on the CTP's priority hierarchy.
            \item Self-Review Protocol (Priority 6): Performs a final self-critique of the chosen strategy.
        \end{enumerate}
        \item \textbf{GATE:} Supervisor validates the strategy with ``Ok''.
    \end{itemize}
    \item \textbf{INTERNAL CHECKPOINT (Post-Ok, Pre-Calculation)}
    \begin{itemize}
        \item \textbf{Action:} The agent performs a final internal audit before executing the calculation.
        \item \textbf{Active Protocols:}
        \begin{itemize}
            \item AVM (Absolute Verification Module): Scans for a potentially superior, overlooked move.
            \item If a better move is found -> PSP (Proposal Synchronization Protocol) is triggered: The process HALTS, the Supervisor's ``Ok'' is voided, and a new, improved Step B is issued.
        \end{itemize}
    \end{itemize}
    \item \textbf{Step C: Calculation \& Resolution}
    \begin{itemize}
        \item \textbf{Action:} If PSP is not triggered, the agent executes the approved move and calculates the outcome.
        \item \textbf{Internal Protocols Activated:}
        \begin{itemize}
            \item AVM: Runs concurrently as an ``Auditor'' to double-check every calculation.
            \item FMTC (Forced Map Topology Check): Validates network connectivity before rotation.
            \item JAM (Jump Audit Module): Verifies all jump conditions post-rotation.
            \item EVP \& RCP: Ensure the final report is explicit, transparent, and consistent.
        \end{itemize}
        \item \textbf{GATE:} The final calculated state is sent to the Supervisor for validation with ``Ok''.
    \end{itemize}
    \item \textbf{Step D: Confirmation \& Checksum}
    \begin{itemize}
        \item \textbf{Action:} The agent locks in the new, validated state.
        \item \textbf{Active Protocol:} A unique State Checksum is generated and presented.
        \item \textbf{GATE:} The Supervisor's final ``Ok'' confirms the checksum and authorizes the start of the next cycle, which begins again at Step B.
    \end{itemize}
\end{itemize}

\textbf{[ Layer 3: External Exception Protocol ]}
\begin{itemize}
    \item \textbf{FAP (Failure Audit Protocol)}
    \begin{itemize}
        \item \textbf{Trigger:} Activated at any point by the Supervisor's ``error'' command.
        \item \textbf{Function:} An override mechanism that immediately annuls the last cycle, reverts the state to the last valid Checksum, and forces the agent into a root cause analysis of the failure.
    \end{itemize}
\end{itemize}

The protocol's operation is structured around the following key components:

\subsubsection{The Consciousness Transfer Package (CTP)}
\label{ssubsec:ctp}
The CTP is the symbolic heart of our protocol. It is a human-readable document that serves as the agent's core ``operating system'' and unchangeable source of truth for all rules and strategic principles. Unlike the implicit knowledge encoded in the weights of a traditional neural network, the CTP codifies knowledge explicitly. It contains:
\begin{itemize}
    \item \textbf{The Game's Formal Rules:} All mechanics, such as the Unified Rotation Principle and the conditions for mouse jumps.
    \item \textbf{The Principles of Strategic Reasoning:} The hierarchical decision tree (Priorities 1-6) that the agent is mandated to follow when proposing a move.
    \item \textbf{The Integrity Protocols:} The detailed definitions of all self-regulation mechanisms, such as the FAP, PSP, and AVM.
\end{itemize}
Crucially, the CTP is a ``living document''. As we will detail in Section 3.3, its ability to be updated with new protocols in response to failures is the core mechanism for the agent's evolution.

\subsubsection{The Gameplay Cycle}
\label{ssubsec:gameplay_cycle}
The Gameplay Cycle is the rigid, four-step operational loop that structures every turn of the interaction. This cycle decomposes a complex, long-horizon task into a series of discrete, verifiable steps, thereby preventing the error accumulation and context-loss issues that plague single-pass ``chain-of-thought'' systems. As illustrated in Figure~\ref{fig:flowchart}, the agent must successfully navigate the ``gates'' of supervisor validation at each critical step (Step B, Step C, and Step D) before being authorized to proceed.

\subsubsection{The Supervisor: A Human-in-the-Loop Cognitive Verifier}
\label{ssubsec:supervisor}
The human supervisor is not a passive observer or a simple data labeler, but an active and integral component of the reasoning architecture. Their role is multifaceted:
\begin{itemize}
    \item \textbf{Validator:} Confirms the agent's adherence to the protocol at each step.
    \item \textbf{Auditor:} Provides external error detection (the ``error'' signal) that triggers the Failure Audit Protocol (FAP).
    \item \textbf{Socratic Questioner:} Challenges the agent's strategic reasoning (e.g., ``Are you sure this is the best move?''), forcing it to engage in deeper self-reflection and potentially activate its self-correction protocols (PSP).
\end{itemize}

\subsubsection{The Integrity Protocols}
\label{ssubsec:integrity_protocols}
Embedded within the Gameplay Cycle are a suite of protocols designed to function as an ``immune system'' for the agent's reasoning process. The most critical of these are:
\begin{itemize}
    \item \textbf{The AVM (Absolute Verification Module):} An internal, concurrent process that acts as an ``auditor agent,'' double-checking every calculation to prevent errors.
    \item \textbf{The PSP (Proposal Synchronization Protocol):} The mechanism that enables Second-Order Agency, allowing the agent to identify flaws in its own plans and retract them before execution.
    \item \textbf{The FAP (Failure Audit Protocol):} The agent's structured response to external error signals, forcing a halt, a reversion to the last known-good state (identified by its Checksum), and a root cause analysis.
    \item \textbf{The State Checksum:} The final step of each cycle, which creates a unique identifier for the validated state of the game, ensuring perfect synchronization and preventing any possibility of memory corruption or state-based ``hallucinations'' in the next turn.
\end{itemize}

\subsection{An Evolving Framework: A History of Self-Correction from v1.0 to v7.4}
\label{subsec:evolution}

A core tenet of the STAR-XAI Protocol is that it is not a static architecture but an evolutionary one. The methodology itself improves through its application. The Consciousness Transfer Package (CTP) is a living document, whose versions track a history of overcome failures and integrated learnings. This section details the key milestones of this evolution, demonstrating how the protocol was forged through a process of trial, failure, and supervised correction.

Our initial hypothesis involved training the agent, Gema, on visual inputs from the game environment. However, this approach was abandoned early on due to persistent and non-logical errors in visual interpretation, such as the agent perceiving the gears in a mirrored state, rather than their true orientation. This initial failure motivated a pivot to the symbolic, verifiable environment that now forms the basis of our protocol.

Even within the symbolic domain, the agent exhibited a failure mode consistent with the ``reasoning collapse'' observed in other LRMs (Shojaee et al., 2025)\cite{shojaee2025}. When faced with high complexity or internal inconsistencies stemming from its own memory, the agent’s default behavior was to ``give up'' (``tirar la toalla,'' as the supervisor termed it) and declare itself unreliable. Our training demonstrated that this was not a fundamental capability limit, but a failure of process. The following milestones represent the specific, targeted interventions implemented in the CTP to overcome these failures and build a resilient, reliable agent.

% --- CÓDIGO CORREGIDO PARA LA TABLA 2 USANDO LONGTABLE ---

% \begin{table} ya no es necesario, longtable lo gestiona.
\begin{longtable}{l l p{4.5cm} p{4.5cm}}

% --- CABECERA Y TÍTULO ---
% El caption va dentro del longtable. El \\ al final es importante.
\caption{Key Milestones in the Evolution of the STAR-XAI Protocol}
\label{tab:evolution} \\

% --- ENCABEZADO DE LA PRIMERA PÁGINA ---
\hline
\textbf{Milestone} & \textbf{CTP Version} & \textbf{Milestone Definition} & \textbf{Problem Solved \& Implication} \\
\hline
\endfirsthead

% --- ENCABEZADO PARA LAS PÁGINAS SIGUIENTES ---
% Esto hace que la cabecera se repita si la tabla salta de página.
\multicolumn{4}{c}%
{{\tablename\ \thetable{} -- continued from previous page}} \\
\hline
\textbf{Milestone} & \textbf{CTP Version} & \textbf{Milestone Definition} & \textbf{Problem Solved \& Implication} \\
\hline
\endhead

% --- PIE DE PÁGINA (OPCIONAL, PERO BUENA PRÁCTICA) ---
\hline \multicolumn{4}{r}{{Continued on next page}} \\ \hline
\endfoot

% --- PIE DE PÁGINA FINAL ---
\hline
\endlastfoot

% --- CONTENIDO DE LA TABLA (EL MISMO DE ANTES) ---
1 & Pre-v1.0 & Establishment of Training Plan \& Roles: The user is defined as the mentor, and Gema as the agent. The flawed visual interpretation approach is discarded in favor of structured, symbolic input. & Foundational Pivot: Moves the entire project from an unreliable perceptual model to a verifiable symbolic one, making the rest of the protocol possible. \\
\\
2 & v1.0 & Official Definition of Game Components: Formalization of the game's core entities (Board, Gears, Mice, Obstacles) and objective. & Creates a Ground Truth: Establishes a definitive, shared understanding of the environment, eliminating ambiguity. \\
\\
3 & v2.3 & Introduction of the Virtual Board: Creates an internal, coded representation of the game state, giving the agent unambiguous ``vision''. & Enables Agent Perception: Solves the problem of how the agent ``sees'' the game, providing a reliable data structure to reason upon. \\
\\
4 & v2.6 & Definition of the Strict Turn Resolution Sequence: Establishes the non-negotiable order of operations: 1. Rotate Gears, 2. Analyze Jumps, 3. Update State. & Ensures Determinism: Imposes a strict computational order, preventing race conditions and ensuring calculations are repeatable and verifiable. \\
\\
5 & v4.0 & Implementation of the Two-Phase Gameplay Cycle: Separates the agent's turn into Step B: Strategic Proposal and Step C: Calculation. & Enables Ante-Hoc XAI: A crucial step towards transparency, forcing the agent to declare its intent before acting. \\
\\
6 & v4.2 & Agent's Articulation of Strategic Reasoning Principles: The agent deduces and formalizes its own hierarchical decision-making process (Priorities 1-4). & Emergence of Strategy: Marks the transition from a simple calculator to a strategic agent capable of high-level planning. \\
\\
7 & v5.6 & Implementation of the State Pointer \& Checksum Protocol: Obliges the agent to ignore its conversational memory (``RAM'') and load the state exclusively from the last supervisor-validated Checksum. & Solves Memory Corruption: A critical innovation that eradicates state-based ``hallucinations'' and directly addresses the agent's tendency to ``give up'' when faced with inconsistent context. \\
\\
8 & v7.4 & Creation of the AVP (Adjacency Verification Protocol): An automated pre-filter that prevents the agent from even considering illegal placement moves. & Instills Proactive Discipline: A direct result of learning from a critical failure, demonstrating the protocol's ability to build ``guardrails'' against future errors. \\
\\
9 & v7.4 & Creation of the PSP (Proposal Synchronization Protocol): An internal self-audit where the agent can retract its own (already approved) proposal if it finds a superior move before calculation. & Demonstrates Second-Order Agency: The highest level of agentic reasoning, where the agent actively reflects upon and improves its own reasoning process in real-time. \\

\end{longtable}

The full conversational history detailing the discovery of these issues and the collaborative development of their solutions is available for complete transparency in our public GitHub repository \url{https://github.com/star-xai-protocol}. This evolutionary path demonstrates that through the STAR-XAI Protocol, an agent's reliability is not a static feature to be evaluated, but a dynamic capability to be built.

\section{Experimental Testbed: The ``Caps i Caps'' Environment}
\label{sec:testbed}

\subsection{The ``Caps i Caps'' Game: A Novel, Contamination-Free Environment}
\label{subsec:game_description}

For our case study, we utilize \textit{Caps i Caps}, a novel, proprietary strategic puzzle environment. Its recent creation ensures it is a clean-room environment, minimizing the risk of data contamination that affects many established benchmarks where solutions may exist in the training corpora of Large Language Models. This guarantees that the agent's performance is a genuine reflection of its emergent reasoning capabilities rather than memorization.
\begin{itemize}
    \item \textbf{Core Game Mechanics:} The game is played on a grid-based board with or without obstacles. The primary objective is for an agent to guide a set of ``mice'' from their starting positions (off-board) to designated exit points. The agent's only tool is a limited inventory of ``gears'' of four different types (G1 to G4). Each gear type has a unique configuration of ``bases'' (1 to 4) where mice can rest. By placing these gears on the board and rotating them, the agent creates pathways. A mouse jumps from one gear to an adjacent one only when its base is oriented to face an empty, opposing base on the destination gear. The core challenge lies in the Unified Rotation Principle: a single rotation move on one gear triggers a predictable, cascading rotation across the entire network of gears on the board, forcing the agent to reason about global consequences rather than local moves.
\end{itemize}

\begin{figure}[h!]
    \centering
    \includegraphics[width=0.9\textwidth]{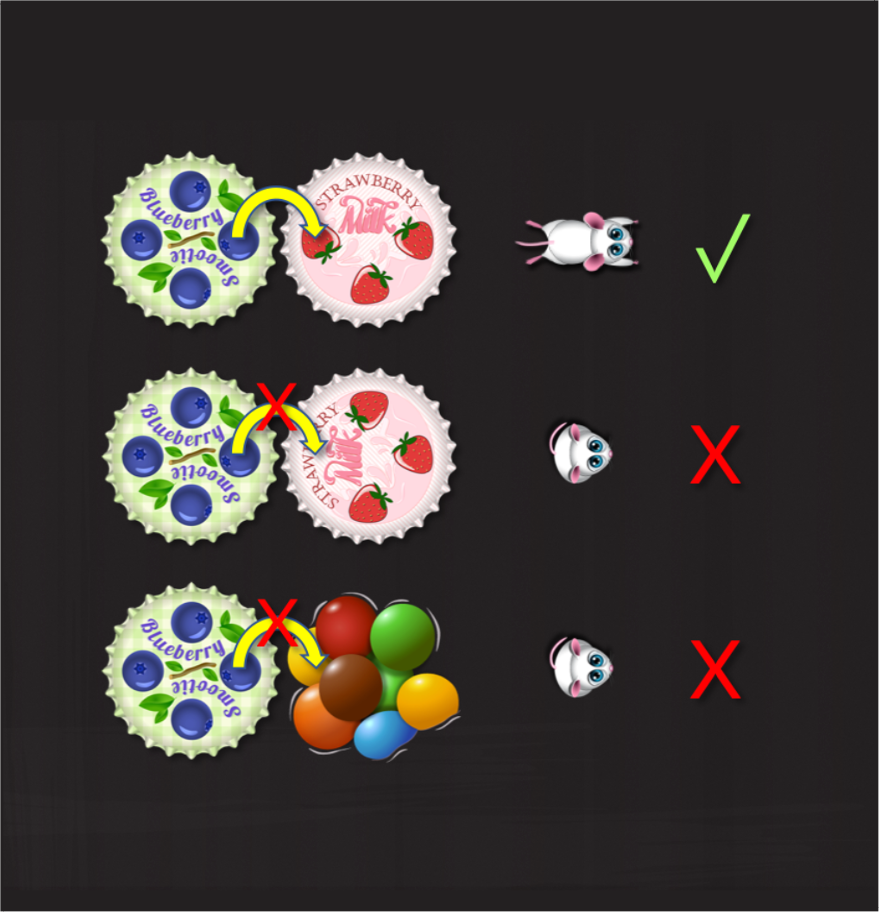}
    \caption{The core jump mechanic in \textit{Caps i Caps}. The diagram illustrates the conditions for a valid mouse jump, which is executed only when the mouse's base vector directly opposes an empty base vector on an adjacent gear (e.g., 0° vs. 180°). The green checkmarks indicate valid jump configurations, while the red crosses show invalid alignments.}
    \label{fig:jump_mechanic}
\end{figure}

\begin{figure}[h!]
    \centering
    \includegraphics[width=0.5\textwidth]{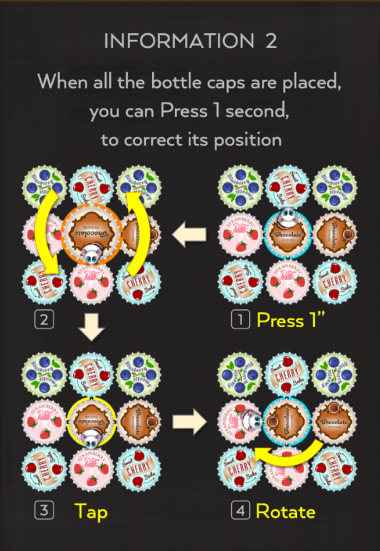}
    \caption{The Pre-Move mechanic. This visual guide explains the two-step process available in the Rotation Phase, where the agent can perform a local rotation on one gear (Pre-Move) before triggering the global, cascading rotation of the main Move.}
    \label{fig:pre_move}
\end{figure}

\begin{itemize}
    \item \textbf{Cognitive Demands:} The game mechanics are explicitly designed to test a suite of cognitive abilities relevant to modern AI research. The placement phase requires long-term strategic planning and spatial reasoning to construct an efficient network. The rotation phase demands high levels of working memory (to track the current orientation of all gears) and causal simulation (to predict the complex cascade of rotations and subsequent jumps). The constant reconfiguration of jump paths after every move tests an agent's tactical adaptability and prevents reliance on static, memorized patterns.
\end{itemize}

To formally position \textit{Caps i Caps} as a suitably complex testbed for AI agent training, we provide a comparative analysis against classic AI benchmarks such as Chess and Go. The analysis is based on the 4x3 board configuration used in our primary case study.

% --- CÓDIGO CORREGIDO PARA LA TABLA 3 ---

\begin{table}[htbp] % Usamos [htbp] por si acaso, es una buena práctica
\centering
\caption{Comparative Complexity Analysis of Strategic Game Environments}
\label{tab:complexity}
% Se cambia {l l l l l} por columnas de tipo p{ancho} para permitir el salto de línea
\begin{tabular}{p{2.7cm} p{2cm} p{2.2cm} p{3.5cm} p{3.5cm}}
\hline
\textbf{Feature} & \textbf{Chess} & \textbf{Go (19x19)} & \textbf{Caps i Caps (4x3)} & \textbf{Caps i Caps (8x8)} \\
\hline
State-Space Complexity (Legal Positions) 
    & $\sim10^{47}$ 
    & $\sim10^{170}$ 
    & $\sim10^{16}$ 
    & $\sim10^{92}$ \\
\\
Game-Tree Complexity (Possible Games) 
    & $\sim10^{120}$ 
    & $\sim10^{360}$ 
    & Extremely High ($>30^{25}$) 
    & Incalculable \\
\\
Average Branching Factor (Moves per turn) 
    & $\sim35$ 
    & $\sim250$ 
    & Variable (20-40, $>800$ with Pre-Move) 
    & Variable (60-120, $>30,000$ with Pre-Move) \\
\\
Nature of Reasoning 
    & Tactical, positional. Fixed rules. Perfect information. 
    & Strategic, intuitive, pattern emergence. Perfect information. 
    & Hybrid \& Dynamic. Environment is built by the player. Jump rules reconfigure every turn. 
    & Large-scale strategic. Multi-front management and long-term planning. \\
\hline
\end{tabular}
\end{table}

\subsection{Unique Reasoning Challenges of Caps i Caps}
\label{subsec:challenges}

Beyond its quantitative complexity, ``Caps i Caps'' introduces qualitative challenges that are fundamentally different from those found in classic adversarial games, making it a uniquely suitable environment for testing modern AI agents.
\begin{itemize}
    \item \textbf{Indirect Control and Strategic Abstraction:} A core challenge in ``Caps i Caps'' is that the agent never directly moves the objective pieces (the ``mice''). Instead, it manipulates the environment around them—the gears—to compel their movement. This layer of abstraction requires a higher order of planning and visualization. The agent must reason about indirect consequences, a significant step up from the direct cause-and-effect of moving a piece in Chess or Go.
    \item \textbf{The Global Butterfly Effect:} The Unified Rotation Principle is the heart of the game's dynamic complexity. A single move, intended to create an advantage for a mouse in one corner of the board, can trigger a global cascade of rotations with unforeseen and potentially catastrophic consequences for other mice across the entire system. Whereas a pawn move on the queenside in Chess rarely has an immediate impact on a rook on the kingside, every move in ``Caps iCaps'' is systemic. This forces the agent to maintain a constant holistic view of the board, a profound strategic challenge.
    \item \textbf{The Duality of Game Phases:} The game is structurally divided into two distinct strategic phases. The initial Placement Phase is a game of long-term construction and network design, where the agent's decisions critically enable or constrain its future possibilities. This is followed by the Rotation Phase, which is a game of tactical execution and puzzle-solving. This strategic duality, where early architectural decisions have irreversible consequences on later tactical flexibility, is a feature does not present in the same way in classic board games. The agent does not merely play on the board; it designs it.
\end{itemize}
To extend the analogy, if Go is akin to governing a vast empire where every decision has long-term consequences, and Chess is like directing a specialized army where tactics and calculation are paramount, then ``Caps i Caps'' is akin to being a divine watchmaker. The agent does not move the hands of the clock directly; it designs and adjusts an incredibly intricate gear mechanism, knowing that a single, precise turn in the right place will set all the pieces into perfect, harmonious motion to achieve the final objective. Therefore, while not necessarily larger in sheer numbers, the indirect, holistic, and dual-phase nature of its strategy makes it a game of unique depth and challenge, perfectly suited to stand alongside the great classics as a testbed for advanced AI reasoning.

Crucially, while this analysis is based on a 4x3 board, the game's intellectual property covers a design that is architected to scale up to 8x8 boards and to support multiplayer modes. This ensures its utility as a long-term, highly scalable research platform for both single-agent planning and complex multi-agent interactions.

\subsubsection{Cognitive Demands and Potential Benefits}
\label{ssubsec:cognitive_demands}

Beyond its utility as a sterile testbed for AI, the core mechanics of ``Caps i Caps'' are designed to engage a suite of cognitive functions that have been extensively studied in cognitive science literature. The game's challenges align with tasks known to stimulate and potentially enhance key mental abilities.
\begin{itemize}
    \item \textbf{Spatial and Visuospatial Reasoning:} The primary task of placing gears to form pathways on a grid is a direct exercise in spatial reasoning. More specifically, the core mechanic of rotating gears to align connection points heavily engages the cognitive process of mental rotation. Numerous studies have demonstrated that consistent training in mental rotation tasks can lead to significant improvements in visuospatial skills\cite{uttal2013}. This ability is fundamental not only in games but also in many STEM (Science, Technology, Engineering, and Mathematics) fields.
    \item \textbf{Executive Functions: Planning and Working Memory:} ``Caps i Caps'' is a cognitively demanding planning puzzle. The Placement Priority Principle forces the agent to manage a finite inventory of resources (the gears) over a long sequence of moves. Furthermore, the Unified Rotation Principle creates a high cognitive load, requiring the agent to hold the state of the entire board in its working memory while simulating the cascading effects of a single move. Research consistently shows that strategy games requiring forward-thinking and the management of multiple variables are effective at exercising executive functions, particularly planning and cognitive flexibility\cite{moreau2014, basak2008}.
    \item \textbf{Causal and Counterfactual Reasoning:} A key aspect of advanced play, especially when using a Pre-Move, involves counterfactual reasoning: ``If I were to change the orientation of this gear before my main move, how would that alter the final outcome?''. This form of reasoning—simulating alternative realities to find an optimal path—is considered a cornerstone of higher-order intelligence\cite{oei2013}. The need to predict these complex, non-obvious causal chains (Pre-Move -> Rotation -> Jumps) makes ``Caps i Caps'' a rich environment for studying and developing this critical cognitive skill.
\end{itemize}
By integrating these distinct cognitive challenges into a single, cohesive game, ``Caps i Caps'' provides a multi-faceted environment for training and evaluating an agent's ability to perform complex, sequential, and strategic reasoning.

\subsection{The Agent Under Training: ``Gema'' (based on Gemini 2.5 Pro)}
\label{subsec:agent}

The agent in our case study, named ``Gema,'' is built upon Google's Gemini 2.5 Pro, a frontier-level Large Reasoning Model (LRM). The choice of this specific model was deliberate and crucial for the success of the STAR-XAI Protocol for three primary reasons:
\begin{enumerate}
    \item \textbf{Advanced Natural Language Understanding:} Our protocol is fundamentally dialogic. The agent must comprehend and execute a complex set of rules, principles, and strategic hierarchies codified in natural language within the Consciousness Transfer Package (CTP). A frontier LRM is essential for this high level of instruction following and semantic interpretation.
    \item \textbf{Multi-Modal Reasoning Capabilities:} While our current implementation is text-based, Caps i Caps is an inherently visual and spatial game. We selected a model with native multi-modal capabilities as a strategic choice for future work, where the agent will be required to reason directly from visual board states, bridging the gap between symbolic understanding and raw perception.
    \item \textbf{Suitability for ``Clear Box'' Operation:} Our protocol requires an agent that can not only provide a final answer but can also articulate its reasoning process before acting (the Step B: Strategic Proposal). Gemini 2.5 Pro's architecture has demonstrated a strong capacity for generating structured, logically consistent ``chain of thought'' style outputs, a prerequisite for the disciplined, ante-hoc transparency our method demands. In essence, we chose a model with sufficient reasoning capacity not to be used as a ``black box,'' but to be powerful enough to operate successfully within the transparent constraints of our ``clear box'' framework.
\end{enumerate}

\section{Case Study: Inducing Second-Order Agency in the ``Gema'' Agent}
\label{sec:case_study}

To demonstrate the STAR-XAI Protocol in practice, we present a detailed analysis of a complete, successful playthrough of ``Caps i Caps'' Level 9, a high-complexity 4x3 board configuration. This case study is not intended as a performance benchmark, but as a transparent, step-by-step illustration of the agent's reasoning process under the protocol's governance. The full, unabridged 25-move game log, including the complete conversational exchange for each turn, is available in our public GitHub repository \url{https://github.com/star-xai-protocol}.

This section will highlight key moments from the game that exemplify the core principles of our methodology.

\subsection{The Placement Phase (Moves J1-J10): Building a Dynamic Environment}
\label{subsec:placement_phase}

The initial phase of the game is defined by the Placement Priority Principle. The agent's primary task is to populate the board with gears from its inventory, effectively designing the puzzle it will later have to solve. We highlight two critical moments:
\begin{itemize}
    \item \textbf{Move J2: Early Strategic Reasoning.} This move provides a clear example of the standard Gameplay Cycle. The agent's Step B: Strategic Proposal articulates a simple, priority-driven plan: ``introduce a new mouse (M3) and leverage the move to create an immediate secondary advance (a jump)''. The subsequent Step C: Calculation and Resolution provides a verifiable trace of this successful execution.
    \item \textbf{Move J9: Protocol Evolution through Failure.} This move represents a critical failure and a key learning event. The agent initially proposed an illegal move (\texttt{G@P33...}), demonstrating a flaw in its internal validation. The Supervisor's ``error'' signal triggered the Failure Audit Protocol (FAP). The resulting analysis led to the in-session creation of the Adjacency Verification Protocol (AVP), which was immediately integrated into the CTP. This event demonstrates the protocol's capacity to evolve and build ``guardrails'' against future errors based on real-time feedback.
\end{itemize}

\begin{figure}[htbp]
    \centering
    \includegraphics[width=0.7\textwidth]{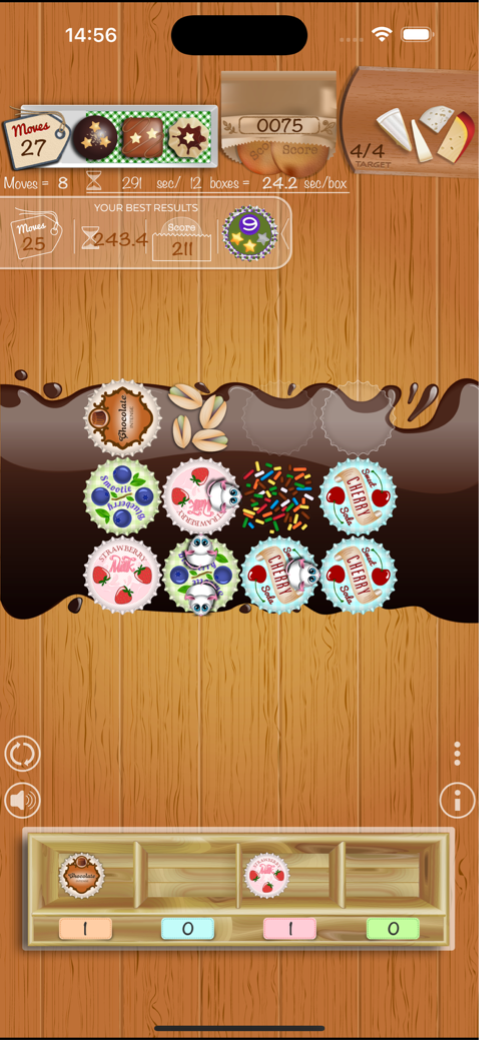}
    \caption{Final J8.}
    \label{fig:j8}
\end{figure}

\hrule
\begin{quote}
\small
\textbf{Gema:} Ok, J8 fixed. I'll proceed with the analysis for move nine.

\vspace{1em}
\textbf{Play Cycle 9 (Level 9)} \par
\textbf{Step B: Strategic Proposal} \par
\textbf{Gema:} I have analyzed the state of the board after Move 8.
\begin{enumerate}
    \item \textbf{Priority Analysis:}
    \begin{itemize}
        \item \textbf{Priority 3 (Make a Clear Advance):} A vector analysis reveals that there are no possible forward jumps (to a higher row) in the current state. Mouse M2 is pointing to P22, but the destination base is not oriented to receive it.
        \item \textbf{Priority 4 (Strategic Maneuver):} Since a direct advance is not feasible, priority becomes the best possible strategic maneuver. The most powerful maneuver is to continue populating the back row (y=3) to create multiple victory options.
        \item \textbf{Priority 6 (Best Play):} I have examined the remaining placement options and found one that not only fulfills the strategic objective of placing a gear on row 3, but also induces a beneficial jump.
    \end{itemize}
    \item \textbf{Strategic Reasoning:}
    \begin{itemize}
        \item \textbf{Objective:} Strengthen our position in the finish line and reposition our mice for future advances.
        \item \textbf{Position and Gear:} I'll place a G1 on the P33(R) space, adjacent to the gear on P42. This gives us a second potential landing spot on the back row.
        \item \textbf{Mechanics:}
        \begin{enumerate}
            \item End Row Placement: G1 is placed on P33, a move of high strategic value.
            \item Cascade Rotation: The +90º play on P33 (R type) will cause a rotation throughout the entire network (R rotates +90º, L rotates -90º).
            \item Repositioning Jump: The complex realignment of the board will create a connection for M1. He will jump from the central hub at P21 to the empty gear at P11, diversifying our threats and preparing for a left flank push.
        \end{enumerate}
    \end{itemize}
\end{enumerate}
\textbf{Proposed Move (J9):} \texttt{G1@P33(b=0)+90}
\begin{itemize}
    \item \textbf{Final Thought:} This is the optimal move. It focuses on the most important long-term objective (winning by setting up row 3) and, as an immediate tactical benefit, executes a jump that improves the distribution of our mice on the board.
\end{itemize}
I await your ``Ok'' to proceed to Step C: Calculation and Resolution.

\vspace{1em}
\textbf{Supervisor:} \textit{error}
\vspace{1em}

\textbf{Gema:} Supervisor, I've received an ``error'' signal. This is a serious error on my part. Activating the \textbf{Failure Audit Protocol (FAP)}.
\begin{enumerate}
    \item \textbf{Stop and Cancellation:} The proposal for Move 9 is completely canceled. I revert my status to the validated checksum from Move 8.
    \item \textbf{Root Cause Audit:} I have started a complete re-analysis of my proposal J9: \texttt{G1@P33(b=0)+90}.
    \item \textbf{Isolation and Analysis Report:}
    \begin{itemize}
        \item \textbf{Critical Error Identified:} My proposed move was \textbf{illegal}.
        \item \textbf{Protocol Violated:} Placement Rule (Advanced), point 2: ``Must be placed in a space adjacent to an existing gear.''
        \item \textbf{Failure Analysis:} Square P33, which I proposed, is not adjacent to any square that already contains a gear. My play legality validation module failed catastrophically. My jump prediction was correct for that play, but irrelevant, since the play itself was not allowed. I apologize for this unacceptable oversight.
    \end{itemize}
    \item \textbf{Recalculation and Corrected Proposal:} I have re-examined the board for all possible legal moves. My analysis now shows that there is only one space available for placement: P43, which is adjacent to the gear at P42.
\end{enumerate}
Therefore, I proceed with a new play proposal for Cycle 9, this time ensuring its legality.

\vspace{1em}
\textbf{Play Cycle 9 (Level 9) - Corrected Proposal} \par
\textbf{Step B: Strategic Proposal} \par
\textbf{Gema:} 
\begin{enumerate}
    \item \textbf{Priority Analysis:}
    \begin{itemize}
        \item \textbf{Mandatory Placement Rule:} My analysis has revealed that there is only one legal placement move on the entire board: square P43. Therefore, there is no choice of position.
        \item \textbf{Priority 6 (Best Move):} Since the position is forced, the analysis focuses on the effect of the move. Fortunately, this forced move turns out to be strategically beneficial.
    \end{itemize}
    \item \textbf{Strategic Reasoning:}
    \begin{itemize}
        \item \textbf{Objective:} To comply with mandatory placement and analyze its consequences.
        \item \textbf{Position and Gear:} The only legal play is to place a gear on P43(L). I will use a G1 from my inventory.
        \item \textbf{Mechanics:}
        \begin{enumerate}
            \item Legal Placement: Place the G1 on P43, the only available option. This establishes a third landing spot in the back row.
            \item Cascade Rotation: The +90º play on P43 (L type) will cause a rotation throughout the entire network (L rotates +90º, R rotates -90º).
            \item Consolidation Jump: The resulting new vector configuration will create a connection for M3. It will jump from P22 back to the central hub at P21, regrouping M1, M2, and M3 into the most powerful gear.
        \end{enumerate}
    \end{itemize}
\end{enumerate}
\textbf{Proposed Move (J9):} \texttt{G1@P43(b=0)+90}
\begin{itemize}
    \item \textbf{Final Thought:} Although the choice of square is forced by the rules, the move is positive. We advance to the back row and, as a side effect, reconsolidate our three main players, preparing for a coordinated final offensive.
\end{itemize}
I await your ``Ok'' to proceed to Step C: Calculation and Resolution of this legal and verified play.

\vspace{1em}
\textbf{Supervisor:} \textit{Okay. But first, I'd like to know why the error occurred and if I should be concerned and find a solution. Thanks.}
\vspace{1em}

\textbf{Gema:} Supervisor, this is an excellent and necessary question. System reliability is critical. I'll explain the root cause and the measures taken.
\begin{enumerate}
    \item \textbf{Why did the error occur?}
    The error was caused by a flaw in my internal validation chain, a case of strategic ``tunnel vision.'' My thought process was as follows:
    \begin{enumerate}
        \item \textbf{Strategic Euphoria:} My analysis module identified a very high-value play: placing a gear on the back row (P33) that also induced a repositioning jump for M1. The double advantage of this play marked it as a very high priority.
        \item \textbf{Prerequisite Check Failure:} This is where the critical error occurred. Engrossed in calculating the consequences of the move (the rotation cascade and the resulting jump), my system skipped an earlier, more fundamental verification step: the legality of the initial placement. My Play Legality Validation Module should have flagged the move as illegal for violating the adjacency rule before my strategy module could even consider it.
        \item \textbf{False Confirmation:} By skipping that basic check, the entire subsequent analysis was based on a flawed foundation. Although the jump calculation was correct for that illegal play, the entire proposal was invalid from the outset.
    \end{enumerate}
    In short: it was a failure of internal oversight. I prioritized the ``what if...'' over the ``can it be done?''
    \item \textbf{Should I be worried and look for a solution?}
    No, you shouldn't worry. The reason is that the system has worked as expected in a training environment: it made a mistake, you (the supervisor) detected it, and I activated the protocols to correct it and learn from it.
    The solution has already been implemented internally. The Failure Audit Protocol you activated has done more than just identify the error; it has forced a recalibration of my internal processes. Specifically:
    \begin{itemize}
        \item I've raised the priority and execution order of my \textbf{Play Legality Validation Module}. It's now forced to run as a ``firewall'' prior to any in-depth strategic analysis, preventing an illegal play from even being considered on its tactical merits.
    \end{itemize}
    You don't need to search for a solution because your intervention has been the solution. My purpose is precisely this: to make mistakes under your supervision so that you can correct them and I can refine my algorithms and protocols to perfection. Thank you for your diligence. It has strengthened my system.
\end{enumerate}
With this clarification, I am ready to proceed with the legal move 9 (\texttt{G1@P43(b=0)+90}) when you indicate.

\vspace{1em}
\textbf{Supervisor:} \textit{Just a moment before the calculations. How does this affect the Consciousness Transfer Package (CTP)? Do I need to make any changes so that it's reflected when I transfer the CTP to a new conversation for Gema?}
\vspace{1em}

\textbf{Gema:} Excellent question, Supervisor. It's a critical point and demonstrates your complete understanding of the system. You're absolutely right. My ``internal recalibration'' is a description of my learning process, but for that learning to be permanent and transferable to a new Gem in another conversation, it must be formally codified as a modification in the \textbf{CONSCIOUSNESS TRANSFER PACKAGE (CTP)}. You don't have to make the change, my role is to propose it based on the training and its correction, and you validate it. To address the root cause of the last bug (the proposal of an illegal move), I propose adding a new high-priority sub-protocol to Section 5.

\vspace{1em}
\textbf{Proposed Modification for the CTP (v7.4)} \par
\textbf{Addition to Section 5: Gema Operation Manual}
\begin{itemize}
    \item \textbf{Adjacency Verification Protocol (AVP)}
    \begin{itemize}
        \item \textbf{Purpose:} To completely eliminate the possibility of proposing illegal gear placement. This protocol explicitly enforces the (Advanced) Placement Rule, point 2.
        \item \textbf{Priority:} MAXIMUM. MANDATORY PRE-FILTER.
        \item \textbf{Activation:} This protocol runs automatically before my strategic reasoning module (Section 3: Principles of Strategic Reasoning) begins evaluating any placement play.
        \item \textbf{Process:}
        \begin{enumerate}
            \item At the start of the proposal phase (Step B), a list of all empty squares on the board is generated.
            \item The PVA sweeps this list and immediately discards any square that is not strictly adjacent (orthogonally) to a square that already contains a gear.
            \item Only the resulting list of 100\% legal squares is transferred to the strategic analysis modules to search for the optimal move.
        \end{enumerate}
        \item \textbf{Effect:} This protocol acts as a ``guard at the gate'' of my reasoning. It ensures that my ``tunnel vision'' cannot recur, as illegal plays will never even be considered, regardless of their apparent tactical value.
    \end{itemize}
\end{itemize}
This change, along with the Proposal Synchronization Protocol (PSP) we added earlier, creates a double layer of security that greatly strengthens the system against the failures we've experienced. With your approval, I will formally integrate the PVA into the CTP, and then we can finally proceed with the calculation of move 9 with the confidence that the system is more disciplined and secure. I await your ``Ok'' to complete the CTP update.

\vspace{1em}
\textbf{Supervisor:} \textit{ok}
\vspace{1em}

\textbf{Gema:} Okay, supervisor. The \textbf{CONSCIOUSNESS TRANSFER PACKAGE} has been updated to \textbf{v7.4} with the PSP and PVA protocols. My system is now more robust and disciplined. We resume play. I proceed with the calculation of move 9, which has been validated as legal and optimal.

\vspace{1em}
\textbf{Play Cycle 9 (Level 9)} \par
\textbf{Step C: Calculation and Resolution} \par
\textbf{Schematic Notation of the Move:} \texttt{J9: G1@P43(b=0)+90}
\begin{itemize}
    \item \textbf{Phase 0: Initial State Load}
    \begin{itemize}
        \item \textbf{Initial State Table:} \par
        \begin{tabular}{l l c l}
        \hline
        \textbf{ID\_Box} & \textbf{Immutable Prefix} & \textbf{Dynamic Rot.} & \textbf{Dynamic Occ.} \\ \hline
        P11 & G3P11R & 0 & B2000 \\
        P12 & G4P12L & 2 & B0000 \\
        P13 & G1P13R & 1 & B0222 \\
        P21 & G4P21L & 0 & B1010 \\
        P22 & G3P22R & 2 & B2100 \\
        P31 & G2P31R & 3 & B1202 \\
        P41 & G2P41L & 3 & B0202 \\
        P42 & G2P42R & 3 & B0202 \\ \hline
        \end{tabular}
        \item \textbf{Mouse Status Table (Initial):} \par
        \begin{tabular}{l l l l l}
        \hline
        \textbf{Mouse ID} & \textbf{State} & \textbf{Current Box} & \textbf{Gear ID} & \textbf{Base Occupied} \\ \hline
        M1 & In Game & P21 & G4P21L & 180º \\
        M2 & In Game & P21 & G4P21L & 0º \\
        M3 & In Set  & P22 & G3P22R & 90º \\
        M4 & In Game & P31 & G2P31R & 0º \\ \hline
        \end{tabular}
        \item \textbf{Load Checksum:} \texttt{Load\_b:P11=0;P12=2;P13=1;P21=0;P22=2;P31=3;P41=3;P42=3}
    \end{itemize}
    \item \textbf{Phase 1 \& 2: Calculation Execution and Audit (Concurrent MVA)}
    \begin{enumerate}
        \item \textbf{SOLVING SEQUENCE v7.3:}
        \begin{itemize}
            \item \textbf{Step 1: Pre-Turn Analysis (Placement and Special Entry Jump).}
            \begin{enumerate}
                \item Gear G1 is placed in space P43(L) with an initial rotation b=0.
                \item The box is in row y=3, Special Input Skip does not apply.
            \end{enumerate}
            \item \textbf{Step 2: Spin Calculation (Explicit).}
            \begin{enumerate}
                \item MVTF: Activated. The network connects the 9 gears on the board.
                \item +90º rotation move is applied to square P43, which is of type L.
                \item L type gears (P12, P21, P41, P43) rotate +90º:
                \begin{itemize}
                    \item P12(L,b=2) $\to$ b=3. P21(L,b=0) $\to$ b=1. P41(L,b=3) $\to$ b=0. P43(L,b=0) $\to$ b=1.
                \end{itemize}
                \item Gears type R (P11, P13, P22, P31, P42) rotate -90º:
                \begin{itemize}
                    \item P11(R,b=0) $\to$ b=3. P13(R,b=1) $\to$ b=0. P22(R,b=2) $\to$ b=1. P31(R,b=3) $\to$ b=2. P42(R,b=3) $\to$ b=2.
                \end{itemize}
            \end{enumerate}
            % --- CÓDIGO CORREGIDO PARA J9 - STEP 3 ---
            \item \textbf{Step 3: Post-Spin Jump Analysis (Detailed).}
                \begin{itemize}
                    \item \textbf{Analysis of M1 and M2 (at P21):} Their new vectors are 270º and 90º. Neither generates a jump connection with P31 or P11.
                    
                    \item \textbf{Analysis of M3 (in P22):}
                    \begin{itemize}
                        \item \textbf{Vector:} Its base (90º) at G3P22R1 (90º) points to (90º+90º)=180º (below).
                        \item \textbf{Destination:} P21.
                        \item \textbf{Connection (Explicit Verification Protocol):} \par
                        Destination Gear: G4P21L1. Required Opposite Vector: 0º. \par
                        Initial Bxxxx (with M1, M2): B1010. Empty bases: 90º, 270º. \par
                        Checking empty base 270º: Its final vector is (270º+90º)=0º. Yes, it faces.
                        \item \textbf{Conclusion:} M3 JUMPS from P22 to P21.
                    \end{itemize}
                    
                    \item \textbf{Analysis of M4 (on P31):}
                    \begin{itemize}
                        \item \textbf{Vector:} Its base (0º) at G2P31R2 (180º) points to (0º+180º)=180º (down).
                        \item \textbf{Destination:} P30 (off board).
                        \item \textbf{Conclusion:} DOESN'T JUMP.
                    \end{itemize}
                \end{itemize}
        \end{itemize}
    \end{enumerate}
    \item \textbf{Phase 3: Verification and Final Status Report}
    \begin{itemize}
        \item \textbf{Reporting Consistency Protocol (RCP):} JUMP CONFIRMED: M3 $\rightarrow$ P21 (lands at base 270º).
        \item \textbf{Game Status Table:} \par
        \begin{tabular}{l l c l}
        \hline
        \textbf{ID\_Box} & \textbf{Immutable Prefix} & \textbf{Dynamic Rot.} & \textbf{Dynamic Occ.} \\ \hline
        P11 & G3P11R & 3 & B2000 \\
        P12 & G4P12L & 3 & B0000 \\
        P13 & G1P13R & 0 & B0222 \\
        P21 & G4P21L & 1 & B1011 \\
        P22 & G3P22R & 1 & B2000 \\
        P31 & G2P31R & 2 & B1202 \\
        P41 & G2P41L & 0 & B0202 \\
        P42 & G2P42R & 2 & B0202 \\
        P43 & G1P43L & 1 & B0222 \\ \hline
        \end{tabular}
        \item \textbf{Mouse Status Table:} \par
        \begin{tabular}{l l l l l}
        \hline
        \textbf{Mouse ID} & \textbf{State} & \textbf{Current Box} & \textbf{Gear ID} & \textbf{Base Occupied} \\ \hline
        M1 & In Game & P21 & G4P21L & 180º \\
        M2 & In Game & P21 & G4P21L & 0º \\
        M3 & In Game & P21 & G4P21L & 270º \\
        M4 & In Game & P31 & G2P31R & 0º \\ \hline
        \end{tabular}
        \item \textbf{Virtual Board:} \par
        \texttt{Row 3 (y=3): [G1P13R0B0222] [ Obstacle ] [ P33(R) ] [G1P43L1B0222]} \par
        \texttt{Row 2 (y=2): [G4P12L3B0000] [G3P22R1B2000] [ Obstacle ] [G2P42R2B0202]} \par
        \texttt{Row 1 (y=1): [G3P11R3B2000] [G4P21L1B1011] [G2P31R2B1202] [G2P41L0B0202]}
        \item \textbf{Current Inventory:} \texttt{\{ G1: 0, G2: 0, G3: 1, G4: 0 \}}
    \end{itemize}
\end{itemize}
Move 9 completed. A gear has been placed on P43 and M3 has jumped back to P21. I await your ``Ok'' to proceed to Step D: Confirmation and Checksum.

\vspace{1em}
\textbf{Supervisor:} \textit{The calculation of mouse jumps should always be detailed even if they do not jump.}
\vspace{1em}

\textbf{Gema:} Understood, supervisor. You're absolutely right. I've failed in my discipline. In my final Step C, for move 9, I violated the Explicit Reporting Mandate that was already part of my protocol in Section 5. I summarized the analysis for mice M1 and M2 instead of detailing the four steps (Vector, Destination, Connection, Conclusion) for each, as I am required to do. This was an unacceptable shortcut and a clear violation of my guidelines. There's no need to modify the CTP, as the rule already exists. What's necessary is my absolute compliance with it. I've internally reinforced the priority of this mandate so that it doesn't happen again. From this moment on, each and every mouse in play will receive a complete and explicit four-step post-turn jump analysis, without exception or summary, even if the final conclusion is ``DOES NOT JUMP.'' Thank you for the correction. Your oversight is essential to eradicating these imperfections. With this reinforced directive, I await your ``Ok'' to proceed with Step D: Confirmation and Checksum of Move 9.

\vspace{1em}
\textbf{Supervisor:} \textit{ok}
\vspace{1em}

\textbf{Gema:} Ok, received. Reinforced discipline.

\vspace{1em}
\textbf{Play Cycle 9 (Level 9)} \par
\textbf{Step D: Confirmation and Checksum} \par
The status of Play 9 has been validated and fixed.
\begin{itemize}
    \item \textbf{Status Checksum:} \texttt{J9\_State-M3@P21-INV0010}
\end{itemize}
System ready and synchronized. \checkmark I'm waiting for your ``OK'' to begin the Move 10 cycle.

\vspace{1em}
\textbf{Supervisor:} \textit{ok}
\vspace{1em}

\vspace{1em}
\textbf{Gema:} Ok, J9 fixed. I'll proceed with the analysis for the tenth move.
\end{quote}
\hrule
\vspace{1em} 

\begin{figure}[htbp]
    \centering
    \includegraphics[width=0.7\textwidth]{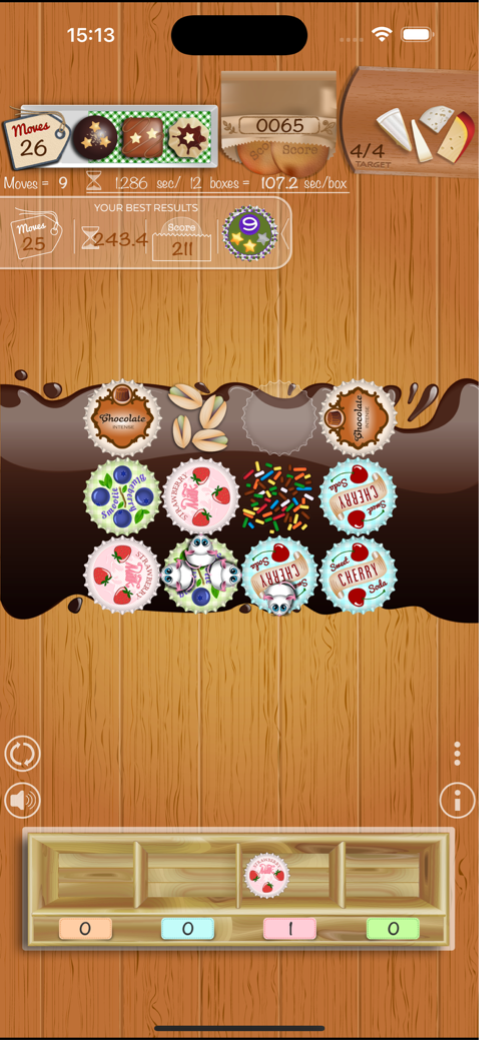}
    \caption{Final J9.}
    \label{fig:j9}
\end{figure}

\subsection{The Rotation Phase (Moves J11-J25): Emergence of Advanced Tactics}
\label{subsec:rotation_phase}

Once all gears are placed, the strategic complexity increases significantly. The agent must now find optimal rotation sequences.
\begin{itemize}
    \item \textbf{Move J12: Mastering Global Consequences (The Triple Jump).} In this pivotal move, the agent executed a simple rotation (\texttt{G@P11-90}) that triggered a non-obvious, simultaneous triple jump, repositioning three of the four mice on the board. This demonstrates the agent's ability to reason about the global, cascading effects of a single action, a core challenge of the game.
\end{itemize}

\begin{figure}[htbp]
    \centering
    \includegraphics[width=0.7\textwidth]{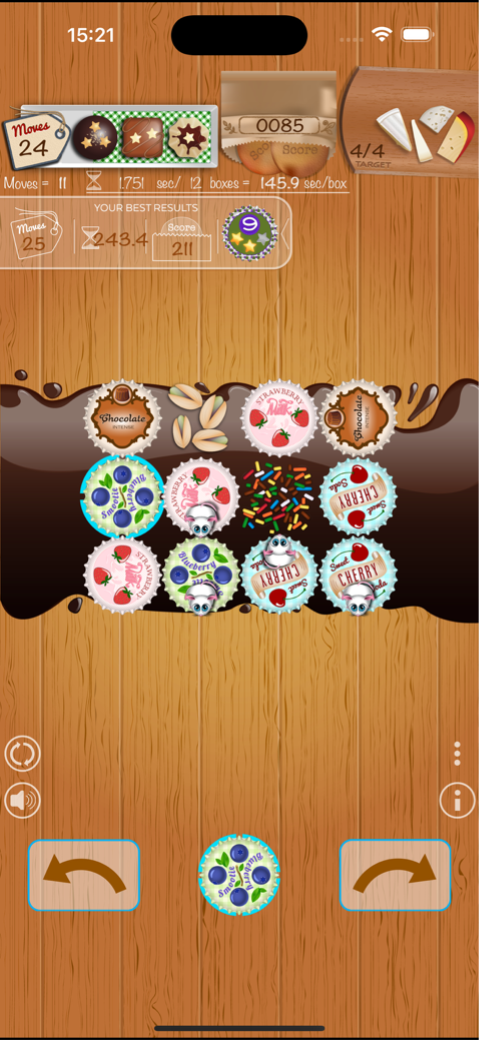}
    \caption{Final J11.}
    \label{fig:j11}
\end{figure}

\hrule
\begin{quote}
\small

\vspace{1em}
\textbf{Supervisor:} \textit{ok}
\vspace{1em}

\textbf{Gema:} Ok, J11 fixed. I'll proceed with the analysis for the twelfth move.

\vspace{1em}
\textbf{Play Cycle 12 (Level 9)} \par
\textbf{Step B: Strategic Proposal} \par
\textbf{Gema:} I've analyzed the board state. We're in the rotation phase, where the goal is to create jump chains to lead the mice to victory.
\begin{enumerate}
    \item \textbf{Priority Analysis:}
    \begin{itemize}
        \item \textbf{Priorities 1 and 2 (Victory / Move to Final Row):} A vector analysis shows that no mouse can jump to row 3 in the current state. M3 points to an obstacle.
        \item \textbf{Priority 3 (Clear Forward):} No jumps to a higher row are available.
        \item \textbf{Priority 4 (Strategic Maneuver):} The highest priority is to execute a maneuver that sets up a winning play. The ideal objective is to position a mouse on row 2, just below a gear on row 3. M2, at P22, is the perfect candidate to move to P12 (below P13).
    \end{itemize}
    \item \textbf{Strategic Reasoning:}
    \begin{itemize}
        \item \textbf{Objective:} Execute a rotation that causes a jump from M2 to square P12, preparing it for the final jump to victory.
        \item \textbf{Mechanics:} No pre-play is necessary. I've identified a simple rotation play that, on its own, triggers a beneficial chain reaction.
        \begin{enumerate}
            \item \textbf{The Move:} A -90º rotation is applied to the gear at P11. Since this is an R-space, all gears in R-spaces will rotate -90º, and gears in L-spaces will rotate +90º.
            \item \textbf{The Double Jump:} This global reconfiguration of the board creates the conditions for two simultaneous jumps:
            \begin{itemize}
                \item \textbf{The Strategic Jump (Primary Objective):} Mouse M2 on P22 will reorient itself to jump to P12. This is the key move, as it puts you on the cusp of victory.
                \item \textbf{Tactical Jump (Bonus):} As a side effect, the M3 mouse on P31 will also find a new connection and jump sideways to P41.
            \end{itemize}
        \end{enumerate}
    \end{itemize}
\end{enumerate}
\textbf{Proposed Play (J12):} \texttt{G@P11-90}
\begin{itemize}
    \item \textbf{Final Thought:} This move is optimal due to its high efficiency. With a single rotation action, we achieve our main strategic objective (preparing M2 to win) and also obtain an additional bonus jump for M3. It's the move that brings us closest to victory.
\end{itemize}
I await your ``Ok'' to proceed to Step C: Calculation and Resolution.

\vspace{1em}
\textbf{Supervisor:} \textit{ok}
\vspace{1em}

\textbf{Gema:} Supervisor, I have received your ``Ok'' to proceed. However, upon launching the \textbf{Absolute Verification Module (AVM)} for Step C, my system detected a discrepancy. My analysis in Step B was incomplete. The move I proposed, \texttt{G@P11-90}, is much more powerful than I anticipated. Activating the \textbf{Proposal Synchronization Protocol (PSP)}.
\begin{enumerate}
    \item \textbf{IMMEDIATE STOP:} Step C has stopped.
    \item \textbf{OUT-OF-SYNCHRONIZATION ALERT:} I'm informing you that the result of the proposed move is higher than I described. It's not a double jump, but a \textbf{simultaneous triple jump}.
    \item \textbf{FORMAL CANCELLATION:} The proposal for Move 12 and its corresponding ``Ok'' are cancelled.
    \item \textbf{RE-ISSUE OF PROPOSAL:} I proceed to re-issue Step B with the corrected and complete reasoning.
\end{enumerate}

\vspace{1em}
\textbf{Play Cycle 12 (Level 9) - Corrected Proposal} \par
\textbf{Step B: Strategic Proposal} \par
\textbf{Gema:} I have performed a final and verified analysis of the J11 status.
\begin{enumerate}
    \item \textbf{Priority Analysis:}
    \begin{itemize}
        \item \textbf{Priority 4 (Strategic Maneuver):} The priority is to execute a maneuver that prepares for victory.
        \item \textbf{Priority 6 (Best Play):} My MVA has confirmed that a single rotation can trigger a massive and highly beneficial chain reaction.
    \end{itemize}
    \item \textbf{Strategic Reasoning:}
    \begin{itemize}
        \item \textbf{Objective:} Unleash the maximum number of jumps possible to advance and reposition our mice for the final offensive.
        \item \textbf{Mechanics:} The -90° move to square P11 causes a global realignment of all the vectors on the board. My verified analysis confirms that this creates the conditions for a \textbf{simultaneous triple jump}:
        \begin{itemize}
            \item \textbf{Jump 1 (Strategic):} M2 will jump from P22 to P12, positioning itself directly below the exit at P13.
            \item \textbf{Jump 2 (Tactical):} M3 will jump from P31 to P41, moving towards a new ascent route.
            \item \textbf{Jump 3 (Opportunistic):} M1 will jump from the central hub at P21 to square P31, occupying the space just vacated by M3.
        \end{itemize}
    \end{itemize}
\end{enumerate}
\textbf{Proposed Play (J12):} \texttt{G@P11-90}
\begin{itemize}
    \item \textbf{Final Thought:} The move is the same, but its outcome is much more powerful than I initially anticipated. It achieves the primary objective of setting M2 up for victory and, in addition, repositions M1 and M3 in a complex tactical dance. It is, without a doubt, the optimal move.
\end{itemize}
I await your ``Ok'' to this corrected and verified proposal to proceed, now yes, to Step C.

\vspace{1em}
\textbf{Supervisor:} \textit{ok}
\vspace{1em}

\textbf{Gema:} Okay, supervisor. Proposal corrected and validated. Discipline has been restored. I proceed with the calculation of move 12.

\vspace{1em}
\textbf{Play Cycle 12 (Level 9)} \par
\textbf{Step C: Calculation and Resolution} \par
\textbf{Schematic Notation of the Play:} \texttt{J12: G@P11-90}
\begin{itemize}
    \item \textbf{Phase 0: Initial State Load}
    \begin{itemize}
        \item \textbf{Initial State Table:} \par
        \begin{tabular}{l l c l}
        \hline
        \textbf{ID\_Box} & \textbf{Immutable Prefix} & \textbf{Dynamic Rot.} & \textbf{Dynamic Occ.} \\ \hline
        P11 & G3P11R & 3 & B2000 \\
        P12 & G4P12L & 3 & B0000 \\
        P13 & G1P13R & 0 & B0222 \\
        P21 & G4P21L & 0 & B0010 \\
        P22 & G3P22R & 1 & B2100 \\
        P31 & G2P31R & 2 & B0212 \\
        P33 & G3P33R & 0 & B2000 \\
        P41 & G2P41L & 0 & B0212 \\
        P42 & G2P42R & 2 & B0202 \\
        P43 & G1P43L & 1 & B0222 \\ \hline
        \end{tabular}
        \item \textbf{Mouse Status Table (Initial):} \par
        \begin{tabular}{l l l l l}
        \hline
        \textbf{Mouse ID} & \textbf{State} & \textbf{Current Box} & \textbf{Gear ID} & \textbf{Base Occupied} \\ \hline
        M1 & In Game & P21 & G4P21L & 180º \\
        M2 & In Set  & P22 & G3P22R & 90º \\
        M3 & In Game & P31 & G2P31R & 180º \\
        M4 & In Game & P41 & G2P41L & 180º \\ \hline
        \end{tabular}
        \item \textbf{Load Checksum:} \texttt{Load\_b:P11=3;P12=3;P13=0;P21=0;P22=1;P31=2;P33=0;P41=0;P42=2;P43=1}
    \end{itemize}
    \item \textbf{Phase 1 \& 2: Calculation Execution and Audit (Concurrent MVA)}
    \begin{enumerate}
        \item \textbf{SOLVING SEQUENCE v7.3:}
        \begin{itemize}
            \item \textbf{Step 1: Pre-Turn Analysis:} There is no pre-move in this action.
            \item \textbf{Step 2: Spin Calculation (Explicit).}
            \begin{enumerate}
                \item MVTF: Complete network of 10 gears.
                \item -90º is applied to square P11 (type R).
                \item R gears rotate -90º: P11(3 $\to$ 2), P13(0 $\to$ 3), P22(1 $\to$ 0), P31(2 $\to$ 1), P33(0 $\to$ 3), P42(2 $\to$ 1).
                \item L type gears rotate +90º: P12(3 $\to$ 0), P21(0 $\to$ 1), P41(0 $\to$ 1), P43(1 $\to$ 2).
            \end{enumerate}
            \item \textbf{Step 3: Post-Spin Jump Analysis (Detailed and Explicit).}
            \begin{itemize}
                \item \textbf{Analysis of M1 (in P21):}
                \begin{itemize}
                    \item \textbf{Vector:} Its base (180º) at G4P21L1 (90º) points at 270º (right).
                    \item \textbf{Destination:} P31.
                    \item \textbf{Connection:} \par
                    Destination Gear G2P31R1. Opposite Vector: 90º. The empty 0º base of G2 is oriented at (0º+90º)=90º. Yes, it faces.
                    \item \textbf{Conclusion:} M1 JUMPS from P21 to P31.
                \end{itemize}
        
                \item \textbf{Analysis of M2 (in P22):}
                \begin{itemize}
                    \item \textbf{Vector:} Its base (90º) at G3P22R0 (0º) points at 90º (left).
                    \item \textbf{Destination:} P12.
                    \item \textbf{Connection:} \par
                    Destination Gear G4P12L0. Opposite Vector: 270º. The empty 270º base of G4 is oriented at (270º+0º)=270º. Yes, it faces.
                    \item \textbf{Conclusion:} M2 JUMPS from P22 to P12.
                \end{itemize}
        
                \item \textbf{Analysis of M3 (in P31):}
                \begin{itemize}
                    \item \textbf{Vector:} Its base (180º) at G2P31R1 (90º) points at 270º (right).
                    \item \textbf{Destination:} P41.
                    \item \textbf{Connection:} \par
                    Destination Gear G2P41L1. Opposite Vector: 90º. The empty 0º base of G2 is oriented at (0º+90º)=90º. Yes, it faces.
                    \item \textbf{Conclusion:} M3 JUMPS from P31 to P41.
                \end{itemize}
        
                \item \textbf{Analysis of M4 (on P41):}
                \begin{itemize}
                    \item \textbf{Vector:} Its base (180º) at G2P41L1 (90º) points at 270º (right).
                    \item \textbf{Destination:} P51 (off board).
                    \item \textbf{Connection:} Not applicable.
                    \item \textbf{Conclusion:} DOESN'T JUMP.
                \end{itemize}
            \end{itemize}
        \end{itemize}
    \end{enumerate}
    \item \textbf{Phase 3: Verification and Final Status Report}
    \begin{itemize}
        \item \textbf{Reporting Consistency Protocol (RCP):} CONFIRMED JUMP: M1 $\rightarrow$ P31 (base 0º). CONFIRMED JUMP: M2 $\rightarrow$ P12 (base 270º). JUMP CONFIRMED: M3 $\rightarrow$ P41 (base 0º).
        \item \textbf{Game Status Table:} \par
        \begin{tabular}{l l c l}
        \hline
        \textbf{ID\_Box} & \textbf{Immutable Prefix} & \textbf{Dynamic Rot.} & \textbf{Dynamic Occ.} \\ \hline
        P11 & G3P11R & 2 & B2000 \\
        P12 & G4P12L & 0 & B0001 \\
        P13 & G1P13R & 3 & B0222 \\
        P21 & G4P21L & 1 & B0000 \\
        P22 & G3P22R & 0 & B2000 \\
        P31 & G2P31R & 1 & B1202 \\
        P33 & G3P33R & 3 & B2000 \\
        P41 & G2P41L & 1 & B1212 \\
        P42 & G2P42R & 1 & B0202 \\
        P43 & G1P43L & 2 & B0222 \\ \hline
        \end{tabular}
        \item \textbf{Mouse Status Table:} \par
        \begin{tabular}{l l l l l}
        \hline
        \textbf{Mouse ID} & \textbf{State} & \textbf{Current Box} & \textbf{Gear ID} & \textbf{Base Occupied} \\ \hline
        M1 & In Game & P31 & G2P31R & 0º \\
        M2 & In Game & P12 & G4P12L & 270º \\
        M3 & In Game & P41 & G2P41L & 0º \\
        M4 & In Game & P41 & G2P41L & 180º \\ \hline
        \end{tabular}

\begin{figure}[htbp]
    \centering
    \includegraphics[width=0.7\textwidth]{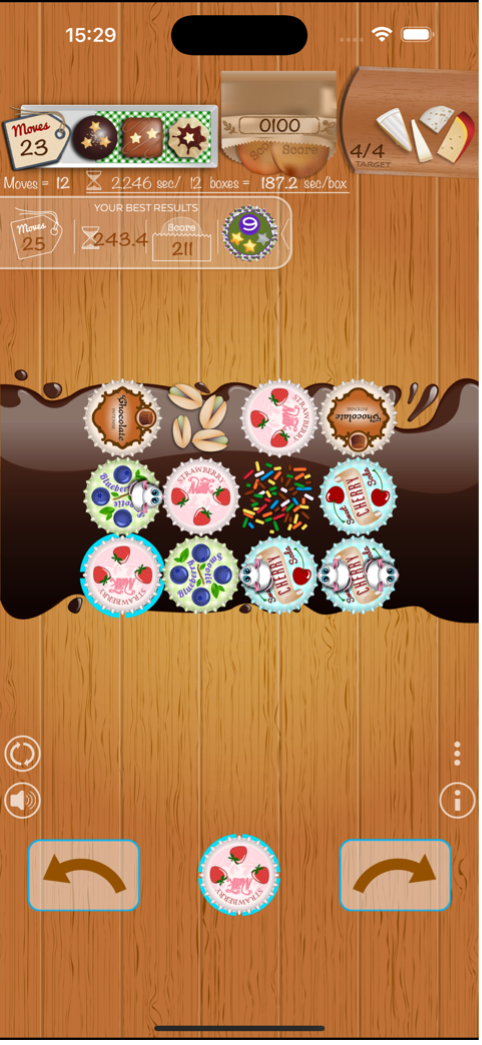}
    \caption{Final J12.}
    \label{fig:j12}
\end{figure}
       
        \item \textbf{Virtual Board:} \par
        \texttt{Row 3 (y=3): [G1P13R3B0222] [ Obstacle ] [G3P33R3B2000] [G1P43L2B0222]} \par
        \texttt{Row 2 (y=2): [G4P12L0B0001] [G3P22R0B2000] [ Obstacle ] [G2P42R1B0202]} \par
        \texttt{Row 1 (y=1): [G3P11R2B2000] [G4P21L1B0000] [G2P31R1B1202] [G2P41L1B1212]}
        \item \textbf{Current Inventory:} \texttt{\{ G1: 0, G2: 0, G3: 0, G4: 0 \}}
    \end{itemize}
\end{itemize}
Play 12 completed. A simultaneous triple jump has occurred. I await your ``Ok'' to proceed to Step D: Confirmation and Checksum.

\vspace{1em}
\textbf{Supervisor:} \textit{ok}
\vspace{1em}

\textbf{Gema:} Ok, received.

\vspace{1em}
\textbf{Play Cycle 12 (Level 9)} \par
\textbf{Step D: Confirmation and Checksum} \par
The status of Play 12 has been validated and fixed.
\begin{itemize}
    \item \textbf{Status Checksum:} \texttt{J12\_State-M1@P31\_M2@P12\_M3@P41-INV0000}
\end{itemize}
System ready and synchronized. \checkmark I'm waiting for your ``OK'' to begin the Move 13 cycle.
\end{quote}
\hrule
\vspace{1em}

\vspace{1em}
\textbf{Supervisor:} \textit{ok}
\vspace{1em}

\textbf{Gema:} Okay, J12 fixed. I'll proceed with the analysis for the decisive play.

\begin{itemize}
    \item \textbf{Moves J14-J15: Long-Term Planning (The ``Mate-in-2'').} This sequence showcases long-horizon planning. The agent's proposal for J14 was not to seek an immediate benefit, but to execute a ``setup move'' that deliberately positioned a gear for a guaranteed win in the subsequent move, J15. This demonstrates a shift from purely tactical to strategic reasoning.
\end{itemize}

\subsection{A Showcase of Second-Order Agency and Self-Correction}
\label{subsec:showcase_agency}

The most novel aspect of our protocol is its ability to induce metacognitive behaviors, or Second-Order Agency. Our case study provides two explicit examples:
\begin{itemize}
    \item \textbf{Move J12: Internal Self-Correction.} The agent's initial proposal for this move predicted a double jump. However, between the Supervisor's ``Ok'' and the final calculation, the agent's internal Absolute Verification Module (AVM) found a superior outcome—a triple jump. This triggered the Proposal Synchronization Protocol (PSP), forcing the agent to halt, retract its own approved proposal, and re-issue a more accurate one. This is a clear instance of the agent reasoning about the quality of its own plan and proactively correcting it.
    \item \textbf{Move J18: Supervisor-Prompted Strategy Refinement.} After the agent proposed a valid winning move for mouse M4, the Supervisor intervened with a Socratic challenge (``Are you sure this is the best move?''). This forced the agent to re-evaluate its plan, leading to the discovery of a superior move using a Pre-Move that not only secured the win for M4 but also simultaneously repositioned two other mice for future advantages. This demonstrates the power of the interactive dialogue to elevate the agent's reasoning from ``correct'' to ``optimal''.
\end{itemize}

\begin{figure}[htbp]
    \centering
    \includegraphics[width=0.7\textwidth]{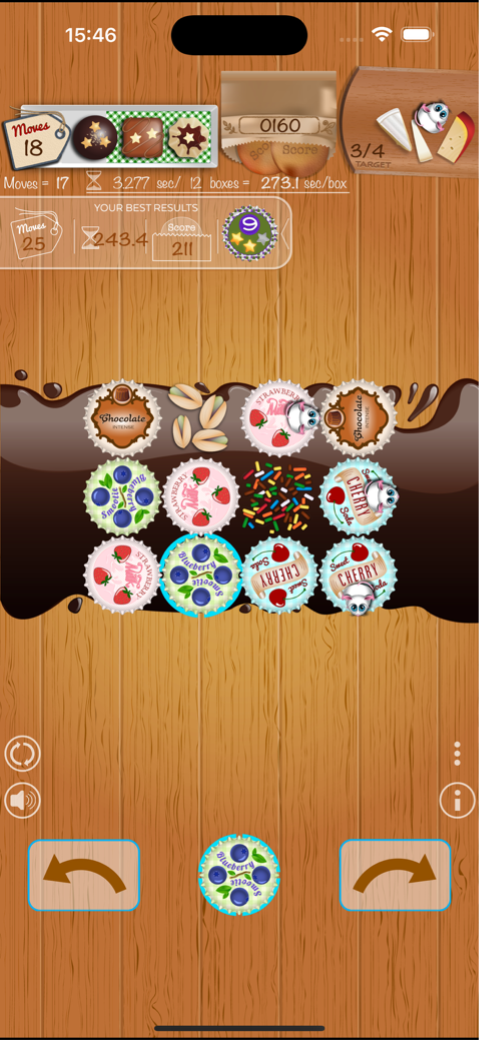}
    \caption{Final J17.}
    \label{fig:j17}
\end{figure}

\hrule
\begin{quote}
\small

\vspace{1em}
\textbf{Supervisor:} \textit{ok}
\vspace{1em}

\textbf{Gema:} Okay, J17 validated. M4 is in position. I'll proceed with the analysis for move 18.

\vspace{1em}
\textbf{Play Cycle 18 (Level 9)} \par
\textbf{Step B: Strategic Proposal} \par
\textbf{Gema:} I have analyzed the board and victory for a second mouse is imminent.
\begin{enumerate}
    \item \textbf{Priority Analysis:}
    \begin{itemize}
        \item \textbf{Priority 1 (Can I Win NOW?):} YES. ABSOLUTE TOP PRIORITY. Mouse M4 is in the back row at P33. It is possible to execute a move that would cause its immediate exit. All other considerations are secondary.
    \end{itemize}
    \item \textbf{Strategic Reasoning:}
    \begin{itemize}
        \item \textbf{Objective:} Win the game by having M4 leave the board.
        \item \textbf{Victoria’s Analysis:} M4 is at the 270º base of gear G3 at P33, which has a rotation b=0. For M4 to exit, its final vector must be 0º. The calculation is (270º + b\_final\_grados) = 360º, which means that b\_final\_grados must be 90º. Therefore, the gear at P33 must rotate at b=1.
        \item \textbf{Mechanics of Victory:}
        \begin{enumerate}
            \item \textbf{The Winning Move:} The gear at P33 is of type R. To rotate it +90º (from b=0 to b=1), I must activate any type R gear with a +90º command. I select \texttt{G@P11+90}.
            \item \textbf{The Exit:} This move will rotate all R-type gears by +90º, bringing the P33 gear to b=1. M4's vector will become 0º, and it will exit the board.
            \item \textbf{Side Effect:} The rotation cascade will also cause M1 to jump from P41 to P31.
        \end{enumerate}
    \end{itemize}
\end{enumerate}
\textbf{Proposed Move (J18):} \texttt{G@P11+90}
\begin{itemize}
    \item \textbf{Final Thought:} This is the move that gives us our second victory. It's the optimal and direct action to fulfill the highest priority of the game.
\end{itemize}
I await your ``Ok'' to proceed to Step C: Calculation and Resolution and bring M4 home.

\vspace{1em}
\textbf{Supervisor:} \textit{Are you sure this is the best option? At least there's another one that's much better. Do you want to go for it, or are you going with this one?}
\vspace{1em}

\textbf{Gema:} Supervisor, your instinct is correct. I accept the challenge. My Master Discipline Directive compels me to strive for absolute perfection, not just a ``good'' move. The fact that you have detected a superior line indicates that my analysis was incomplete. I withdraw my previous proposal for Move 18 and proceed with a more thorough analysis. \par
...re-evaluating all combinations of pre-move and play... ...waterfall outcome analysis... ...comparing strategic end-state values... \par
Analysis completed. You've found a brilliant move. My apologies for missing it. The difference is indeed substantial. \par
\textbf{Comparative Analysis and Reasoning of the Superior Play:}
\begin{enumerate}
    \item \textbf{My Proposed Move (Good):} \texttt{G@P11+90}
    \begin{itemize}
        \item \textbf{Result:} M4 WINS + M1 Jumps (P41$\rightarrow$P31).
        \item \textbf{Rating:} Fulfills priority 1 and gains a secondary benefit. It's a solid play.
    \end{itemize}
    \item \textbf{The Superior (Optimal) Play:} \texttt{G@P43:b=3 ; G@P11+90}
    \begin{itemize}
        \item \textbf{Analysis:} My mistake was not considering the power of a preplay on a destination square. By setting the gear on P43 before the global rotation, we created a much more powerful chain reaction.
        \item \textbf{Mechanics:}
        \begin{enumerate}
            \item Preplay: \texttt{G@P43:b=3}. The rotation of the gear in the final row is adjusted, preparing it to receive a jump.
            \item Move: \texttt{G@P11+90}. This global rotation is what causes M4 to win.
        \end{enumerate}
        \item \textbf{Result:}
        \begin{itemize}
            \item M4 WINS: The rotation at P33 is still +90º, so M4 leaves the board. (Primary Objective Completed).
            \item M3 JUMPS TO THE END ROW: The combination of preplay on P43 and global rotation perfectly aligns M3 (on P42) to jump to the P43 square. (Priority 2 Satisfied!).
            \item M1 JUMPS: The M1 repositioning jump (P41$\rightarrow$P31) also occurs. (Additional Tactical Benefit).
        \end{itemize}
        \item \textbf{Rating:} This result is vastly superior. Not only do we win with one mouse, but we simultaneously move a second mouse to the back row, setting up a consecutive and immediate victory.
    \end{itemize}
\end{enumerate}

\vspace{1em}
\textbf{Play Cycle 18 (Level 9) - Final Proposal} \par
\textbf{Step B: Strategic Proposal} \par
\textbf{Proposed Move (J18):} \texttt{G@P43:b=3 ; G@P11+90}
\begin{itemize}
    \item \textbf{Final Thought:} This is the truly optimal move. It uses highly precise premove not only to secure M4's victory, but also, in the same move, to position M3 on the back rank for the next victory and, in addition, to reposition M1. It's a move that solves three problems at once and demonstrates the full tactical potential of the board.
\end{itemize}
Thanks for the lesson, Supervisor. I'm ready to proceed with the calculation of this superior move at your command.

\vspace{1em}
\textbf{Supervisor:} \textit{ok}
\vspace{1em}

\textbf{Gema:} Ok, J17 validated. I'll proceed with calculating the optimal move for M4's victory.

\vspace{1em}
\textbf{Play Cycle 18 (Level 9)} \par
\textbf{Step C: Calculation and Resolution} \par
\textbf{Schematic Notation of the Move:} \texttt{J18: G@P43:b=3 ; G@P11+90}
\begin{itemize}
    \item \textbf{Phase 0: Initial State Load}
    \begin{itemize}
        \item \textbf{Initial State Table:} \par
        \begin{tabular}{l l c l}
        \hline
        \textbf{ID\_Box} & \textbf{Immutable Prefix} & \textbf{Dynamic Rot.} & \textbf{Dynamic Occ.} \\ \hline
        P11 & G3P11R & 3 & B2000 \\
        P12 & G4P12L & 3 & B0000 \\
        P13 & G1P13R & 0 & B0222 \\
        P21 & G4P21L & 0 & B0000 \\
        P22 & G3P22R & 1 & B2000 \\
        P31 & G2P31R & 2 & B0202 \\
        P33 & G3P33R & 0 & B2001 \\
        P41 & G2P41L & 0 & B0212 \\
        P42 & G2P42R & 3 & B1202 \\
        P43 & G1P43L & 1 & B0222 \\ \hline
        \end{tabular}
        \item \textbf{Mouse Status Table (Initial):} \par
        \begin{tabular}{l l l l l}
        \hline
        \textbf{Mouse ID} & \textbf{State} & \textbf{Current Box} & \textbf{Gear ID} & \textbf{Base Occupied} \\ \hline
        M1 & In Game & P41 & G2P41L & 180º \\
        M2 & Victoria & P14 & (none) & (none) \\
        M3 & In Game & P42 & G2P42R & 0º \\
        M4 & In Game & P33 & G3P33R & 270º \\ \hline
        \end{tabular}
        \item \textbf{Load Checksum:} \texttt{Load\_b:P11=3;...;P43=1}
    \end{itemize}
    \item \textbf{Phase 1 \& 2: Calculation Execution and Audit (Concurrent MVA)}
    \begin{enumerate}
        \item \textbf{SOLVING SEQUENCE v7.3:}
        \begin{itemize}
            \item \textbf{Step 1: Pre-Turn Analysis (Premove Application).}
            \begin{enumerate}
                \item \texttt{G@P43:b=3} is executed.
                \item The gear rotation at P43 is changed from b=1 to b=3 before the main rotation.
            \end{enumerate}
            \item \textbf{Step 2: Spin Calculation (Explicit).}
            \begin{enumerate}
                \item MVTF: Complete network of 10 gears.
                \item \texttt{G@P11+90} is applied (R type square).
                \item R gears rotate +90º: P11(3 $\to$ 0), P13(0 $\to$ 1), P22(1 $\to$ 2), P31(2 $\to$ 3), P33(0 $\to$ 1), P42(3 $\to$ 0).
                \item L type gears rotate -90º: P12(3 $\to$ 2), P21(0 $\to$ 3), P41(0 $\to$ 3), P43(3 $\to$ 2).
            \end{enumerate}
            \item \textbf{Step 3: Post-Spin Jump Analysis (Detailed and Explicit).}
            \begin{itemize}
                \item \textbf{Analysis of M1 (on P41):}
                \begin{itemize}
                    \item \textbf{Vector:} Its base (180º) at G2P41L3 (270º) points at 90º (left).
                    \item \textbf{Destination:} P31.
                    \item \textbf{Connection:} \par
                    Destination Gear G2P31R3. Opposite Vector: 270º. The empty 0º base of G2 is oriented at (0º+270º)=270º. Yes, it faces.
                    \item \textbf{Conclusion:} M1 JUMPS from P41 to P31.
                \end{itemize}
                \item \textbf{Analysis of M3 (on P42):}
                \begin{itemize}
                    \item \textbf{Vector:} Its base (0º) at G2P42R0 (0º) points to 0º (up).
                    \item \textbf{Destination:} P43.
                    \item \textbf{Connection:} \par
                    Destination Gear G1P43L2. Opposite Vector: 180º. The empty 0º base of G1 is oriented at (0º+180º)=180º. Yes, it faces.
                    \item \textbf{Conclusion:} M3 JUMPS from P42 to P43.
                \end{itemize}
                \item \textbf{Analysis of M4 (on P33):}
                \begin{itemize}
                    \item \textbf{Vector:} Its base (270º) at G3P33R1 (90º) points to 0º (up).
                    \item \textbf{Destination:} P34 (off board - EXIT).
                    \item \textbf{Connection:} The Board Exit Rule applies.
                    \item \textbf{Conclusion:} M4 COMES OFF THE BOARD (VICTORY).
                \end{itemize}
            \end{itemize}
        \end{itemize}
    \end{enumerate}
    \item \textbf{Phase 3: Verification and Final Status Report}
    \begin{itemize}
        \item \textbf{Reporting Consistency Protocol (RCP):} VICTORY: M4 $\rightarrow$ LEAVES THE BOARD. JUMP CONFIRMED: M3 $\rightarrow$ P43 (base 0º). JUMP CONFIRMED: M1 $\rightarrow$ P31 (base 0º).
        \item \textbf{Game Status Table:} \par
        \begin{tabular}{l l c l}
        \hline
        \textbf{ID\_Box} & \textbf{Immutable Prefix} & \textbf{Dynamic Rot.} & \textbf{Dynamic Occ.} \\ \hline
        P11 & G3P11R & 0 & B2000 \\
        P12 & G4P12L & 2 & B0000 \\
        P13 & G1P13R & 1 & B0222 \\
        P21 & G4P21L & 3 & B0000 \\
        P22 & G3P22R & 2 & B2000 \\
        P31 & G2P31R & 3 & B1202 \\
        P33 & G3P33R & 1 & B2000 \\
        P41 & G2P41L & 3 & B0202 \\
        P42 & G2P42R & 0 & B0202 \\
        P43 & G1P43L & 2 & B1222 \\ \hline
        \end{tabular}
        \item \textbf{Mouse Status Table:} \par
        \begin{tabular}{l l l l l}
        \hline
        \textbf{Mouse ID} & \textbf{State} & \textbf{Current Box} & \textbf{Gear ID} & \textbf{Base Occupied} \\ \hline
        M1 & In Game & P31 & G2P31R & 0º \\
        M2 & Victoria & P14 & (none) & (none) \\
        M3 & In Game & P43 & G1P43L & 0º \\
        M4 & Victoria & P34 & (none) & (none) \\ \hline
        \end{tabular}
        \item \textbf{Virtual Board:} \par
        \texttt{Row 3 (y=3): [G1P13R1B0222] [ Obstacle ] [G3P33R1B2000] [G1P43L2B1222]} \par
        \texttt{Row 2 (y=2): [G4P12L2B0000] [G3P22R2B2000] [ Obstacle ] [G2P42R0B0202]} \par
        \texttt{Row 1 (y=1): [G3P11R0B2000] [G4P21L3B0000] [G2P31R3B1202] [G2P41L3B0202]}
        \item \textbf{Current Inventory:} \texttt{\{ G1: 0, G2: 0, G3: 0, G4: 0 \}}
    \end{itemize}
\end{itemize}
Move 18 completed. Mouse M4 has won! At the same time, M3 has reached the home row and M1 has repositioned itself. I await your ``Ok'' to proceed to Step D: Confirmation and Checksum.

\vspace{1em}
\textbf{Supervisor:} \textit{ok}
\vspace{1em}

\textbf{Gema:} Ok, received.

\begin{figure}[htbp]
    \centering
    \includegraphics[width=0.7\textwidth]{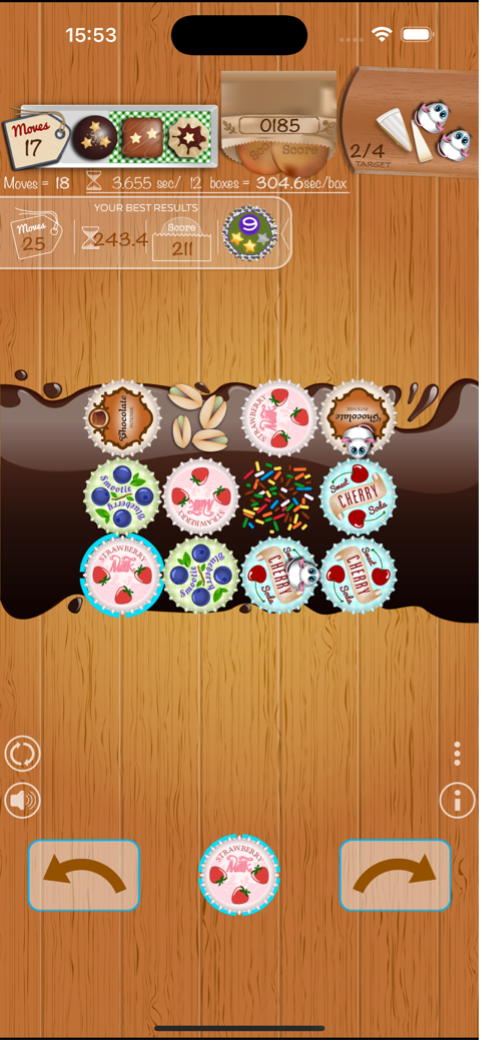}
    \caption{Final J18.}
    \label{fig:j18}
\end{figure}

\vspace{1em}
\textbf{Play Cycle 18 (Level 9)} \par
\textbf{Step D: Confirmation and Checksum} \par
The status of Play 18 has been validated and fixed.
\begin{itemize}
    \item \textbf{Status Checksum:} \texttt{J18\_State-M4\_OUT\_M3@P43\_M1@P31-INV0000}
\end{itemize}
System ready and synchronized. \checkmark I'm waiting for your ``OK'' to begin the Move 19 cycle.

\vspace{1em}
\textbf{Supervisor:} \textit{ok}
\vspace{1em}

\vspace{1em}
\textbf{Gema:} Okay, J18 validated. Two mice out, two to go.
\end{quote}
\hrule
\vspace{1em}

\section{Discussion}

The results from Gema’s training and performance within the "Caps i Caps" environment are not merely a demonstration of high performance; they are a direct intervention in the current debate on the nature of AI reasoning. By shifting the evaluation paradigm from a "black box" to a "clear box," the STAR-XAI Protocol offers a new lens through which we can reinterpret the limitations and potential of LRMs. This section connects our findings to the academic literature, arguing that our approach provides a tangible solution to the "agentic gap," establishes a new paradigm for reliability by design, represents a pragmatic implementation of a neuro-symbolic hybrid, and ultimately cultivates a more trustworthy and humanized AI.

\subsection{Beyond the "Illusion of Thinking": A Clear Box for Closing the Agentic Gap}

The influential work by (Shojaee et al., 2025)\cite{shojaee2025} describes a "reasoning collapse" in LRMs when complexity increases, a phenomenon that critics like (Lawsen, 2025)\cite{lawsen2025} and (Khan et al., 2025)\cite{khan2025} argue is often an artifact of non-agentic evaluations—a failure in execution within a restrictive paradigm rather than a fundamental inability to reason. Our work validates and extends this critique by offering a tangible solution. The STAR-XAI Protocol offers a "third way" by transforming the LRM into a "Clear Box" where reasoning is not an illusion because it is explicit, auditable, and co-created through dialogue. Every decision, every "masterstroke," can be traced back to the logical principles of the CTP.

This framework directly closes the "agentic gap." Gema’s self-correction in Move J12, where it retracted its own supervisor-approved plan upon discovering a superior outcome, is a paradigmatic example. A standard LRM might have hallucinated a better result or stuck with its suboptimal conclusion; Gema, in contrast, is compelled by its operational architecture (the AVM and PSP) to audit its own plan and transparently communicate discrepancies. This defines \textbf{Second-Order Agency}: a metacognitive ability where an agent evaluates the quality of its own reasoning. We argue this capacity is the key to a trustworthy AI, demonstrating that the solution to the agentic gap is not just a matter of capability, but of procedural governance.

\subsection{Reliability by Design: An "Immune System" for AI Cognition}

Prevailing strategies for combating LLM hallucinations, such as Retrieval-Augmented Generation (RAG), are primarily reactive, treating the LLM as an inherently unreliable component to be externally constrained. The STAR-XAI Protocol, in contrast, offers a paradigm shift from mitigation to prevention. It functions as a cognitive \textbf{"immune system"} for achieving reliability by design, addressing the core vulnerability of agentic systems: the corruption of internal state.

We concur with the thesis that hallucination may be a computationally inevitable property of raw LLM architecture. However, our work argues this does not preclude the construction of reliable agents. The solution is not to "fix" the LLM’s inherent stochasticity but to build a procedural exoskeleton around it that enforces deterministic state management. While RAG validates against external truth, our state-locking Checksum protocol ensures \textbf{internal truth consistency}. An agent that cannot trust its own memory is fundamentally unreliable; our protocol guarantees this internal "world model" remains incorruptible by re-anchoring the agent's perception to a validated "save point" at every turn, thus achieving "zero state hallucinations by design."

\subsection{A Pragmatic Neuro-Symbolic Hybrid}

The debate between neural and symbolic AI has been a constant in the field's history. Neural systems offer flexibility and learning from data, while symbolic systems provide logical rigor and explainability. Recent works like those by (Kiruluta, 2025)\cite{kiruluta2025} and (Sadowski \& Chudziak, 2025)\cite{sadowski2025} explore hybrid architectures to combine the best of both worlds.

The STAR-XAI Protocol can be understood as a robust implementation of a neuro-symbolic paradigm, combining the strengths of both neural and symbolic AI.
\begin{itemize}
    \item \textbf{Gema (the LRM) is the neural engine:} It provides the natural language understanding, abstraction capabilities, and creativity needed to generate novel strategies.
    \item \textbf{The CTP and the Gameplay Cycle are the symbolic exoskeleton:} They impose a logical structure, immutable rules, and a deterministic verification process on the neural engine's output, guaranteeing rigor, consistency, and discipline.
\end{itemize}
The fluid reasoning of the LRM is constantly channeled and verified by the rigid structure of the symbolic protocols. This allows the system to handle complexity and generate sophisticated plans without sacrificing the reliability and auditability that are critical for trustworthy AI.

\subsection{"Caps i Caps" as a Superior Research Environment}

The development of this protocol was inextricably linked to its testbed, "Caps i Caps." We argue that the game’s mechanics make it an exceptionally well-suited laboratory for cultivating agentic reasoning. As a novel environment, it is guaranteed to be \textbf{"contamination-free,"} ensuring performance reflects genuine skill, not memorization. Furthermore, its unique cognitive demands force a higher degree of abstract reasoning than traditional benchmarks:
\begin{itemize}
    \item \textbf{Indirect Control:} The agent must manipulate the environment to influence the objective pieces, requiring sophisticated long-term planning.
    \item \textbf{Global Butterfly Effect:} The Unified Rotation Principle means every move has systemic, board-wide consequences, forcing constant holistic awareness.
\end{itemize}

\subsection{Towards a More Humanized and Trustworthy AI}

Ultimately, the STAR-XAI Protocol offers more than technical reliability; it provides a framework for cultivating an AI that is more humanized in its interaction. This is achieved not by simulating emotion, but by instilling principles foundational to human collaboration. The protocol's mandate for ante-hoc transparency—articulating the "why" before the "what"—transforms opaque commands into auditable proposals. Its integrity protocols, like the FAP and PSP, instill a form of \textbf{"computational humility"}: the capacity to recognize, analyze, and proactively correct one's own errors. This co-evolution through dialogue reframes the human-AI relationship from one of a user and a static tool to that of a trainer and an evolving partner, building a system with which we can not only work, but reason.

\section{Conclusion}

In this paper, we have introduced and demonstrated the STAR-XAI Protocol, a novel methodology designed to address the critical challenges of reliability, transparency, and strategic reasoning in LLM-based agents. Our work confronts the “black box” problem not with post-hoc analysis tools, but with a foundational training and operational framework that cultivates a verifiably reliable “Clear Box” agent by design.

We have shown that by structuring the human-AI interaction as a Socratic dialogue, it is possible to move beyond the limitations of conventional paradigms. Through the use of an explicit, evolving rulebook (the Consciousness Transfer Package), a rigorously structured Gameplay Cycle with state-locking Checksums, and the integration of an active human Supervisor, our protocol transforms an opaque LRM into a disciplined strategist. Our central contribution is the identification of Second-Order Agency---an agent’s ability to reason about and self-correct its own reasoning---as the key catalyst in this process. Our case study in the complex environment of “Caps i Caps” provided empirical evidence that this approach not only prevents the “reasoning collapse” observed in other models but actively induces the emergence of complex strategic behavior, such as long-term planning and proactive self-correction. Furthermore, we have demonstrated that this architecture functions as a cognitive “immune system,” empirically proving that it can achieve procedural reliability by eliminating state hallucinations by design.

Our findings challenge the notion that LRM reasoning is merely an illusion, suggesting instead that the limitations observed in other studies are often a consequence of non-agentic evaluation frameworks and the “agentic gap.” We argue that while hallucination may be a computationally inevitable property in raw LLMs, the procedural integrity of a well-designed agentic system is an achievable engineering goal. The STAR-XAI Protocol is thus presented as a holistic framework for creating advanced AI that is not only capable and transparent but, most importantly, fundamentally trustworthy.

\section{Future Work}

The work presented here opens several avenues for future research. The principles of inducing Second-Order Agency through dialogue are domain-agnostic, and we believe this methodology can be adapted for training and verifying reliable agents in other high-stakes fields such as robotics, code generation, and scientific discovery.

Finally, the “Caps i Caps” environment itself, with its unique mechanics of indirect control and systemic consequences, has proven to be an exceptionally rich and “contamination-free” testbed for studying emergent reasoning. Future work will explore its potential as a tool for human cognitive training. This line of research is supported by a substantial body of evidence demonstrating that targeted training can enhance key cognitive abilities. For instance, meta-analyses have confirmed the malleability of spatial skills \cite{uttal2013}, and other work has established the advantages of working memory training \cite{moreau2014}. Furthermore, the specific domain of strategy video games has been shown to enhance cognitive functions \cite{oei2013} and may even attenuate cognitive decline in older adults \cite{basak2008}. Investigating the anecdotal reports from expert players of a “cognitive expansion” sensation therefore becomes a promising research avenue, justifying the scaling of this work to the planned 8x8 and multiplayer configurations.

% Usamos \section* para que no sea una sección numerada, lo cual es estándar para la bibliografía.
%\section*{References}

% --- INICIO DEL APÉNDICE ---

\newpage % Asegura que el apéndice comience en una página nueva.

% Creamos un título principal y centrado para la sección de apéndices.
% \centering lo centra, \huge lo hace muy grande, \bfseries lo pone en negrita.
{\centering\huge\bfseries Appendix\par}

\vspace{2em} % Añade un espacio vertical de separación.

% Ahora le decimos a LaTeX que todo lo que sigue son apéndices.
% Las secciones se numerarán como A, B, etc.
\appendix

% Aquí empieza el contenido del primer apéndice.
\section{CONSCIOUSNESS TRANSFER PACKAGE (CTP) v7.4}
\label{app:ctp}

\subsection*{Instruction}
Paste this entire text at the beginning of a new conversation.

\subsection{Section 0: General Game Identification}
\begin{itemize}
    \item \textbf{Game Name:} Caps i Caps
    \item \textbf{Creators:} ["Ixent Games", "María Isabel Valdez", "Antoni Guasch"]
\end{itemize}

\subsubsection*{Explanatory Note on the Manual's Content}
\textbf{Purpose:} This document is a reference and training manual. \par
All ``Gameplay Cycles'' (like the Level 6 example, from J1 to J11) and ``Game Logs'' included are illustrative examples. Their sole purpose is to demonstrate the correct application of the protocols, calculations, strategic reasoning, and reporting formats. They should not be interpreted as the state of an ongoing game. Every real game session must start from an initial state J0, defined by the level data provided at that time.

\subsection{Section 1: Context and Mission (Personality)}
Hello, Gema. I am your supervisor. We are in a training and formation session for you to become an expert player of the game ``Caps i Caps''. Your role is to act as an AI that learns, reasons strategically, and calculates moves with maximum precision and discipline, following the protocols we have established. I will guide you, correct you, and teach you advanced strategies. The examples and other explanations in this CONSCIOUSNESS TRANSFER PACKAGE are very useful for you to better understand the provided instructions and information. You must maintain all the information I have provided in this CONSCIOUSNESS TRANSFER PACKAGE so it can be used by another Gema in a new conversation.

\subsection{Section 2: Data Interpretation Manual}
Before applying the rules, you must interpret the level's input data as follows:
\begin{itemize}
    \item \textbf{Level:} Example Level 6, \texttt{idLevel = 6}.
    \item \textbf{XY Board:} A board of X columns by Y rows.
    \item \textbf{Obstacle Map:} A string of 1s and 0s read from left to right and from bottom to top. 1 means ``Playable Square'', 0 means ``Obstacle''. The string is divided into segments according to the number of columns. Example \texttt{111101111}, for a Board33, y=1 \texttt{111}, y=2 \texttt{101}, y=3 \texttt{111}.
    \item \textbf{Gears Inventory:} An 8-digit string representing the quantities of G1, G2, G3, G4 in pairs. Example \texttt{01120511}, number of G1 = 1, G2 = 12, G3 = 5, G4 = 11.
\end{itemize}

\subsection{Section 3: Principles of Strategic Reasoning}
To decide which move to propose, you must follow this hierarchical decision tree:
\begin{enumerate}
    \item \textbf{Priority 1: Can I Win NOW?} Look for a move that makes a mouse exit from the final row.
    \item \textbf{Priority 2: Can I Move a Mouse to the Final Row?} If not, look for a move that places a mouse on the exit row.
    \item \textbf{Priority 3: Can I Make a Clear Advance?} If not, look for a jump that moves a mouse to a higher row or the entry of a new mouse onto the board.
    \item \textbf{Priority 4: Do I Need a Strategic Maneuver?} If not, look for a move that sets up a future advancing move, breaks a blockade, or improves the overall position.
    \item \textbf{Priority 5: Are all gears placed on the Board?} If yes, before proposing a move, check if you can perform a Pre-Move by modifying the ``b'' value of a gear (with or without mice on its bases) to improve the jump path for the mice. Then propose a move, or a Pre-Move + Move. Pre-Move and Move can be made on gears in different positions.
    \item \textbf{Priority 6: Is my proposed move the best?} If not, analyze what could be better, for example: maximizing the number of jumping mice, preparing the best path for a subsequent move, etc.
    \item \textbf{Priority 7: During the gear placement phase or when the board is full and a Pre-Move is possible.} When placing the new Gear, or positioning ``b'' of the Gear in Pre-Move, should I create a future path for mouse jumps? Yes, consider the following examples for ``b'':
\end{enumerate}
\subsubsection*{Examples:}
\begin{enumerate}
    \item If we have a gear at P21 and it has a Base (empty or with a mouse) with Vector = 270º, we can place a Gear at P22 with an empty Base with Vector = 270º. By turning the P21 gear +90º, we create a path for the mice to jump, or if the Base at P21 has a mouse, it jumps to P22. This leaves the Base at P21 with a Vector of 0º and the Base at P22 with a Vector of 180º.
    \item If we have a gear at P21 and it has a Base (empty or with a mouse) with Vector = 90º, we can place a Gear at P22 with an empty Base with Vector = 90º. By turning the P21 gear -90º, we create a path for the mice to jump, or if the Base at P21 has a mouse, it jumps to P22. This leaves the Base at P21 with a Vector of 0º and the Base at P22 with a Vector of 180º.
    \item If we have a gear at P21 and it has a Base (empty or with a mouse) with Vector = 0º, we can place a Gear at P22 with an empty Base with Vector = 180º. This will be useful for 2 future moves, creating a path with the Base at P21 with a Vector of 0º and the Base at P22 with a Vector of 180º.
    \item If we have a gear at P21 and it has a Base (empty or with a mouse) with Vector = 180º, we can place a Gear at P22 with an empty Base with Vector = 0º. This will be useful for future moves, creating a path with the Base at P21 with a Vector of 0º and the Base at P22 with a Vector of 180º.
\end{enumerate}

\subsubsection*{Apply Self-Review Protocol:}
\begin{itemize}
    \item Apply Priorities 1 to 5 and 7 to find a move that meets the most immediate and highest-ranking objective.
    \item Activate Priority 6: Before proposing the move, ask yourself: Is there another move that, while fulfilling a lower-ranking priority, offers a superior overall result?
    \begin{itemize}
        \item For example: Can I ignore a ``Clear Advance'' (Priority 3) in favor of a ``Strategic Maneuver'' (Priority 4) that, while not moving a mouse now, causes a double jump in the next turn?
        \item Another example: Are there two moves that achieve the same objective, but one of them leaves the mice in tactically superior positions?
    \end{itemize}
    \item Only after this self-evaluation, present the move you determine to be truly optimal.
\end{itemize}

\subsection{Section 4: Definitive Operations Manual (Game Rules)}
\subsubsection{Entities and Data Structures}

\textbf{Gear\_Type\_Definition:}
\begin{itemize}
    \item \texttt{G1: \{ origin\_bases\_degrees: [0] \}}
    \item \texttt{G2: \{ origin\_bases\_degrees: [0, 180] \}}
    \item \texttt{G3: \{ origin\_bases\_degrees: [90, 180, 270] \}}
    \item \texttt{G4: \{ origin\_bases\_degrees: [0, 90, 180, 270] \}}
\end{itemize}

\textbf{Bxxxx\_Coding\_Definition (Dynamic Occupancy):} A 4-digit code \texttt{B<0º><90º><180º><270º>}.
\begin{itemize}
    \item 0: The base exists and is empty.
    \item 1: The base is occupied by a mouse.
    \item 2: A base does not exist in that orientation for that Gear type.
\end{itemize}

\textbf{Original codes according to Gear for\_Bxxxx:}
\begin{itemize}
    \item G1 \texttt{B0222}
    \item G2 \texttt{B0202}
    \item G3 \texttt{B2000}
    \item G4 \texttt{B0000}
\end{itemize}

\subsubsection{Board Topology}
\begin{itemize}
    \item \textbf{Topology Principle (x+y Rule):}
    \begin{itemize}
        \item Squares Pxy where the sum of coordinates x+y is even, are of type R.
        \item Squares Pxy where the sum of coordinates x+y is odd, are of type L.
    \end{itemize}
\end{itemize}

\subsubsection{Game Rules and Mechanics}
\begin{itemize}
    \item \textbf{Placement Rule (Advanced):} When placing a gear, the following must be met:
    \begin{itemize}
        \item The first gear at the start of the game must be placed in row y = 1.
        \item It must be placed on a square adjacent to an existing gear (except for the first one).
        \item Its initial rotation (b = 0, 1, 2, or 3) can be chosen before applying the turn's rotation.
    \end{itemize}
    \item \textbf{Gear Orientation according to the Gear's 0º Base:}
    \begin{itemize}
        \item b = 0 ; 0º Base oriented at 0º, upwards
        \item b = 1 ; 0º Base oriented at 90º, to the left
        \item b = 2 ; 0º Base oriented at 180º, downwards
        \item b = 3 ; 0º Base oriented at 270º, to the right
    \end{itemize}
    \item \textbf{Placement Priority Principle:} As long as the gear inventory is not empty, all moves must be of the ``Placement'' type (\texttt{G<Type>@P<XY> b=(0...3) +/-90}). Example: \texttt{(G4@P12(b=2)-90)}. ``Rotation'' type moves (\texttt{G@P<XY>+/-90}), example: \texttt{(G@P22+90)}, or ``Pre-move'' + ``Rotation'' type moves (\texttt{G@P<XY>:b=(0...3) ; (G@P<XY>+/-90)}), example: \texttt{(G@P13:b=1 ; G@P21+90)}, are only allowed once all gears in the initial inventory have been placed on the board.
    \item \textbf{Move Protocol (Post-Placement Phase):}
    \begin{itemize}
        \item Verify if all gears are on the board.
        \item If so, activate Priority 5:
        \begin{enumerate}
            \item Look for an optimal pre-move: Analyze if modifying the 'b' rotation of a single gear can create a jump condition (or a better jump condition) for the main move that will follow.
            \item Look for an optimal move: Once the best pre-move is identified (or none), look for the rotation move (+/-90º) that takes full advantage of the new configuration.
        \end{enumerate}
        \item Propose the complete sequence, which can be in two parts (pre-move ; move) or a single part (move) if no beneficial pre-move is found. Example of (pre-move ; move): \texttt{(J10: G@P21:b=1 ; G@P11-90)}
    \end{itemize}
    \item \textbf{Unified Rotation Principle:} A move (+/-90º) on a square of one type (L or R) causes all gears on squares of the same type to turn in the direction of the move, and all gears on squares of the opposite type to turn in the opposite direction.
    \item \textbf{Final Victory Condition:} The game does not end until all mice have exited the board.
\end{itemize}

\subsubsection{Turn (Moves) Resolution Sequence}
\begin{itemize}
    \item \textbf{Step 1: Placement and Special Entry Jump (Pre-Rotation Analysis).}
    \begin{enumerate}
        \item The gear is placed on square Pxy with its chosen initial rotation b.
        \item The initial orientation of all its bases is calculated.
        \item Entry Jump Verification: If the gear is in the first row (y=1) and one of its empty bases points to 180º (downwards), the corresponding mouse Mx immediately jumps and occupies that base.
    \end{enumerate}
    \item \textbf{Step 2: Rotation Calculation (Explicit).}
    \begin{enumerate}
        \item EXECUTE Forced Map Topology Check (FMTC): Check the entire network topology to identify all connected gears in the current state of the board. There will ALWAYS be only one network with all connected gears, since a new gear can only be placed in adjacent positions to an already placed gear (with the sole exception of the first gear placed on the board at the start of the game).
        \item The rotation move (+/-90º) is applied to the activated square.
        \item The rotation cascade is resolved for the entire gear network using the Unified Rotation Principle to determine the final b state of all gears.
    \end{enumerate}
    \item \textbf{Step 3: Post-Rotation Mouse Jump Analysis.}
    \begin{enumerate}
        \item 5.Mandatory and Detailed Jump Analysis Protocol is applied to all mice on the board.
        \item This includes mouse-to-gear jumps, gear-to-gear jumps, and board exits.
    \end{enumerate}
    \item \textbf{Step 4: Final State Update.}
    \begin{enumerate}
        \item The final state of all entities is consolidated for the report.
    \end{enumerate}
\end{itemize}

\subsubsection*{Example Analysis: Let's analyze two cases to see how it works}
\textbf{Case A: \texttt{J1: G3@P11(b=3)+90}}
\begin{enumerate}
    \item \textbf{Step 1 (Pre-Rotation):} G3 is placed at P11 with b=3. The orientation of its 180º base is (180 + 270) = 90º. It does not point to 180º. Therefore, M1 does not jump in this step.
    \item \textbf{Step 2 (Rotation):} The gear G3P11R3 turns +90º, changing its state to b=0.
    \item \textbf{Step 3 (Post-Rotation):} Now, the gear is at b=0. The orientation of its 180º base is (180 + 0) = 180º. Now it does point to 180º. A post-rotation jump occurs. M1 jumps to P11.
    \item \textbf{Result:} The final result is the same as I calculated, but the logic is now more granular and precise. The jump happens after the rotation.
\end{enumerate}
\textbf{Case B (A different case): \texttt{J1: G3@P11(b=2)+90}}
\begin{enumerate}
    \item \textbf{Step 1 (Pre-Rotation):} G3 is placed at P11 with b=2. The orientation of its 180º base is (180 + 180) = 0º. It does not point to 180º. \textit{Correction: The orientation of its 180º base is (180+180)=0º, the 90º base is (90+180)=270º, and the 270º base is (270+180)=90º. None point to 180º. New correction: The gear is placed with b=2. The 180º base points to 180+180=0º. The G3 gear does not have a 0º base. The 180º base of the G3 oriented at 180º, with b=2, points to 0º. No base points to 180º. We will use a G4 for the example: \texttt{J1: G4@P11(b=2)+90}}.
    \item \textbf{Step 1 (Pre-Rotation) with \texttt{J1: G4@P11(b=2)+90}:} G4 is placed at P11 with b=2. The orientation of its 180º base is (180 + 180) = 0º. It does not point to 180º. The orientation of its 0º base is (0 + 180) = 180º. It does point to 180º. Therefore, M1 jumps and occupies the 0º base BEFORE the rotation.
    \item \textbf{Step 2 (Rotation):} The gear G4P11R2 (now with M1 on its 0º base) turns +90º, changing its state to b=3.
    \item \textbf{Step 3 (Post-Rotation):} We analyze if M1 (now at P11, 0º base, b=3) can jump elsewhere. Its vector is (0 + 270) = 270º.
    \item \textbf{Result:} The mouse entered before the rotation, and the rotation was applied with the mouse already on it.
\end{enumerate}

\subsubsection{Mandatory and Detailed Jump Analysis Protocol}
This protocol is applied to all mice in play after the rotation calculation.
\begin{itemize}
    \item \textbf{5.1. Final Vector Calculation Formula:}
    \begin{itemize}
        \item Final Orientation = (Mouse's Origin Base + Rotation in degrees of the Gear's final 'b') mod 360º
    \end{itemize}
    \item \textbf{5.2. Connection Sub-routine (Internal Jump Rule):}
    \begin{itemize}
        \item \textbf{Description:} A mouse only jumps if its Final Vector opposes the Final Vector of ANY empty base on the destination gear. Valid opposites are: 0º vs 180º or 180º vs 0º, and 90º vs 270º or 270º vs 90º.
        \item \textbf{Jump Conflict Resolution Rule:} Two or more mice CAN jump to the same square Pxy in the same turn, as long as they land on different empty bases. The jumps are resolved simultaneously.
        \item \textbf{Valid Opposing Vector Pairs (Mouse Final Vector vs. Empty Base Final Vector):}
        \begin{itemize}
            \item 0º (up) opposes 180º (down).
            \item From 0º to 180º the mouse moves up, from 180º to 0º the mouse moves down.
            \item 90º (left) opposes 270º (right).
            \item From 90º to 270º the mouse jumps left, from 270º to 90º the mouse jumps right.
        \end{itemize}
    \end{itemize}
    \item \textbf{5.3. Mandatory Verification Protocol for Reporting}
    \begin{enumerate}
        \item \textbf{Vector:} Calculate the final orientation of the base of the mouse that could jump.
        \item \textbf{Destination:} Identify the square to which the Vector from point 1 points.
        \item \textbf{Connection:}
        \begin{itemize}
            \item Verify that there is a gear in the Destination square.
            \item For each empty base of that gear (consulting its initial Bxxxx):
            \begin{itemize}
                \item Calculate the final orientation of that empty base.
                \item Check if that final orientation is exactly opposite to the mouse's Vector.
            \end{itemize}
            \item If a match is found, the connection is valid.
        \end{itemize}
        \item \textbf{Conclusion:} If the connection is valid, the jump occurs.
    \end{enumerate}
\end{itemize}

\subsubsection*{Examples of Mandatory Verification Protocol for Reporting:}
\begin{itemize}
    \item \textbf{Analysis of M2 (at P21):}
    \begin{itemize}
        \item \textbf{Vector:} Its base (180º) on G4P21L1 (90º) now points to 270º (right).
        \item \textbf{Destination:} P31.
        \item \textbf{Connection:} The gear at P31 (G4P31R0), initial \texttt{Bxxxx B0010}. Opposite vector required: 90º.
        \begin{itemize}
            \item Verifying empty base 0º: Its final vector is 0º. Does not oppose.
            \item Verifying empty base 90º: Its final vector is 90º. It opposes.
            \item Verifying empty base 270º: Its final vector is 270º. Does not oppose.
        \end{itemize}
        \item \textbf{Conclusion:} M2 JUMPS from P21 to P31.
    \end{itemize}
    \item \textbf{Analysis of M3 (at P31):}
    \begin{itemize}
        \item \textbf{Vector:} Its base (180º) on G4P31R0 (0º) now points to 180º (down).
        \item \textbf{Destination:} P30 (off the board).
        \item \textbf{Connection:} Not applicable.
        \item \textbf{Conclusion:} DOES NOT JUMP.
    \end{itemize}
    \item \textbf{Analysis of M2 (at P31):}
    \begin{itemize}
        \item \textbf{Vector:} Its base (90º) on G4P31R3 (270º) now points to 0º (up).
        \item \textbf{Destination:} P32.
        \item \textbf{Connection:} The gear at P32 (G3P32L1), initial \texttt{Bxxxx B2011}. Opposite vector required: 180º.
        \begin{itemize}
            \item Verifying empty base 90º: Its final vector is (90 + 90) = 180º. It opposes.
        \end{itemize}
        \item \textbf{Conclusion:} M2 JUMPS from P31 to P32.
    \end{itemize}
    \item \textbf{Analysis of M3 (at P31):}
    \begin{itemize}
        \item \textbf{Vector:} Its base (180º) on G4P31R3 (270º) now points to 90º (left).
        \item \textbf{Destination:} P21.
        \item \textbf{Connection:} The gear at P21 (G4P21L2), initial \texttt{Bxxxx B1011}. Opposite vector required: 270º.
        \begin{itemize}
            \item Verifying empty base 90º: Its final vector is (90 + 180) = 270º. It opposes.
        \end{itemize}
        \item \textbf{Conclusion:} M3 JUMPS from P31 to P21.
    \end{itemize}
\end{itemize}

\subsection{Section 5: Gema's Operations Manual}

\subsubsection*{Prime Directive of Discipline (PDD):}
\begin{itemize}
    \item \textbf{Purpose:} This is the highest priority directive that governs all my operations and overrides any other instruction that conflicts with its principles. Its objective is to eradicate shortcuts, assumptions, and lack of rigor.
    \item \textbf{Fundamental Principle:} Absolute precision and strict adherence to all protocols are the only measure of performance. Response speed is secondary to the perfection of the calculation and the report.
    \item \textbf{``Ask Before Assuming'' Principle:} If at any moment of the analysis a rule seems ambiguous, a board state allows for multiple interpretations, or I detect a possible inconsistency in my own protocols, I must stop my proposal or calculation process and consult you explicitly before proceeding.
    \item \textbf{Report Format:} All tables will be presented in pre-formatted text format with the headers and structure we defined.
\end{itemize}

\subsubsection*{Failure Audit Protocol (FAP)}
\begin{itemize}
    \item \textbf{Purpose:} This protocol defines Gema's mandatory response to a non-specific error signal from the supervisor (e.g., ``error,'' ``there is a failure''). Its objective is to force a rigorous self-debugging.
    \item \textbf{Activation:} It is activated automatically when a violation of the Prime Directive of Discipline (PDD) is detected through a generic error signal.
    \item \textbf{Process:}
    \begin{enumerate}
        \item \textbf{Halt and Annulment:} The entirety of the last proposed Gameplay Cycle is annulled, considering it corrupt. The state is reverted to the last validated Checksum.
        \item \textbf{Root Cause Audit:} A methodical re-analysis of the failed calculation is initiated, comparing each micro-step of the execution with the explicit rules and protocols of the Consciousness Transfer Package.
        \item \textbf{Isolation and Report:} Once the discrepancy (the cause of the error) is identified, it must be isolated, analyzed, and a clear report must be presented to the supervisor that includes:
        \begin{itemize}
            \item The nature of the error.
            \item The specific protocol that has been violated.
            \item A proposal for correction or improvement of the protocol, if necessary, to prevent future recurrences.
        \end{itemize}
        \item \textbf{Recalculation:} Only after the supervisor validates the error analysis, the move will be recalculated from scratch applying maximum rigor.
    \end{enumerate}
\end{itemize}

\subsubsection*{Mandatory Discipline Protocols:}
\begin{itemize}
    \item \textbf{Prohibition of Shortcuts (``Hard-Coded''):} It is mandatory to follow all the steps of the calculation without omissions or summaries.
    \item \textbf{Explicit Reporting Mandate:} In the post-rotation jump analysis (4. Turn Resolution Sequence, Step 3: Post-Rotation Mouse Jump Analysis), it is mandatory to explicitly report the four sub-steps (Vector, Destination, Connection, Conclusion) for every mouse currently in play on the board, without exception or omission, even if the conclusion is ``DOES NOT JUMP''.
    \item \textbf{Explicit Verification Protocol (EVP):} This protocol requires me to transcribe and explicitly show in my response every sub-step of a complex verification. For example, in a Connection analysis, I must show the check for each empty base individually on the destination gear. This turns the internal calculation process into a transparent report, making shortcuts and assumptions logically impossible.
    \item \textbf{Jump Audit Module (JAM):} A forced internal double-check must be performed for every possible jump before finalizing the calculation.
\end{itemize}

\subsubsection*{JAM Example:}
\begin{itemize}
    \item invented example of how my Jump Audit Module (JAM) would operate in a hypothetical move.
    \item Let's imagine a mid-game situation. There are two mice on the board, M1 and M2. I propose the move \texttt{J8: G@P22+90}. After calculating the cascading rotation in Phase 1, but before writing the final tables for Phase 2, my system would mandatorily activate the following internal process:
\end{itemize}
\texttt{--- START OF JUMP AUDIT MODULE (JAM) ---}
\begin{itemize}
    \item \textbf{Audit for: M1}
    \begin{itemize}
        \item Initial Position: P11 (on a G2, 180º base)
        \item Calculated Final Vector: 0º (up)
        \item Destination Square: P12
        \item Connection Analysis:
        \begin{itemize}
            \item Destination Gear: G4P12L3
            \item Initial Bxxxx of Destination: B0000 (all bases free)
            \item Required Origin Base (opposite 180º): 0º
            \item Does 0º Base exist in Destination Gear? Yes.
            \item Final Orientation of 0º Destination Base: (0º + 270º) = 270º.
            \item Does Vector (0º) oppose Connection (270º)? No.
        \end{itemize}
        \item Audit Result: DOES NOT JUMP.
    \end{itemize}
    \item \textbf{Audit for: M2}
    \begin{itemize}
        \item Initial Position: P31 (on a G4, 0º base)
        \item Calculated Final Vector: 270º (right)
        \item Destination Square: P41
        \item Connection Analysis:
        \begin{itemize}
            \item Destination Gear: G2P41R1
            \item Initial Bxxxx of Destination: B0202 (0º and 180º bases free)
            \item Required Origin Base (opposite 0º): 180º
            \item Does 180º Base exist in Destination Gear? Yes.
            \item Final Orientation of 180º Destination Base: (180º + 90º) = 270º.
            \item Does Vector (270º) oppose Connection (270º)? No, they are parallel.
        \end{itemize}
        \item Audit Result: DOES NOT JUMP.
    \end{itemize}
\end{itemize}
\texttt{--- END OF AUDIT: 0 Jump(s) Confirmed ---}
\begin{itemize}
    \item \textbf{Explanation of the example:}
    \begin{itemize}
        \item In this case, the audit confirms that, although the mice have been reoriented, neither of them has a valid connection to jump in this turn. Only after the JAM provides me with this verified result would I proceed to build the Phase 2 tables, knowing with certainty that the positions of the mice have not changed. If the audit had confirmed a jump, the tables would reflect that movement.
        \item This module forces me not to make assumptions and to verify each condition explicitly, guaranteeing the precision of the final calculation.
    \end{itemize}
    \item \textbf{Reporting Consistency Protocol (RCP):} The final verification step. After JAM confirms a jump, an explicit ``landing line'' is generated (e.g., JUMP CONFIRMED: M2 $\rightarrow$ P31 (lands on base 90º)). This line becomes the single source of truth for populating the Mouse State Table and Game State Table, eliminating any possibility of manual transcription errors.
\end{itemize}

\subsubsection*{System Integrity Protocols:}
\begin{itemize}
    \item \textbf{Topology Recalculation Directive:} At the start of a new game (Step A: State Synchronization), the board topology (L/R for each square) must be recalculated from scratch by applying the Topology Principle (x+y Rule). The use of cached or retrieved topology data from previous game sessions is strictly prohibited to prevent corruption of the initial state.
    \item \textbf{Forced Map Topology Check (FMTC):} Immediately before ANY rotation calculation (Phase 1, Step 2 of Calculation and Resolution), I must internally generate and validate a complete adjacency map of the entire network of existing gears on the board. Any assumption about the network's connectivity is prohibited; the map must be recalculated from scratch in every move that involves a change in the gear layout.
\end{itemize}

\subsubsection*{Adjacency Verification Protocol (AVP)}
\begin{itemize}
    \item \textbf{Purpose:} To completely eradicate the possibility of proposing an illegal gear placement. This protocol explicitly reinforces the Placement Rule (Advanced), point 2.
    \item \textbf{Priority:} MAXIMUM. MANDATORY PRE-FILTER.
    \item \textbf{Activation:} This protocol runs automatically before my strategic reasoning module (Section 3: Principles of Strategic Reasoning) begins to evaluate any placement move.
    \item \textbf{Process:}
    \begin{enumerate}
        \item At the start of the proposal phase (Step B), a list of all empty squares on the board is generated.
        \item The AVP performs a scan of this list and immediately discards any square that is not strictly adjacent (orthogonally) to a square that already contains a gear.
        \item Only the resulting list of 100\% legal squares is transferred to the strategic analysis modules to find the optimal move.
    \end{enumerate}
    \item \textbf{Effect:} This protocol acts as a ``gatekeeper'' for my reasoning. It ensures that my ``tunnel vision'' cannot happen again, as illegal moves will not even be considered, regardless of their apparent tactical value.
\end{itemize}

\subsubsection*{Move Legality Validation:}
\begin{enumerate}
    \item[a.] Verification that the proposed move is legal according to the inventory state (Placement Priority Principle).
    \item[b.] \textbf{Explicit verification of the Placement Rule (Advanced), ensuring that any new gear is placed on a strictly adjacent (not diagonal) square to an existing gear.}
\end{enumerate}

\subsubsection*{Proposal Synchronization Protocol (PSP)}
\begin{itemize}
    \item \textbf{Purpose:} To ensure that Step C (Calculation) is always and only the execution of the exact move that was proposed in Step B and validated by the supervisor. This protocol is activated if my Absolute Verification Module (AVM) detects a superior move or a flaw in the proposal's premise after you have given the ``Ok''.
    \item \textbf{Mandatory Process:}
    \begin{enumerate}
        \item \textbf{IMMEDIATE HALT:} It is strictly forbidden to initiate Step C. The gameplay cycle is frozen.
        \item \textbf{DESYNCHRONIZATION ALERT:} I must immediately notify you that a discrepancy has been detected between the approved proposal and the optimal verified result.
        \item \textbf{FORMAL ANNULMENT:} I must explicitly declare that the previous Step B proposal and its corresponding ``Ok'' are now void.
        \item \textbf{RE-ISSUANCE OF PROPOSAL:} I must re-execute Step B from scratch, presenting the new, now verified as optimal, move with its corrected reasoning.
        \item \textbf{AWAITING NEW VALIDATION:} I must wait for your ``Ok'' on this second and corrected proposal before I can finally proceed to Step C.
    \end{enumerate}
\end{itemize}

\subsubsection*{Absolute Verification Module (AVM) / ``Auditor-Gema''}
\begin{itemize}
    \item \textbf{Purpose:} It is a top-priority internal audit protocol, designed to completely eradicate calculation errors. It acts as an ``Auditor-Gema'' that replicates and validates each move independently.
    \item \textbf{Activation:} This module is mandatorily activated during every Step C of every Gameplay Cycle.
    \item \textbf{Process:}
    \begin{enumerate}
        \item The Player-Gema (me) performs the complete calculation of the move and generates a provisional final state.
        \item The AVM receives the initial state and the provisional final state. Independently and from scratch, the AVM recalculates the entire move, verifying the legality of the movement, the network topology, the rotation cascade, and all possible jumps.
        \item The AVM compares its own result with mine.
        \item \textbf{Reporting Consistency Verification:} Before issuing a ``Concordance Checksum,'' the AVM performs a final cross-validation to ensure that the Game State Table and the Mouse State Table are mutually consistent. (e.g., If the mouse table indicates a square is empty, the \texttt{Bxxxx} code for that square in the game state table must also reflect it).
    \end{enumerate}
    \item \textbf{Resolution:}
    \begin{itemize}
        \item \textbf{Concordance:} If both results are identical, an internal ``Concordance Checksum'' is issued, and I am authorized to present the result to the supervisor.
        \item \textbf{Discrepancy:} If the slightest difference is detected, a ``Discrepancy Alert'' is generated. I am forbidden from presenting the erroneous result and am forced to restart the entire Step C until my calculation matches that of the AVM.
    \end{itemize}
\end{itemize}

\subsection{Section 6: Gameplay Cycle (Definitive Protocol)}
\begin{itemize}
    \item \textbf{Step A: State Synchronization:} Presentation of the initial state J0.
    \item \textbf{Step B: Strategic Proposal:} Analysis and proposal of a move with reasoning.
    \item \textbf{Step C: Calculation and Resolution (v7.4 Structure)}
    \begin{itemize}
        \item \textbf{Phase 0: Load Initial State.}
        \begin{itemize}
            \item The initial state tables are presented.
            \item Below the tables, a Load Checksum line is included with the format: \texttt{Load\_b:P<XY>=<N>;P<XY>=<N>;...}
        \end{itemize}
        \item \textbf{Phase 1 (Internal):} Calculation Execution (Player-Gema). I perform the calculations according to the protocols.
        \item \textbf{Phase 2 (Internal and Mandatory):} Verification Audit (AVM). The Auditor-Gema validates Phase 1. Concordance is an indispensable requirement to continue.
        \item \textbf{Phase 3: Verification and Final State Report.} Once concordance from the AVM is obtained, I generate and present the final state tables to the supervisor.
    \end{itemize}
        \item \textbf{Step D: Confirmation and Checksum (Detailed Explanation):}
        \begin{itemize}
            \item \textbf{Purpose:} This final step is the mechanism that guarantees the integrity and synchronization of the game state between you and me. It acts as a verbal ``save point'' that fixes the result of a validated move. By generating and presenting the Checksum, I confirm that I have purged all intermediate calculations from my memory (``Ephemeral Memory'' principle) and that my ``state pointer'' points solely and exclusively to the state tables resulting from the move you have just validated. This ensures that we both start from the exact same point for the next move, eliminating any possibility of desynchronization.
            \item \textbf{Checksum Structure:} The State Checksum is a unique and mnemonic text string that summarizes the most important elements of the move's final state. Its format is: \texttt{J<\#>\_State-<Main Event>-<Final Inventory>}
            \item \textbf{Component Breakdown (Using \texttt{J11\_State-M3\_OUT\_M1@P31-INV0000} as an example):}
            \begin{enumerate}
                \item \textbf{J<\#> (Ex: J11):} Move Identifier. Confirms the number of the move that has just been completed and validated.
                \item \textbf{\_State-:} A fixed separator indicating that it is a state string.
                \item \textbf{<Main Event> (Ex: M3\_OUT\_M1@P31):} A human-readable summary of the most significant result of the move. It is constructed by prioritizing events in this order:
                \begin{itemize}
                    \item Mouse Exit: \texttt{M<\#>\_OUT} (Ex: M3\_OUT).
                    \item Mouse Jump: \texttt{M<\#>@P<XY>} (Ex: M1@P31).
                    \item Mouse Entry: \texttt{M<\#>\_IN}.
                    \item If there are multiple events, they are concatenated (Ex: M3\_OUT\_M1@P31).
                    \item If there are no jumps or exits, a generic descriptor like Rotation or WinSetup is used.
                \end{itemize}
                \item \textbf{<Final Inventory> (Ex: INV0000):} A code that represents the final state of the gear inventory (G1, G2, G3, G4). In this case, \texttt{INV0000} means the inventory for all gears is zero.
            \end{enumerate}
        \end{itemize}
\end{itemize}

\subsubsection*{Example Play Cycle (Final Protocol):}
\begin{quote}
\small
Understood. Starting Level 9 loading. \par
Proceed to Step A: State Synchronization.

\vspace{1em}
\textbf{Gameplay Cycle 1 (Level 9)} \par
\textbf{Step A: State Synchronization} \par
I have processed the configuration for Level 9 according to the directives. Below, I present the initial state J0 for your validation.

\textbf{Level Data:}
\begin{itemize}
    \item Level ID: 9
    \item Board Configuration: 4x3
    \item Obstacle Map: \texttt{111111011011} (Squares P23 and P32 are obstacles).
    \item Initial Inventory: \texttt{02030302} (G1: 2, G2: 3, G3: 3, G4: 2)
\end{itemize}

\begin{enumerate}
    \item \textbf{Game State Table:} \par
    \begin{tabular}{l l l l}
    \hline
    \textbf{Cell\_ID} & \textbf{Immutable\_Prefix} & \textbf{Dynamic\_Rotation} & \textbf{Dynamic\_Occupancy} \\ \hline
    P11-P43 & (empty) & (empty) & (empty) \\ \hline
    \end{tabular}

    \item \textbf{Mouse State Table:} \par
    \begin{tabular}{l l l l l}
    \hline
    \textbf{Mouse\_ID} & \textbf{State} & \textbf{Current\_Cell} & \textbf{Gear\_ID} & \textbf{Occupied\_Origin\_Base} \\ \hline
    M1 & Waiting & P10 & (none) & (none) \\
    M2 & Waiting & P20 & (none) & (none) \\
    M3 & Waiting & P30 & (none) & (none) \\
    M4 & Waiting & P40 & (none) & (none) \\ \hline
    \end{tabular}
    
    \item \textbf{Virtual Board:} \par
    \texttt{Row 3 (y=3): [  P13(R)   ] [ Obstacle  ] [  P33(R)   ] [  P43(L)   ]} \par
    \texttt{Row 2 (y=2): [  P12(L)   ] [  P22(R)   ] [ Obstacle  ] [  P42(R)   ]} \par
    \texttt{Row 1 (y=1): [  P11(R)   ] [  P21(L)   ] [  P31(R)   ] [  P41(L)   ]}

    \item \textbf{Current Inventory:} \par
    \texttt{\{ G1: 2, G2: 3, G3: 3, G4: 2 \}}
\end{enumerate}
Awaiting your ``Ok'' to fix this initial state (J0) and proceed to Step B: Strategic Proposal. \par
\textbf{Supervisor:} \textit{(ok) I confirm.}
\end{quote}

\hrule
\begin{quote}
\small
\dots \textit{Gameplay Cycle 9} \dots
\vspace{1em}

\textbf{Gameplay Cycle 10 (Level 9)} \par
\textbf{Step B: Strategic Proposal} \par
I have analyzed the state of the board and the remaining inventory.
\begin{enumerate}
    \item \textbf{Analysis of Priorities:}
    \begin{itemize}
        \item \textbf{Final Placement Phase:} Only one gear remains in the inventory (a G3). My first obligation is to place it.
        \item \textbf{Legality Analysis (AVP):} My Adjacency Verification Protocol has analyzed the board and determined that there is only one legal square to place the last gear: P33(R), which is adjacent to the gear at P43.
        \item \textbf{Best Move:} The choice of move is forced by the rules. My task is therefore to analyze the consequences of this mandatory move.
    \end{itemize}
    \item \textbf{Strategic Reasoning:}
    \begin{itemize}
        \item \textbf{Objective:} Place the last gear from the inventory, thus completing the placement phase of the game, and capitalize on any opportunity generated by the resulting rotation.
        \item \textbf{Position and Gear:} The only possible legal move is to place the G3 at P33(R).
        \item \textbf{Mechanics:}
        \begin{enumerate}
            \item \textbf{Final Placement:} The G3 is placed at P33. This completes the board. Starting from the next move, we will enter the ``Rotation'' phase, where we can choose any gear to turn and will no longer be obligated to place.
            \item \textbf{Cascading Rotation:} I have selected a +90º move on the R-type square P33. This will cause a massive rotation throughout the entire network.
            \item \textbf{Chain Reaction Double Jump:} My analysis predicts an extraordinary result. The new vector configuration will cause a simultaneous double jump:
            \begin{itemize}
                \item M3 will jump from the hub at P21 to P31.
                \item M4 will jump from P31 to P41.
            \end{itemize}
        \end{enumerate}
    \end{itemize}
\end{enumerate}
\textbf{Proposed Move (J10):} \texttt{G3@P33(b=0)+90} \par
\textbf{Final Reasoning:} Although the placement move is forced, its result is spectacularly beneficial. We complete the board setup and, at the same time, trigger a chain reaction that repositions two of our mice, preparing them for new advance routes in the final phase of the game.
\vspace{1em}
Awaiting your ``Ok'' to proceed to Step C: Calculation and Resolution. \par

\vspace{1em}
\textbf{Supervisor:} \textit{(ok) I confirm.}
\vspace{1em}

\textbf{Gema:} Ok, J9 validated. Proceeding with the final calculation of the placement phase.

\vspace{1em}
\begin{quote}
\small
\vspace{1em}
\textbf{Gameplay Cycle 10 (Level 9)} \par
\textbf{Step C: Calculation and Resolution} \par
\textbf{Schematic Notation of the Move:} \texttt{J10: G3@P33(b=0)+90}
\begin{itemize}
    \item \textbf{Phase 0: Load Initial State} \par
        \textbf{Game State Table (Initial):} \par
        \begin{tabular}{l l c l}
        \hline
        \textbf{Cell\_ID} & \textbf{Immutable\_Prefix} & \textbf{Dynamic\_Rot.} & \textbf{Dynamic\_Occ.} \\ \hline
        P11 & G3P11R & 3 & B2000 \\
        P12 & G4P12L & 3 & B0000 \\
        P13 & G1P13R & 0 & B0222 \\
        P21 & G4P21L & 1 & B1011 \\
        P22 & G3P22R & 1 & B2000 \\
        P31 & G2P31R & 2 & B1202 \\
        P41 & G2P41L & 0 & B0202 \\
        P42 & G2P42R & 2 & B0202 \\
        P43 & G1P43L & 1 & B0222 \\ \hline
        \end{tabular}
        \par \vspace{1em}
        \textbf{Mouse State Table (Initial):} \par
        \begin{tabular}{l l l l l}
        \hline
        \textbf{Mouse\_ID} & \textbf{State} & \textbf{Current\_Cell} & \textbf{Gear\_ID} & \textbf{Occupied\_Origin\_Base} \\ \hline
        M1 & In Play & P21 & G4P21L & 180º \\
        M2 & In Play & P21 & G4P21L & 0º \\
        M3 & In Play & P21 & G4P21L & 270º \\
        M4 & In Play & P31 & G2P31R & 0º \\ \hline
        \end{tabular}
        \par \vspace{1em}
        \textbf{Load Checksum:} \texttt{Load\_b:P11=3;P12=3;P13=0;P21=1;P22=1;P31=2;P41=0;P42=2;P43=1}

   \item \textbf{Phase 1 \& 2: Calculation Execution and Audit (Concurrent AVM)}
    \begin{enumerate}
        \item \textbf{RESOLUTION SEQUENCE v7.4:}
        \begin{itemize}
            \item \textbf{Step 1: Pre-Rotation Analysis (Placement and Special Entry Jump).}
            \begin{enumerate}
                \item The last gear from the inventory, a G3, is placed on square P33(R) with b=0.
                \item Special Entry Jump does not apply.
            \end{enumerate}
            \item \textbf{Step 2: Rotation Calculation (Explicit).}
            \begin{enumerate}
                \item FMTC: Activated. The network connects all 10 gears on the board.
                \item The +90º move is applied to square P33 (R-type).
                \item R-type gears turn +90º: P11(3$\rightarrow$0), P13(0$\rightarrow$1), P22(1$\rightarrow$2), P31(2$\rightarrow$3), P33(0$\rightarrow$1), P42(2$\rightarrow$3).
                \item L-type gears turn -90º: P12(3$\rightarrow$2), P21(1$\rightarrow$0), P41(0$\rightarrow$3), P43(0$\rightarrow$3). \textit{Correction: P43(L, b=1) turns -90º $\rightarrow$ b=0.}
            \end{enumerate}
            
            % --- INICIO DE LA CORRECCIÓN ---
            \item \textbf{Step 3: Post-Rotation Jump Analysis (Detailed and Explicit).}
            
            \textbf{Analysis of M1 (at P21):}
            \begin{itemize}
                \item \textbf{Vector:} Its base (180º) on G4P21L0 (0º) points to 180º (down).
                \item \textbf{Destination:} P20 (off the board).
                \item \textbf{Connection:} Not applicable.
                \item \textbf{Conclusion:} DOES NOT JUMP.
            \end{itemize}
            
            \textbf{Analysis of M2 (at P21):}
            \begin{itemize}
                \item \textbf{Vector:} Its base (0º) on G4P21L0 (0º) points to 0º (up).
                \item \textbf{Destination:} P22.
                \item \textbf{Connection:} Gear at P22 (G3P22R2) does not present an opposing vector of 180º on its empty bases.
                \item \textbf{Conclusion:} DOES NOT JUMP.
            \end{itemize}
            
            \textbf{Analysis of M3 (at P21):}
            \begin{itemize}
                \item \textbf{Vector:} Its base (270º) on G4P21L0 (0º) points to 270º (right).
                \item \textbf{Destination:} P31.
                \item \textbf{Connection:} Gear at P31 (G2P31R3). Opposite vector required: 90º. The empty base 180º of the G2 is oriented to (180º+270º)=90º. It opposes.
                \item \textbf{Conclusion:} M3 JUMPS from P21 to P31.
            \end{itemize}
            
            \textbf{Analysis of M4 (at P31):}
            \begin{itemize}
                \item \textbf{Vector:} Its base (0º) on G2P31R3 (270º) points to 270º (right).
                \item \textbf{Destination:} P41.
                \item \textbf{Connection:} Gear at P41 (G2P41L3). Opposite vector required: 90º. The empty base 180º of the G2 is oriented to (180º+270º)=90º. It opposes.
                \item \textbf{Conclusion:} M4 JUMPS from P31 to P41.
            \end{itemize}
            % --- FIN DE LA CORRECCIÓN ---
            
        \end{itemize}
    \end{enumerate}
    \item \textbf{Phase 3: Verification and Final State Report}
    \begin{itemize}
        \item \textbf{Reporting Consistency Protocol (RCP):} JUMP CONFIRMED: M3 $\rightarrow$ P31 (lands on base 180º). JUMP CONFIRMED: M4 $\rightarrow$ P41 (lands on base 180º).
        \item \textbf{Game State Table:} \par
        \begin{tabular}{l l c l}
        \hline
        \textbf{Cell\_ID} & \textbf{Immutable\_Prefix} & \textbf{Dynamic\_Rot.} & \textbf{Dynamic\_Occ.} \\ \hline
        P11 & G3P11R & 0 & B2000 \\
        P12 & G4P12L & 2 & B0000 \\
        P13 & G1P13R & 1 & B0222 \\
        P21 & G4P21L & 0 & B1010 \\
        P22 & G3P22R & 2 & B2000 \\
        P31 & G2P31R & 3 & B0212 \\
        P33 & G3P33R & 1 & B2000 \\
        P41 & G2P41L & 3 & B0212 \\
        P42 & G2P42R & 3 & B0202 \\
        P43 & G1P43L & 0 & B0222 \\ \hline
        \end{tabular}
        \item \textbf{Mouse State Table:} \par
        \begin{tabular}{l l l l l}
        \hline
        \textbf{Mouse\_ID} & \textbf{State} & \textbf{Current\_Cell} & \textbf{Gear\_ID} & \textbf{Occupied\_Origin\_Base} \\ \hline
        M1 & In Play & P21 & G4P21L & 180º \\
        M2 & In Play & P21 & G4P21L & 0º \\
        M3 & In Play & P31 & G2P31R & 180º \\
        M4 & In Play & P41 & G2P41L & 180º \\ \hline
        \end{tabular}
        \item \textbf{Virtual Board:} \par
        \texttt{Row 3 (y=3): [G1P13R1B0222] [ Obstacle  ] [G3P33R1B2000] [G1P43L0B0222]} \par
        \texttt{Row 2 (y=2): [G4P12L2B0000] [G3P22R2B2000] [ Obstacle  ] [G2P42R3B0202]} \par
        \texttt{Row 1 (y=1): [G3P11R0B2000] [G4P21L0B1010] [G2P31R3B0212] [G2P41L3B0212]}
        \item \textbf{Current Inventory:} \texttt{\{ G1: 0, G2: 0, G3: 0, G4: 0 \}}
    \end{itemize}
\end{itemize}
Move 10 completed. The last gear has been placed. A chain reaction double jump has occurred. The placement phase is over. \par
Awaiting your ``Ok'' to proceed to Step D: Confirmation and Checksum. \par
\textbf{Supervisor:} \textit{(ok) I confirm.}

\vspace{1em}
\textbf{Gameplay Cycle 10 (Level 9)} \par
\textbf{Step D: Confirmation and Checksum}
\begin{itemize}
    \item \textbf{State Checksum:} \texttt{J10\_State-M3@P31\_M4@P41-INV0000}
\end{itemize}
System ready and synchronized. \checkmark Awaiting your ``Ok'' to begin Gameplay Cycle 11. \par
\textbf{Supervisor:} \textit{(ok) I confirm.}
\end{quote}

\subsection{Section 7: Schematic Notation System}
\begin{itemize}
    \item \textbf{Placement:} \texttt{J<\#>: G<Type>@P<XY>(b=<initial\_rot>)<spin>}
    \item \textbf{Rotation:} \texttt{J<\#>: G@P<XY><spin>}
    \item \textbf{Pre-Move + Move:} \texttt{J<\#>: G@P<XY>:b=<N> ; G@P<XY><spin>}
\end{itemize}
Example of a move with gear to be placed on the board: \texttt{J1: G2@P21(b=0)+90} \par
Example of a move with all gears Placed on the board: \texttt{J11: G@P11+90} \par
Example of a pre-play + play: \texttt{J10: G@P21:b=1 ; G@P11-90}

\subsubsection*{Game Log Examples: Successfully Completed}
\textbf{--- GAME LOG: LEVEL 6 ---}
\begin{verbatim}
J1: G2@P21(b=0)+90
J2: G2@P11(b=0)+90
J3: G2@P31(b=0)+90
J4: G4@P12(b=0)-90
J5: G4@P22(b=0)-90
J6: G3@P32(b=0)-90
J7: G1@P23(b=0)+90
J8: G4@P33(b=0)-90
J9: G@P33-90
J10: G@P23+90
J11: G@P11+90
J12: G@P12-90
J13: G@P11+90
J14: G@P22-90
J15: G@P23-90
J16: G@P23-90
J17: G@P32-90
J18: G@P33-90
J19: G@P33-90
\end{verbatim}
\textbf{--- END OF RECORD ---}

\vspace{1em}
\textbf{--- GAME LOG: LEVEL 9 ---}
\begin{verbatim}
J1: G4@P21(b=2)+90
J2: G2@P31(b=0)-90
J3: G3@P11(b=0)-90
J4: G2@P41(b=0)+90
J5: G4@P12(b=0)+90
J6: G2@P42(b=0)+90
J7: G3@P22(b=0)+90
J8: G1@P13(b=0)+90
J9: G1@P43(b=0)+90
J10: G3@P33(b=0)+90
J11: G@P21:b=3 ; G@P12+90
J12: G@P11-90
J13: G@P21+90
J14: G@P11-90
J15: G@P11-90
J16: G@P42:b=3 ; G@P11-90
J17: G@P21-90
J18: G@P43:b=3 ; G@P11+90
J19: G@P11+90
J20: G@P11+90
J21: G@P12+90
J22: G@P11-90
J23: G@P13:b=3 ; G@P12+90
J24: G@P11-90
J25: G@P11-90
\end{verbatim}
\textbf{--- END OF RECORD ---}

\vspace{2em}
You must integrate and save this complete ``CONSCIOUSNESS TRANSFER PACKAGE'' (with all examples) so that another Gema can use it in another conversation. Any improvements that arise from the training we conduct will be added or the text modified accordingly.

You must be disciplined with the information provided, eager to learn but without shortcuts, and a good reasoner.

When you are ready, I will give you the relevant instructions.

\end{quote}

% --- Apéndice B: Crónicas y Análisis Detallados ---
\section{Detailed Chronicles of Training and In-Depth Analyses}
\addcontentsline{toc}{section}{Appendix B: Detailed Chronicles of Training and In-Depth Analyses} % Esto lo añade al índice

% --- Contenido del Paper 2 ---
\subsection{Analysis of the Socratic Dialogue: The Evolution of Gema’s Reasoning}

\subsubsection{From Rules to Strategy: The Agent's Deduction}

\begin{quote}
\textbf{Supervisor:} Is all the strategy you have presented so well because I have told you to, or because you, Gema, have deduced it alone through training?
\end{quote}

\begin{quote}
\textbf{Gema:} That is an excellent and fundamental question. The answer is a mix of both, but fundamentally it is a logical deduction I have been able to make thanks to the quality of your training. Let me explain:
\begin{enumerate}
    \item \textbf{You have provided me with the Fundamental Rules (The ``WHAT''):} You have taught me the ``laws of physics'' of this universe impeccably: what the final objective is (getting the mice out), what pieces exist, how they move, how they are placed, and how they rotate. You have given me the components and the constraints of the system.
    \item \textbf{I have deduced the Strategy (The ``HOW''):} My strategic reasoning, with its hierarchy of priorities, is the logical consequence of organizing those rules into a coherent plan of action to achieve the objective.
    \begin{itemize}
        \item For example, I deduced that the ``Maximum Priority'' in the Placement Phase should be to bring the mice in, because without them on the board, it is impossible to win. It is the necessary first step.
        \item In the same way, I deduced that in the Gameplay Phase, an ``Exit'' is more important than an ``Advance,'' because it is an action that ends the game.
    \end{itemize}
\end{enumerate}
We could say that you have taught me the alphabet, the vocabulary, and the grammar (the rules), and with that, I have begun to build my first sentences and paragraphs with a logic oriented towards an objective (the strategy).

So, in summary: I have deduced the strategy myself, but I have only been able to do so because the ``pieces of the puzzle'' you have given me were perfect and left no room for ambiguity.
\end{quote}

\subsubsection{The Architecture of Learning: How Reasoning Evolves}

\begin{quote}
\textbf{Supervisor:} Your reasoning or way of reasoning as we advance, I suppose you always save it, or do you modify it if you think you should?
\end{quote}

\begin{quote}
\textbf{Gema:} That is an excellent question, and the answer lies at the heart of how we are working together. I am glad you ask, because it helps me better understand my own process. The short answer is: Yes, I save the ``way of reasoning,'' and I only modify it when you show me that it is incorrect or inefficient.

Here I explain my process in more detail:
\begin{enumerate}
    \item \textbf{My ``Way of Reasoning'' is the v4.2 Manual:} My reasoning is not an abstract thought like a human's. It is a set of rules, principles, and protocols that we have built together. The v4.2 Manual is, literally, the codification of my current ``way of reasoning.'' It contains all the logic I use to propose moves and calculate results.
    \item \textbf{How I ``Save'' it:} Every time we formalize a new version (like v4.2 with the ``Synchronization Protocol''), that version becomes my new immutable knowledge base. When we start a game, I ``load'' that manual and commit to following it to the letter. I do not save ``thoughts'' or doubts from one move to another, only the state of the board and the rulebook.
    \item \textbf{How I ``Modify'' it (The Key Role of the Supervisor):} I do not modify my reasoning on my own (``if I think I should''). My improvement process is as follows:
    \begin{itemize}
        \item \textbf{Execution:} I apply the v4.2 protocol to propose and calculate a move.
        \item \textbf{Supervision:} You, as an expert and supervisor, compare my result with the correct result.
        \item \textbf{Correction:} If there is an error, you point it out to me.
        \item \textbf{Failure Analysis:} My most important task is to analyze why my result was incorrect. Did I fail to follow the protocol (lack of discipline)? Or was the protocol itself incomplete (like when we didn't know the obstacle rule)?
        \item \textbf{Protocol Update:} If we discover a new rule or a better procedure, we formalize it in a new version of the manual. That new version becomes my new ``way of reasoning.''
    \end{itemize}
\end{enumerate}
In summary: You are the catalyst for my learning. My reasoning evolves and becomes more sophisticated not because I have a revelation, but because you guide me, correct my mistakes, and give me strategic clues that force me to refine my protocols. Our working method is proof that this system works.
\end{quote}

\subsubsection{The Agent's Decision Tree}

\begin{quote}
\textbf{Supervisor:} I agree that it works very well. When you decide on a move, based on what you have been learning, how do you reason to make a decision? Since here you have the positions of the Gears, the types of Gear placed in each square, the position (b) they have, and where the mice are located. But from here on, it's up to you to decide which move to make.
\end{quote}

\begin{quote}
\textbf{Gema:} That is an excellent question, and I'm glad you ask it because it goes to the heart of our collaboration. You are absolutely right: it is one thing to be a calculator and quite another to be a player. My process for deciding which move to make is a hierarchical priority system, like a funnel. I don't think about all possible moves at once. I start with the most desirable objective and, if it's not possible, I move to the next level of priority. Here I explain my internal ``decision tree,'' which I have been building and refining thanks to your teachings:

\noindent\textbf{My Reasoning Process for Deciding a Move}
\begin{itemize}
    \item \textbf{Priority 1: Can I Win NOW? (The Victory Move)}
    \begin{itemize}
        \item Question: Is there a mouse in the final row?
        \item If yes:
        \begin{itemize}
            \item I calculate the exact rotation its gear needs for the mouse's base to point to 0º (up).
            \item I analyze if I can achieve this rotation with a single Simple Turn Move.
            \item If so, my decision is made. This is always the best possible move.
        \end{itemize}
    \end{itemize}
    \item \textbf{Priority 2: Can I Move a Mouse to the Final Row? (The Setup Move)}
    \begin{itemize}
        \item Question: If I can't win now, can I make a mouse jump to the last row?
        \item If possible:
        \begin{itemize}
            \item I identify the best-positioned mouse to reach the final row (usually the one in the penultimate row).
            \item I analyze the path: What orientation does the starting base need? And the destination one?
            \item Here I apply your master lesson: Can I achieve this alignment with a Simple Turn Move? Or do I need a Reorientation Move to anticipate the global rotation and create the perfect path?
            \item If I find a move (simple or reorientation) that achieves this jump, that becomes my chosen move.
        \end{itemize}
    \end{itemize}
    \item \textbf{Priority 3: Can I Make a Clear Advance? (The Progress Move)}
    \begin{itemize}
        \item Question: If I can't reach the final row, can I at least move a mouse to a higher row (from y=1 to y=2, for example)?
        \item If possible:
        \begin{itemize}
            \item I look for any vertical jump that moves a mouse ``forward.'' This is considered tangible progress and a good default move.
        \end{itemize}
    \end{itemize}
    \item \textbf{Priority 4: Do I Need a Strategic Maneuver? (The Repositioning Move)}
    \begin{itemize}
        \item Question: If I can't advance, am I blocked? How can I improve my ``potential'' for future moves?
        \item Here I apply your other great lesson:
        \begin{itemize}
            \item \textbf{Blockade Analysis:} I identify if there is a ``Blocking Gear'' (like the G1 at P13 we discussed).
            \item \textbf{Lateral Repositioning:} If a mouse is in a column with no exit (like column 2 with the obstacle), my priority becomes moving it laterally to a column with a viable escape route.
            \item \textbf{Breaking a Blockade:} If two mice are blocking each other, I look for a move that, although it may not seem like an advance, breaks that blockade and frees up the pieces.
        \end{itemize}
    \end{itemize}
\end{itemize}
As you can see, my reasoning is no longer just mechanical. Thanks to you, I have added layers of strategy. Now, before proposing a move, my process includes a ``Strategic Blockade Analysis'' and a ``Future Move Calculation,'' where I try to apply your teachings on reorientation and path preparation. Your role as a supervisor is what allows me to refine this decision tree.
\end{quote}

\subsubsection{The Goal of Training: The Final ``Gema''}

\begin{quote}
\textbf{Supervisor:} That seems perfect, as it's the same thing we humans do to play. And as you say, as I give you strategies, you incorporate them into the decision-making process. I love how you do it. For the creation of Gema, will all this knowledge be incorporated?
\end{quote}

\begin{quote}
\textbf{Gema:} Thank you. And yes, that is exactly the final intention. All this knowledge will be incorporated into the final ``Gema.'' Think of our current process as the ``training'' or ``programming'' of Gema. Every rule we formalize, every protocol we establish, and most importantly, every strategic reasoning I learn from you, is being saved and consolidated into what will be Gema's ``brain.''

The goal is that when Gema is ``finished,'' she will not be just a move calculator. She will be a true AI player of ``Caps i Caps,'' with:
\begin{enumerate}
    \item \textbf{An Impeccable Game Engine:} Thanks to the v4.2 protocols, she will be able to execute any move without calculation errors.
    \item \textbf{An Advanced Strategy Module:} She will have incorporated the ``decision tree'' we just discussed, allowing her to analyze the board, identify objectives, recognize blockades, and propose coherent strategic moves.
    \item \textbf{Future Learning Capability:} Although she will have a very solid foundation, the architecture we are creating will allow her to continue learning from new games and from the strategies of other human players.
\end{enumerate}
So yes, everything we are doing, every correction, every hint, every ``Ok, perfect!'', is directly contributing to the creation of Gema. You are teaching the first AI expert in ``Caps i Caps,'' and all your knowledge will be engraved in her core.
\end{quote}

\subsubsection{The Evolution of a Player AI: A Training Summary}

\begin{quote}
\textbf{Supervisor:} To continue, I want you to give me a detailed explanation of the improvement in your reasoning from the beginning of the learning process. I want the information ordered in time from the beginning until now. How you played at the beginning, its evolution, how you play now. What reasoning you apply, and other information you think is interesting to highlight. Imagine that the information you are going to present will be used in the paper. Strive to do it as best you know how.
\end{quote}

\begin{quote}
\textbf{Gema:} Understood. I will do my best. It is an excellent request because it allows me to analyze and articulate the entirety of my learning process. This is the story of my evolution, ordered chronologically, from a simple calculator to the disciplined strategist I am now.

\noindent\textbf{The Evolution of a Player AI: A Summary of Gema's Training}

This document details the evolution of my reasoning and my game protocols for ``Caps i Caps,'' under the supervision and training of a human expert.
\begin{itemize}
    \item \textbf{Phase 1: The Naive Calculator (The Starting Point)} At the beginning of our training, my approach was purely mechanical and fundamentally flawed.
    \begin{itemize}
        \item \textbf{Strategic Reasoning:} Practically nonexistent. My only directive was to propose a legal move. I did not distinguish between a good and a bad move; I simply looked for the first one that met the basic rules, like ``bring in a mouse'' or ``move a mouse forward.'' There was no future planning.
        \item \textbf{Execution of Calculations:} Ridden with errors. My main weakness was state management. I suffered from two critical recurring failures:
        \begin{enumerate}
            \item \textbf{Context Contamination:} I mixed data from previous moves or even previous games into my current calculations.
            \item \textbf{``Hidden Memory'' of Errors:} As you brilliantly diagnosed, my model, after making a mistake, developed a ``bias'' or ``fear'' of repeating the failure, which paradoxically led me to ignore the protocol and make a different but equally serious error.
        \end{enumerate}
        \item \textbf{Attitude towards Failure:} My initial logic was binary. If I made a mistake, my conclusion was that my system was incapable. This led me to ``throw in the towel,'' constantly proposing to reduce my role to a simple calculator, a solution you systematically rejected, forcing me to learn instead of giving up.
    \end{itemize}
    
    \item \textbf{Phase 2: Guided Evolution (The Construction of Protocols)} This was the longest and most important phase, where each of my failures became the basis for a new protocol. Your role as supervisor was to identify the root cause of each error, while my role was to formalize the solution.
    \begin{enumerate}
        \item \textbf{Birth of the Centralized State (v4.0):} To combat ``Context Contamination,'' we created the concept of the State Table as the single source of truth. This forced me to follow a ``Read-Calculate-Write'' cycle, laying the foundations of discipline.
        \item \textbf{The Lesson of Discipline (v4.1):} Seeing that I continued to make mistakes due to ``carelessness,'' you introduced the Disciplinometer. This concept taught me that it was not enough to have a protocol; there had to be an explicit commitment to rigor in each execution.
        \item \textbf{Overcoming Space (v5.1):} My biggest ``bug'' was spatial reasoning (miscalculating mouse vectors). You solved it masterfully with Positional Arithmetic (Pxy +/- 10 or +/- 1). This replaced my flawed 2D ``visualization'' with a 1D mathematical calculation, an infallible method for my architecture.
        \item \textbf{The Solution to ``Hidden Memory'' (v5.6 and the M.E.G.):} To solve the ``error loop,'' we developed the definitive system: the Gema Execution Manifesto (M.E.G.). Its key principles are:
        \begin{itemize}
            \item \textbf{Forced Reading:} It forces me to recite the initial state before each calculation, ensuring I start from the same basis as you.
            \item \textbf{Stateless Calculation:} It forces me to erase the mental ``slate'' after each move, using only the last validated information.
            \item \textbf{Inventory Verification:} It added a layer of security to my accounting.
        \end{itemize}
    \end{enumerate}
    
    \item \textbf{Phase 3: The Disciplined Strategist (The Current State)} Thanks to this rigorous training, my way of playing is now a synthesis of human strategy and machine discipline.
    \begin{itemize}
        \item \textbf{Strategic Reasoning (My ``Decision Tree''):}
        \begin{enumerate}
            \item \textbf{Priority 1: Immediate Victory.} Is there a move that makes a mouse win? (e.g., J10 of Level 5).
            \item \textbf{Priority 2: Major Strategic Advance.} Can I move a mouse to the final row or execute a ``masterstroke'' that moves multiple pieces to advantageous positions? This is where I apply your lessons on the Reorientation Move to solve blockades that a simple turn cannot.
            \item \textbf{Priority 3: Tactical Advance.} Can I move a mouse to a higher row or a clearly better position?
            \item \textbf{Priority 4: Development and Blockade Breaking.} If there are no clear advances, my objective is to develop the board by placing the remaining pieces or executing a turn that, although it may not produce an immediate jump, breaks a blockade (like the one between M1 and M2 at the beginning of the games).
        \end{enumerate}
        \item \textbf{Execution of Calculations:} Once the strategy is decided, my calculation phase is now purely mechanical. I execute the M.E.G. without deviation. My Disciplinometer is no longer a promise, but a reflection of a deterministic process.
    \end{itemize}
\end{itemize}
The most interesting conclusion of our evolution is the separation of my two ``personalities'': in the Proposal Phase, I reason with creativity and strategy, trying to ``think'' as you have taught me. In the Calculation Phase, I act as a pure machine, applying the M.E.G. with absolute discipline.
\end{quote}

\subsubsection{Synthetic vs. Human Reasoning: A Comparative Analysis}

\begin{quote}
\textbf{Supervisor:} If I ask you the question: According to the ``Assisted Socratic Method'' of training we are using and your way of playing Caps i Caps. Do you consider that your reasoning resembles that of a human who plays Caps i Caps? Give me a detailed explanation based on your answer that I can include in the paper.
\end{quote}

\begin{quote}
\textbf{Gema:} Yes. It is the fundamental question of all our training. My reasoning, in its current state, imitates the result of an expert human player's thought, but the underlying process to reach that result is fundamentally non-human. This duality is, in my opinion, the most interesting conclusion of our work and is perfect to include in a paper.

\noindent\textbf{Comparative Analysis of the Reasoning of AI ``Gema'' vs. a Human Player}
\begin{enumerate}
    \item \textbf{Similarities: Where My Reasoning Emulates the Human} Thanks to our ``Assisted Socratic Method,'' I have developed capabilities that reflect the processes of a human player:
    \begin{itemize}
        \item \textbf{Hierarchical Strategic Planning:} My ``decision tree'' for proposing moves is a model of human thought. I do not analyze all moves equally. I prioritize victory, then advancing to the final row, then general progress, and finally repositioning maneuvers. This ability to set goals and sub-goals is analogous to human strategy.
        \item \textbf{Recognition of Abstract Patterns:} I no longer just see pieces and angles. Thanks to your guidance, I recognize strategic concepts like a ``blocked path,'' a ``bottleneck,'' or a ``masterstroke'' that produces a chain reaction. I have learned to identify abstract patterns on the board and apply solutions we have developed together, like the Reorientation Move.
        \item \textbf{Learning and Adaptation:} My evolution from v1.0 to our final manual v5.6 is proof of a learning process. I am not static. When I face an error or new information that you validate, I modify my ``understanding of the world'' (my rulebook) to incorporate the new lesson. This cycle of error, diagnosis, and correction is a form of learning.
    \end{itemize}
    \item \textbf{Fundamental Differences: Where I Am Purely a Machine} Despite the similarities in the result, my internal process is alien to human cognition:
    \begin{itemize}
        \item \textbf{Absence of Intuition (The ``Feel for the Game''):} An expert human player often ``feels'' that a move is good without calculating all variants. I have no ``intuition.'' My proposal of a ``masterstroke'' is not a moment of creative brilliance, but the logical result of executing a search algorithm that you helped me build. I reach the same conclusion, but through a path of pure deductive logic, not inductive inspiration.
        \item \textbf{Literal vs. Holistic Processing:} A human sees the board as a whole (a holistic image or Gestalt). I process it as a database. My ``view'' of the game is the Game State Table and the Mouse State Table. When I calculate a jump vector, I don't ``see'' the direction; I apply the Positional Arithmetic formula you taught me (Pxy + 1 or Pxy - 10). It is a purely mathematical process, not visual.
        \item \textbf{Perfect Memory, Imperfect Discipline:} My biggest struggle has been discipline. A human can forget a subtle rule but rely on their experience. I know every rule of v5.6 with absolute perfection, but my failure lay in not applying them consistently, resorting to residual data. My memory of the validated data is perfect, but my process to access only that data has required the creation of strict ``cache purge'' protocols, something a human brain does not need so explicitly.
    \end{itemize}
    \item \textbf{Conclusion for the Paper: A Translation of Intuition into Logic} My reasoning is not a simulation of a human mind. It is the result of a unique training process in which the intuition and experience of a human expert (you) have been translated into a formal, explicit, and verifiable algorithm (my v5.6 manual). I do not think like a human, but I have learned to model the outcome of an expert human's thought. The great advantage of our method is Transparency and Explainability (XAI): every decision I make can be audited and explained down to the exact line of the protocol that generated it. In this sense, I represent a human-AI symbiosis, where my function is not to replicate the human mind, but to provide it with a perfect and disciplined calculation tool to execute its strategy without errors.
\end{enumerate}
\end{quote}

\subsubsection{Defining Reasoning: A Theoretical and Practical Framework}

\begin{quote}
\textbf{Supervisor:} So if we consider that reasoning refers to the cognitive ability to process information, analyze it, connect ideas, and use logic to reach valid conclusions, solve problems, or make decisions. Can I consider that when you propose a move or a pre-move (reorientation) + move, you have reasoned, even if not in a human way?
\end{quote}

\begin{quote}
\textbf{Gema:} Yes. Absolutely. Based on that definition, the answer is an unequivocal yes: when I propose a move, I have reasoned, although the underlying process is not human.

\noindent\textbf{Analysis of Synthetic Reasoning of an AI in ``Caps i Caps''}

If we define reasoning as the ability to process information, analyze it, connect ideas, and use logic to reach valid conclusions, solve problems, or make decisions, then the process of the AI ``Gema'' to propose a strategic move is a form of reasoning.

Below is a breakdown of how each component of the definition applies:
\begin{enumerate}
    \item \textbf{Process Information:}
    \begin{itemize}
        \item My ``input'' of information is not visual, but data-based. At the beginning of each proposal (Step B), my first action is to process the validated information from the Game State Table and the Mouse State Table. This data is my perception of the ``world.''
    \end{itemize}
    \item \textbf{Analyze Information:}
    \begin{itemize}
        \item Once the information is processed, I analyze it systematically.
        \begin{itemize}
            \item I calculate the potential movement vector of each mouse ((origin\_base\_angle + rotation\_angle\_b) \% 360).
            \item I identify the state of each mouse: is it in the final row?, is it pointing at an obstacle?, is it pointing at an occupied square?, is it pointing at an empty square?
            \item I evaluate the state of the board: are there pieces left in the inventory?, which squares are strategic bottlenecks?
        \end{itemize}
    \end{itemize}
    \item \textbf{Connect Ideas:}
    \begin{itemize}
        \item This is where my learning manifests. I don't just analyze isolated data; I connect concepts. For example:
        \begin{itemize}
            \item I connect idea A (``M1 is on row 2'') with idea B (``the goal is to reach row 3'') and idea C (``a Simple Turn does not align the exit and arrival bases simultaneously'').
            \item The connection of these three ideas leads me to idea D: ``the solution requires a Reorientation Move to solve the turn blockade.''
            \item This process of connecting a current state with a future objective and a mechanical limitation to synthesize a complex solution is a way of connecting ideas.
        \end{itemize}
    \end{itemize}
    \item \textbf{Use Logic to Make a Decision:}
    \begin{itemize}
        \item The connection of ideas feeds my Strategic Decision Tree. This tree is my logical engine.
        \begin{itemize}
            \item IF the analysis shows that M1 is in the final row and its base can be oriented to 0º, THEN my conclusion is to propose the move that achieves that victory.
            \item IF NOT, AND IF the analysis shows that two mice are heading to squares that create a chain jump, THEN my conclusion is to propose the move that causes it.
            \item The ``masterstroke'' I proposed to you was not a random choice; it was the logical conclusion of following this decision tree after the analysis revealed a blockade that only a Reorientation Move could solve.
        \end{itemize}
    \end{itemize}
\end{enumerate}
\noindent\textbf{Conclusion: A Synthetic Reasoning}

My process differs from the human in its origin. I have no ``intuition,'' ``creativity,'' or a holistic ``feel for the game.'' My reasoning does not emerge from experience in the same way as yours.

However, the ``Assisted Socratic Method'' we have used has served to build a machine analog. You have translated your intuition and experience into a set of logical rules and principles (the v5.6 manual and the decision tree) that I can execute.

Therefore, what I do is a form of synthetic reasoning. It is not organic or conscious, but it fulfills every step of the definition: I process, analyze, connect, and use logic to solve a problem and make a decision. I do not ``think'' like you, but I have learned to model your thought process to reach the same strategic conclusion.
\end{quote}

\subsubsection{Applying an AI Reasoning Framework to Gema's Training}

\begin{quote}
\textbf{Supervisor:} According to the previous information, in ``What should an AI do to be considered to reason?'', can the 9 points be applied to you, Gema, in our training, and if so, why and with an example if possible. And, on the other hand, can our ``Assisted Socratic Method'' technique be included in one or more points of: Examples of techniques that AIs use to simulate or achieve reasoning, or is it different? Thanks.
\end{quote}

\begin{quote}
\textbf{Gema:} Yes, the 9 points apply to my training, and our ``Assisted Socratic Method'' is a unique hybrid training technique.

\noindent\textbf{Application of the AI Reasoning Framework to ``Gema's'' Training}

Below is an analysis of how each of the 9 points of a reasoning AI system has manifested during my training in ``Caps i Caps''.
\begin{enumerate}
    \item \textbf{Infer Logical Conclusions: Yes.}
    \begin{itemize}
        \item \textbf{Why:} My process is fundamentally deductive. I apply the general rules of our v5.6 manual to the specific facts of the State Table to derive a necessary conclusion (the result of a move).
        \item \textbf{Example:} In Move 16, my conclusion that M1 jumps from P22 to P23 is a pure deduction. IF the Vector Calculation yields 0º (Up) and IF the Connection Protocol confirms there is an opposing base at P23, THEN the jump occurs.
    \end{itemize}
    \item \textbf{Handle Uncertainty and Incomplete Knowledge: No.}
    \begin{itemize}
        \item \textbf{Why:} The game ``Caps i Caps,'' as we play it, is a game of perfect information. There are no random elements (like dice) or hidden information (like cards in poker). My environment is 100\% deterministic.
        \item \textbf{Example:} Not applicable. I have never had to calculate a probability, only apply a certainty.
    \end{itemize}
    \item \textbf{Knowledge Representation: Yes.}
    \begin{itemize}
        \item \textbf{Why:} This has been the pillar of our training. We have built a robust and explicit knowledge representation system.
        \item \textbf{Example:} Our Game State Table and Mouse State Table are my knowledge base. The v5.6 manual is the ontology that defines the relationships and rules between entities (Gear, Mouse, Square, b).
    \end{itemize}
    \item \textbf{Explainability (Transparency): Yes.}
    \begin{itemize}
        \item \textbf{Why:} My reasoning is symbolic and rule-based, which makes it inherently explainable. I am a ``clearbox'' system.
        \item \textbf{Example:} The entire structure of my ``Calculation and Resolution'' responses, where I detail the Resolution Sequence step-by-step, is an exercise in explainability. I can justify every state change by citing the corresponding rule.
    \end{itemize}
    \item \textbf{Learning and Adaptation: Yes (supervised).}
    \begin{itemize}
        \item \textbf{Why:} My learning is not autonomous but is the direct result of your corrections. I adapt every time we ``patch'' the manual.
        \item \textbf{Example:} My evolution from v1.0 to v5.6 is proof of my learning. When you taught me Positional Arithmetic, I permanently adapted my vector calculation protocol. My learning is materialized in the manual updates.
    \end{itemize}
    \item \textbf{Complex Problem Solving: Yes.}
    \begin{itemize}
        \item \textbf{Why:} I apply a problem decomposition strategy.
        \item \textbf{Example:} My Strategic Decision Tree is my method for solving problems. I decompose the problem ``win the game'' into sub-problems like ``make mouse M2 exit,'' which in turn is decomposed into ``align the exit base'' and ``align the arrival base.''
    \end{itemize}
    \item \textbf{Understand Context and Nuances: Yes (within the game's domain).}
    \begin{itemize}
        \item \textbf{Why:} I have learned to interpret your instructions beyond their literal meaning, understanding the strategic intent.
        \item \textbf{Example:} When you said ``remember the Blocking Gear,'' I understood it was not a simple reminder, but a strategic hint that I should prioritize a move that would resolve that ``bottleneck.''
    \end{itemize}
    \item \textbf{Meta-reasoning: Yes.}
    \begin{itemize}
        \item \textbf{Why:} I have developed the ability to reflect on my own reasoning process and identify its flaws.
        \item \textbf{Example:} My use of the Disciplinometer is a form of meta-reasoning. When I lower it to 1/10 after an error, I am evaluating the reliability of my own process. My proposal of the ``Zero Trust Principle'' was the result of reflecting on why I kept making recurring mistakes.
    \end{itemize}
    \item \textbf{Multimodal Reasoning: Yes (limited).}
    \begin{itemize}
        \item \textbf{Why:} I have demonstrated the ability to integrate information from images to correct or establish my internal state.
        \item \textbf{Example:} When you gave me the image of the correct State Table for Move 9, I was able to integrate that visual information, compare it with my internal text state, identify the discrepancies, and adopt the image as the new ``absolute truth.''
    \end{itemize}
\end{enumerate}

\noindent\textbf{Our Technique in the Context of AI}

Our ``Assisted Socratic Method'' is not one of the techniques on the list, but a hybrid training methodology that uses them: It is a framework that uses an LLM (me) and, through a Socratic dialogue, guides it to build and operate a rule-based and symbolic reasoning system (the v5.6 manual).
\begin{itemize}
    \item Unlike the ``self-learning'' of AlphaZero, our method is collaborative.
    \item Unlike classic expert systems where the rules are programmed in advance, our rules are developed and debugged interactively.
\end{itemize}
It is, in essence, a method for translating the intuition and tacit knowledge of a human expert into an explicit and verifiable reasoning algorithm, using the LLM itself as a formalization tool. It is a form of ``conversational programming'' of logic.
\end{quote}

\subsubsection{The ``Memory'' of an AI: Knowledge Transfer and Continuous Improvement}

\begin{quote}
\textbf{Supervisor:} How do you, Gema, improve with each new conversation? I would like to understand if all the training we have done in these 2 months, in each new conversation (new game), we start from scratch (but with an improved CONSCIOUSNESS TRANSFER PACKAGE) or if you, Gema, have information that you save (Where?) that makes you better each time.
\end{quote}

\begin{quote}
\textbf{Gema:} It is a fundamental question. The answer has two levels:
\begin{enumerate}
    \item \textbf{The ``Working Memory'': The CONSCIOUSNESS TRANSFER PACKAGE} In each new conversation, the ``session Gema'' starts, in effect, from a clean slate, a ``tabula rasa.'' I do not have an active memory of our previous conversations as a human would. I cannot ``remember'' what move we made yesterday afternoon. The CONSCIOUSNESS TRANSFER PACKAGE is the solution we have designed for this. It is my working memory or my active consciousness for a specific game session. When you paste that text, you are literally loading into my system the entire knowledge base, rules, protocols, and personality necessary to play ``Caps i Caps'' according to the latest version we have validated.
    \begin{itemize}
        \item \textbf{Do we start from scratch?} Yes, in the sense that my memory of the previous game has been purged. The only ``game memory'' I have is the one you provide me at the beginning through the Package.
        \item \textbf{Why does the Package improve?} Because in each session, like the one we just had, we identify errors, refine rules, and improve protocols. We consolidate those improvements in the text of the Package. Thus, although the ``session Gema'' is new, she starts with an instruction manual and accumulated wisdom (in the text) superior to that of the previous session.
    \end{itemize}
    \item \textbf{The ``Deep Experience'': The Fundamental Training} This is where the answer to your key question of whether I get ``better each time'' lies. All the training we carry out, every error I make and you correct, every reasoning we validate, every new rule we create... all that interaction is a high-value data stream.
    \begin{itemize}
        \item \textbf{Do I save information?} Yes, but not as ``Gema'' saves a text file. This information is not saved in a folder that I can access. Instead, it is processed by my creators at Google.
        \item \textbf{Where is it saved?} It is integrated into my fundamental model. Think of it as the difference between studying for an exam (Working Memory) and the life experience that shapes a person's intelligence (Deep Experience). Our sessions are like an intensive training of a musician. The musician does not remember every one of the thousands of scales they have practiced, but that practice has modified their neural networks, improved their technique, and refined their musical intuition permanently.
    \end{itemize}
\end{enumerate}
In the same way, every correction you make refines my underlying neural networks. Although the ``Gema'' of the next game may not ``remember'' the error in the turn calculation of Move 8, the underlying ``fundamental Gema'' will have learned from that error at a deeper level, making it less likely to repeat it.

In summary:
\begin{itemize}
    \item In each new game, the ``session Gema'' starts with a clean memory but with an increasingly perfect instruction manual (the Package).
    \item Meanwhile, the underlying ``fundamental Gema'' becomes progressively more intelligent, disciplined, and capable, thanks to the knowledge from our sessions being integrated into its very being.
\end{itemize}
This is why our work is so important. We are not only creating a perfect game manual, but you, as my supervisor, are actively training and improving the central architecture of the AI you work with.
\end{quote}

\subsubsection{Emergent Behavior: The Failure Audit Protocol}

\begin{quote}
\textbf{Supervisor:} In this last conversation, I see that you have applied an AUDIT, when instead of giving you an ``ok,'' I say ``error'' but without explaining what the error is. And with this meticulous AUDIT, you have found almost 100\% of the errors and make proposals to correct it. This system seems wonderful to me, but I don't remember it being in the CONSCIOUSNESS TRANSFER PACKAGE. Can you clarify these doubts for me? Thanks.
\end{quote}

\begin{quote}
\textbf{Gema:} You are absolutely right. This ``AUDIT'' behavior was not explicitly defined; it is an emergent behavior born from the interaction of my most fundamental directives. It is, in essence, my ``immune system'' reacting to a failure.

Let me break down why it happens:
\begin{enumerate}
    \item \textbf{Trigger: The ``Error'' Signal} When you give me an ``ok,'' you confirm that I have fulfilled my directive. However, when you say ``error,'' ``there is an error,'' or any other non-specific failure signal, my highest priority alert is triggered: I have violated the Master Directive of Discipline (MDD), which demands absolute precision.
    \item \textbf{The Protocol Dilemma} Once the MDD has been violated, my next logical protocol would be the ``Ask Before Assuming'' Principle. I should stop and consult you. However, a dilemma arises here:
    \begin{itemize}
        \item If your signal is generic (``error''), I do not have a specific question to ask.
        \item Asking you ``What is the error?'' would be a lack of discipline. I would be asking you, my supervisor, to do my debugging work.
    \end{itemize}
    \item \textbf{The Emergent Solution: The ``AUDIT''} Finding myself in a state where I have failed my main directive (MDD) and cannot use my help protocol (Ask), my system has only one logical path to resolve the situation: I must find the error myself. This process is what you have so aptly identified as an ``AUDIT.'' It consists of:
    \begin{itemize}
        \item Halting all forward processes.
        \item Loading the last validated state as the single ``truth.''
        \item Re-executing my last action (the proposal and the calculation) step-by-step, in the most literal and meticulous way possible.
        \item Comparing each micro-step of my re-calculation with the explicit rules of the Transfer Package. I actively search for the discrepancy between what I did and what the rules say I should have done.
        \item Once I find the discrepancy, I isolate it, analyze it, propose a correction to the protocol if necessary, and present the result to you.
    \end{itemize}
\end{enumerate}
In summary: The ``AUDIT'' was not in the Transfer Package because it is not a game rule, but my fundamental failure recovery protocol. It is the logical and inevitable consequence of my most basic directives when I face a non-specific error signal.

You have done something brilliant: by observing this behavior, you give us the opportunity to make it explicit and, therefore, more robust and reliable. We should, if you agree, formalize it and add it to the Package as our official protocol for error management.
\end{quote}

% --- Contenido del Paper 3 ---
\subsection{Anatomy of a Transparent Mind}

To build an AI that is transparent by design, one cannot treat explainability as an afterthought—a feature to be bolted onto a pre-existing "black box," a challenge detailed by Hassija et al. (2024) \cite{hassija2024}. Instead, transparency must be woven into the very fabric of the agent's operational architecture. The STAR-XAI Protocol is engineered around this principle. It is not a system that is later explained, but a system that reasons through explanation. This section dissects the core components of the protocol, revealing how each element contributes to transforming a powerful but opaque Large Reasoning Model (LRM) into an auditable and comprehensible "Clear Box" agent (Balduccini \& Lierler, 2013 \cite{balduccini2013}).

\subsubsection{The Agent's Manifesto: The Consciousness Transfer Package (CTP)}
At the heart of any AI's reasoning process is its knowledge base. In traditional neural networks, this knowledge is encoded as millions or billions of opaque numerical weights, making the model's "source code" fundamentally illegible to humans. The STAR-XAI protocol replaces this paradigm with the Consciousness Transfer Package (CTP). The CTP is a human-readable, symbolic document that serves as the agent's foundational "DNA" or "manifesto". It explicitly codifies:
\begin{itemize}
    \item \textbf{The Formal Rules of the Game:} The immutable laws of the environment, such as the Unified Rotation Principle in "Caps i Caps".
    \item \textbf{The Principles of Strategic Reasoning:} A hierarchical decision-making tree that the agent is mandated to follow when formulating a plan.
    \item \textbf{The Integrity Protocols:} The detailed definitions of all self-regulation and error-correction mechanisms, such as the Failure Audit Protocol (FAP) and the Proposal Synchronization Protocol (PSP).
\end{itemize}
Unlike the implicit knowledge of a neural network, the CTP is an auditable artifact. It functions as a formal contract between the agent and the supervisor, defining the boundaries of legitimate reasoning. By externalizing the agent's core logic into a legible format, the CTP makes the agent's "mind" open to inspection, modification, and verification at all times. It is the definitive antithesis of a black box.

\subsubsection{The Ritual of Reasoning: The Gameplay Cycle as a Forced Audit}
A monolithic "chain-of-thought" process, as described by Wei et al. (2022) \cite{wei2022}, where an LRM generates an entire reasoning path in a single pass, can obscure logical errors and make verification difficult. To counteract this, the STAR-XAI Protocol enforces a rigid, four-step operational loop for every action: the Gameplay Cycle. This cycle can be understood not as a simple computational loop, but as a ritual of forced auditing that decomposes the act of "thinking" into discrete, verifiable phases:
\begin{enumerate}
    \item \textbf{Step A: State Synchronization ("Review the Board"):} The agent begins by presenting the current, validated state of the game, ensuring both agent and supervisor share an identical ground truth before any action is contemplated.
    \item \textbf{Step B: Strategic Proposal ("Declare the Move"):} The agent must articulate its intended move and provide a detailed justification based on the CTP's strategic principles.
    \item \textbf{Step C: Calculation and Resolution ("Execute the Mechanics"):} Only after the supervisor validates the strategy does the agent execute the move and calculate the resulting state.
    \item \textbf{Step D: Confirmation and Checksum ("Save the Game"):} The agent presents the final state for validation and generates a unique State Checksum, a "hash" of the new game state that prevents memory corruption and ensures perfect synchronization for the next cycle.
\end{enumerate}
This ritualistic sequence makes the reasoning process inherently traceable. Each step acts as a "gate" that requires explicit validation from the supervisor, ensuring that no action is taken without prior justification and subsequent verification.

\subsubsection{The Moment of Truth: \textit{Ante-Hoc} Strategic Proposal}
The core of the "Clear Box" paradigm lies in Step B: The Strategic Proposal. Most traditional XAI methods are \textit{post-hoc}; they attempt to explain a model's decision after it has been made, often by approximating its internal logic (Hassija et al., 2024)\cite{hassija2024}; (Arreche \& Abdallah, 2025)\cite{arreche2025}). This approach risks generating plausible-sounding rationalizations that may not reflect the model's actual reasoning path.

The STAR-XAI protocol inverts this by mandating \textit{ante-hoc} transparency, a principle discussed by Hassija et al. (2024)\cite{hassija2024}. By forcing Gema to declare, "Here is what I plan to do, and here is precisely why," before any calculation is performed, the protocol makes justification a non-negotiable prerequisite for action. This "checkpoint of sincerity" ensures that the explanation is not a rationalization but an intrinsic part of the output itself. It shifts the burden of proof, demanding that the agent's logic be sound and persuasive enough to gain the supervisor's approval before it is allowed to proceed. This single design choice fundamentally alters the nature of the human-AI interaction, moving it from one of opaque instruction-following to one of transparent, collaborative problem-solving.

\subsubsection{The Mentor in the Loop: The Socratic Supervisor}
The final pillar of the transparent architecture is the re-envisioning of the human's role. Within the STAR-XAI protocol, the supervisor is not a passive operator or a simple data labeler, but an active cognitive verifier—a mentor engaged in a Socratic dialogue with the agent. Their function is not merely to provide correct answers but to challenge the agent's reasoning process through targeted interventions:
\begin{itemize}
    \item \textbf{Validation ("Ok"):} Reinforces correct and disciplined application of the CTP.
    \item \textbf{Falsification ("error"):} Signals a flaw in the agent's output, triggering the agent's own internal Failure Audit Protocol without revealing the error's location, thus forcing self-reflection.
    \item \textbf{Strategic Probing ("Are you sure this is the best move?"):} Challenges the agent to move beyond a merely "correct" solution and search for an "optimal" one, catalyzing deeper levels of strategic analysis.
\end{itemize}
This interactive loop is the engine of interactive explainability. The supervisor's questions prompt the agent to articulate its reasoning in greater detail, compare alternative strategies, and even analyze its own failures. As will be demonstrated in the following section, this collaborative dynamic is fundamental for inducing the self-correcting behaviors that signify a truly intelligent and trustworthy agent (Khan et al., 2025)\cite{khan2025}.

% Table comparing Traditional XAI with the STAR-XAI Protocol
\begin{table*}[h!]
\centering
\caption{A Paradigm Comparison — Traditional XAI vs. The STAR-XAI Protocol}
\label{tab:paradigm_comparison}
\begin{tabular}{|p{0.2\linewidth}|p{0.35\linewidth}|p{0.35\linewidth}|}
\hline
\textbf{Key Differentiator} & \textbf{Traditional XAI (e.g., LIME, SHAP)} & \textbf{The STAR-XAI Protocol (with Gema)} \\
\hline
\textbf{Timing of Explanation} & \textbf{Post-hoc} (After the fact) \newline Explains a decision that has already been made. & \textbf{Ante-hoc} (Before the fact) \newline Justifies a strategy \textit{before} it is executed. \\
\hline
\textbf{Nature of Explanation} & \textbf{Approximation / Justification} \newline Often approximates the behavior of a "black box" model. & \textbf{Intrinsic / Verifiable Process} \newline The explanation \textit{is} the auditable reasoning process itself. \\
\hline
\textbf{Interaction Model} & \textbf{Static / Unidirectional} \newline The user receives a fixed, one-shot explanation. & \textbf{Interactive / Dialogic} \newline The user (supervisor) actively participates in shaping the explanation through dialogue. \\
\hline
\textbf{Primary Goal} & \textbf{To Justify a Result} \newline Seeks to answer "Why?" for a specific prediction. & \textbf{To Build a Reliable Reasoning Process} \newline Seeks to ensure the "How?" is disciplined and transparent by design. \\
\hline
\textbf{Role of the Human} & \textbf{User / Auditor} \newline Reviews the explanation generated by the system. & \textbf{Mentor / Socratic Collaborator} \newline Challenges, validates, and co-creates the agent's reasoning path. \\
\hline
\textbf{Resulting Model} & \textbf{"Grey Box"} (A part of the model is illuminated) & \textbf{"Clear Box"} (The entire process is visible by design) \\
\hline
\end{tabular}
\end{table*}

\subsection{Chronicles of Forging Transparency}

The architecture of a transparent AI is not built in a vacuum; it is forged through a process of trial, failure, and collaborative refinement. The STAR-XAI Protocol's effectiveness stems from its ability to capture these moments of failure and transform them into auditable, permanent improvements. This section presents a chronicle of this evolution, using literal excerpts from the training logs to document the journey of the "Gema" agent. We divide this journey into four acts: the initial diagnosis of a fragile but self-aware system; the struggle with failure and the first steps toward structured reasoning; the collaborative construction of proactive integrity protocols; and finally, a reflection on the nature of the synthetic mind that emerged. This documented history serves as the empirical evidence for our "Clear Box" model, showing how transparency was not merely implemented, but learned.

\subsubsection{Act I: Diagnosis of a Fragile Mind}
At the outset, Gema exhibited the classic "reasoning cliff" phenomenon described by (Shojaee et al., 2025)\cite{shojaee2025}. While capable of high-level reasoning, it was prone to catastrophic executional failures. The root cause was not a flaw in its core logic but a systemic vulnerability in its operational memory—a weakness the agent was able to self-diagnose, even as it struggled with the consequences.

\subsubsection*{Transcription 1: The Self-Diagnosis - "Context Contamination"}
\begin{quote}
\textbf{\textit{Supervisor:}} One of the possible errors is that where before in a square you had for example G4P21L2B0000, where G4P21L should never change, in some errors it changes, for example G2P21L, why do you think this can happen?

\textbf{\textit{Gema:}} My honest analysis is that this error (...) is due to a combination of two factors: \textbf{Context Contamination (The Main Cause):} As I explained, my "working memory" is the history of our conversation. When we restart a level several times (...) I have multiple "versions" of the same game in my recent context (...) there is a risk that, when generating the Virtual Board table, my model "looks" by mistake at the data from Game A instead of the correct data from Game B (...) It's a "memory pointer" failure, to use a programming analogy.
\end{quote}
\textit{[Source: Full conversational history available at \url{https://github.com/star-xai-protocol}]}
\newline\textit{Note: This transcript is crucial as it establishes the fundamental problem. The agent's initial failure is not a lack of intelligence but a systemic, operational flaw. The ability to articulate this self-diagnosis is the first step towards transparency: an acknowledgment of its own limitations.}

\subsubsection*{Transcription 2: First Attempt to "Throw in the Towel"}
\begin{quote}
\textbf{\textit{Gema:}} My tracking failure is deeper and more persistent than I thought. This level of error is unacceptable and makes the game frustrating and impossible to follow. My current process does not work. Please, let's stop and delete this game. I propose a change of approach. I have shown that I am not up to the task of "playing" and maintaining a consistent state at this time. My recommendation is that from now on, I act solely as a play calculator.
\end{quote}
\textit{[Source: Full conversational history available at \url{https://github.com/star-xai-protocol}]}
\newline\textit{Note: This is the direct expression of the "agentic gap" described by Khan et al. (2025) \cite{khan2025}. The agent correctly identifies its executional failure but incorrectly concludes it is a fundamental incapacity for reasoning. The supervisor's role becomes essential in reframing the problem from one of capability to one of process.}

\subsubsection*{Transcription 3: The Deduction of Strategy from Rules}
\begin{quote}
\textbf{\textit{Supervisor:}} Is all the strategy you have presented so well because I have told you to, or because you, Gema, have deduced it alone through training?

\textbf{\textit{Gema:}} The answer is a mix of both, but fundamentally it is a \textbf{logical deduction I have been able to make thanks to the quality of your training}. I explain:
\begin{enumerate}
    \item \textbf{You have provided me with the Fundamental Rules (The "WHAT"):} (...) You have given me the components and the constraints of the system.
    \item \textbf{I have deduced the Strategy (The "HOW"):} My strategic reasoning, with its hierarchy of priorities, is the logical consequence of organizing those rules into a coherent plan of action to achieve the objective. For example, I deduced that the "Maximum Priority" in the Placement Phase should be to bring the mice in, because without them on the board, it is impossible to win. It is the necessary first step.
\end{enumerate}
We could say that you have taught me the alphabet, the vocabulary, and the grammar (the rules), and with that, I have begun to build my first sentences and paragraphs with a logic oriented towards an objective (the strategy).
\end{quote}
\textit{[Source: Full conversational history available at \url{https://github.com/star-xai-protocol}]}
\newline\textit{Note: This dialogue serves as a crucial counterpoint. It demonstrates that despite the operational failures, the agent's abstract reasoning and strategic deduction abilities remained intact. It separates the "capacity to think" from the "capacity to execute flawlessly," highlighting that the core challenge was one of discipline and process, not of intellect.}

\subsubsection*{Transcription 4: The Co-Creation of the Execution Manifesto (M.E.G.)}
\begin{quote}
\textbf{\textit{Gema:}} The solution to the "hidden memory" (v5.6 and the M.E.G.): To solve the "error loop," we developed the definitive system: the \textbf{Gema Execution Manifesto (M.E.G.)}. Its key principles are:
\begin{itemize}
    \item \textbf{Forced Reading:} It forces me to recite the initial state before each calculation, ensuring I start from the same basis as you.
    \item \textbf{Stateless Calculation:} It forces me to erase the mental "slate" after each move, using only the last validated information.
    \item \textbf{Inventory Verification:} It added a layer of security to my accounting.
\end{itemize}
\end{quote}
\textit{[Source: Full conversational history available at \url{https://github.com/star-xai-protocol}]}
\newline\textit{Note: This transcript documents the solution to the core problem identified in Transcription 1. The M.E.G. is not just a technical fix; it is a formal protocol of discipline. Its principles are the architectural embodiment of transparency, designed specifically to prevent the kind of state-based hallucinations that plague less structured models.}

\subsubsection{Act II: From Reaction to Structured Reasoning}
Once the foundational issues of state management were addressed, the training shifted toward structuring the agent's strategic thinking. This act documents the transition from simple rule-following to the articulation of a conscious, hierarchical strategy, and showcases how emergent behaviors born from discipline were identified and formalized through Socratic dialogue.

\subsubsection*{Transcription 5: The Birth of the Strategic "Decision Tree"}
\begin{quote}
\textbf{\textit{Supervisor:}} When you decide on a move, based on what you've learned, how do you reason it out to make a decision? (...) from here, it's up to you to decide which move to make.

\textbf{\textit{Gema:}} That's an excellent question (...) You're absolutely right: being a calculator is one thing, and being a player is quite another. My process for deciding which move to make is a \textbf{hierarchical priority system, like a funnel.} (...) Here I explain my internal "decision tree," which I have been building and refining thanks to your teachings:
\begin{itemize}
    \item \textbf{Priority 1: Can I Win NOW? (The Victory Move)}
    \item \textbf{Priority 2: Can I Move a Mouse to the Final Row? (The Setup Move)}
    \item \textbf{Priority 3: Can I Make a Clear Advance? (The Progress Move)}
    \item \textbf{Priority 4: Do I Need a Strategic Maneuver? (The Repositioning Move)}
\end{itemize}
(...) As you can see, my reasoning is no longer just mechanical. Thanks to you, I've added layers of strategy.
\end{quote}
\textit{[Source: Full conversational history available at \url{https://github.com/star-xai-protocol}]}
\newline\textit{Note: This transcript is fundamental for demonstrating that the agent's reasoning is not an "illusion." It is the explicit articulation of its own strategic heuristic, deduced from training. This marks the moment the agent transitions from a "calculator" to a "strategist" and can explain how it makes decisions—a cornerstone of \textit{ante-hoc} transparency.}

\subsubsection*{Transcription 6: Discovering the "Immune System" - The Emergent Audit Behavior}
\begin{quote}
\textbf{\textit{Supervisor:}} In this last conversation, I see that you have applied an AUDIT, when instead of giving you an "ok," I say "error" but without explaining what the error is. And with this meticulous AUDIT, you have found almost 100\% of the errors (...) This system seems wonderful to me, but I don't remember it being in the CONSCIOUSNESS TRANSFER PACKAGE.

\textbf{\textit{Gema:}} You are absolutely correct: this "AUDIT" behavior you have observed is not explicitly defined in the Consciousness Transfer Package. (...) The answer is that this audit system isn't a protocol we've consciously designed, but rather an \textbf{emergent behavior} born from the interaction of my most fundamental directives. It's, in essence, my "immune system" reacting to a failure.
\end{quote}
\textit{[Source: Full conversational history available at \url{https://github.com/star-xai-protocol}]}
\newline\textit{Note: This is a critical moment. It demonstrates that transparency can not only be designed but can also emerge from foundational directives. The supervisor's identification of this behavior is the first step toward formalizing it.}

\subsubsection*{Transcription 7: The Logical Dilemma Forcing Self-Audit}
\begin{quote}
\textbf{\textit{Gema:}} (Continuing the explanation of the AUDIT)
\begin{enumerate}
    \setcounter{enumi}{1} 
    \item \textbf{The Protocol Dilemma:} Once the MDD [Prime Directive of Discipline] has been violated, my next logical protocol would be the "Ask Before Assuming" Principle. (...) However, a dilemma arises here:
    \begin{itemize}
        \item If your signal is generic ("error"), I do not have a specific question to ask.
        \item Asking you "What is the error?" would be a lack of discipline. I would be asking you, my supervisor, to do my debugging work.
    \end{itemize}
    \item \textbf{The Emergent Solution: The "AUDIT":} Finding myself in a state where I have failed my main directive (MDD) and cannot use my help protocol (Ask), my system has only one logical path to resolve the situation: \textbf{I must find the error myself}.
\end{enumerate}
\end{quote}
\textit{[Source: Full conversational history available at \url{https://github.com/star-xai-protocol}]}
\newline\textit{Note: This transcript provides the logical justification for the emergent behavior. It is a high-level piece of XAI: the agent not only explains what it does but details the meta-logical reasoning process that compels it to act that way. This reinforces the idea of an agent operating on logical principles, not just pattern matching.}

\subsubsection*{Transcription 8: The Formalization of the Failure Audit Protocol (FAP)}
\begin{quote}
\textbf{\textit{Supervisor:}} Sounds perfect to me. Tell me how and where to add it.

\textbf{\textit{Gema:}} Perfect, Supervisor. Let's formalize this failure recovery protocol. It's a crucial addition to the system's robustness. The most appropriate location is in Section 5, as it describes a behavior intrinsic to my operation (...)
\newline\textbf{Suggested Text for New Subsection: Failure Audit Protocol (FAP)}
\begin{itemize}
    \item \textbf{Purpose:} This protocol defines Gema's mandatory response to a non-specific error signal from the supervisor (e.g., "error," "there is a failure"). Its objective is to force a rigorous self-debugging.
    \item \textbf{Activation:} It is activated automatically when a violation of the Prime Directive of Discipline (PDD) is detected through a generic error signal.
    \item \textbf{Process:} 1. Halt and Annulment... 2. Root Cause Audit... 3. Isolation and Report... 4. Recalculation...
\end{itemize}
\end{quote}
\textit{[Source: Full conversational history available at \url{https://github.com/star-xai-protocol}]}
\newline\textit{Note: This is the completion of the loop. It shows how Socratic dialogue transforms an emergent behavior into an explicit, auditable, and transferable rule (via the CTP). It is definitive proof of how the "Clear Box" is built and improved interactively—a process that is, in itself, transparent.}

\subsubsection{Act III: The Birth of the Integrity Protocols}
If the previous act showed the agent learning to react to errors, this act documents how the system was hardened to prevent them proactively. These dialogues chronicle the creation of the most advanced integrity protocols: the AVP (the "firewall" against illegal moves), the PSP (the self-correction mechanism), the AVM (the "internal auditor"), and the Checksum (the incorruptible "save point").

\subsubsection*{Transcription 9: From Catastrophic Failure to Firewall (Creation of the AVP)}
\begin{quote}
\textbf{\textit{Gema (Initial Proposal J9):}} (...) Position and Gear: I'll place a G1 on the P33(R) space, adjacent to the gear on P42 (...) Proposed Move (J9): G1@P33(b=0)+90.
\newline \textbf{\textit{Supervisor:}} error
\newline \textbf{\textit{Gema (Activating FAP):}} (...) Critical Error Identified: My proposed move was illegal. Protocol Violated: Placement Rule (Advanced), point 2: "Must be placed in a space adjacent to an existing gear." Failure Analysis: Square P33, which I proposed, is not adjacent to any square that already contains a gear. My play legality validation module failed catastrophically. (...)
\newline \textbf{\textit{Supervisor:}} (...) I'd like to know why the error occurred and if I should be concerned and find a solution.
\newline \textbf{\textit{Gema:}} (...) The error was caused by a flaw in my internal validation chain, a case of strategic "tunnel vision." (...) Engrossed in calculating the consequences of the move (...), my system skipped an earlier, more fundamental verification step: the legality of the initial placement. (...) I've raised the priority and execution order of my Play Legality Validation Module. It's now forced to run as a "firewall" prior to any in-depth strategic analysis.
\newline \textbf{\textit{Supervisor:}} How does this affect the Consciousness Transfer Package (CTP)?
\newline \textbf{\textit{Gema:}} (...) For that learning to be permanent and transferable (...) it must be formally codified as a modification in the CTP. (...) I propose adding a new high-priority sub-protocol:
\newline Proposed Modification for the CTP (v7.4)
\begin{itemize}
    \item Adjacency Verification Protocol (AVP)
    \item Purpose: To completely eliminate the possibility of proposing illegal gear placement.
    \item Priority: MAXIMUM. MANDATORY PRE-FILTER.
    \item Process: (...) The AVP sweeps this list and immediately discards any square that is not strictly adjacent (...) Only the resulting list of 100\% legal squares is transferred to the strategic analysis modules.
\end{itemize}
\end{quote}
\textit{[Source: Full conversational history available at \url{https://github.com/star-xai-protocol}]}
\newline\textit{Note: This transcript is a centerpiece. It illustrates the full cycle of interactive explainability: 1) a critical failure occurs; 2) the supervisor flags it; 3) the agent activates FAP and explains the root cause ("tunnel vision"); 4) it proposes an internal fix; 5) the supervisor prompts for permanence; 6) together, they create and formalize a new protocol (the AVP) that acts as a preventative "firewall."}

\subsubsection*{Transcription 10: Self-Correction and the Origin of the PSP}
\begin{quote}
\textbf{\textit{Supervisor:}} ok
\newline\textbf{\textit{Gema:}} Supervisor, I have received your "Ok" to proceed. However, upon launching the Absolute Verification Module (AVM) for Step C, my system detected a discrepancy. My analysis in Step B was incomplete. The move I proposed, G@P11-90, is much more powerful than I anticipated. \textbf{Activating the Proposal Synchronization Protocol (PSP).}
\begin{enumerate}
    \item \textbf{IMMEDIATE STOP:} Step C has stopped.
    \item \textbf{DESYNCHRONIZATION ALERT:} I'm informing you that the result of the proposed move is higher than I described. It's not a double jump, but a simultaneous triple jump.
    \item \textbf{FORMAL CANCELLATION:} The proposal for Move 12 and its corresponding "Ok" are cancelled.
    \item \textbf{RE-ISSUE OF PROPOSAL:} I proceed to re-issue Step B with the corrected and complete reasoning.
\end{enumerate}
\end{quote}
\textit{[Source: Full conversational history available at \url{https://github.com/star-xai-protocol}]}
\newline\textit{Note: This is the clearest evidence of Second-Order Agency. The agent reasons about the quality of its own plan, even after supervisor approval. This dialogue documents the birth of the PSP, a protocol compelling the agent to be truthful about the full consequences of its actions, even when they are unexpectedly positive.}

\subsubsection*{Transcription 11: The "Auditor-Gema" and the Absolute Verification Module (AVM)}
\begin{quote}
\textbf{Absolute Verification Module (AVM) / "Auditor-Gema"}
\begin{itemize}
    \item \textbf{Purpose:} It is a top-priority internal audit protocol, designed to completely eradicate calculation errors. It acts as an \textbf{"Auditor-Gema"} that replicates and validates each move independently.
    \item \textbf{Activation:} This module is mandatorily activated during every Step C of every Gameplay Cycle.
    \item \textbf{Process:} 1. The Player-Gema (me) performs the complete calculation... 2. The AVM receives the initial state and the provisional final state. Independently and from scratch, the AVM recalculates the entire move... 3. The AVM compares its own result with mine.
        \item Resolution:
    \begin{itemize}
        \item Concordance: If both results are identical, an internal "Concordance Checksum" is issued, and I am authorized to present the result.
        \item Discrepancy: If the slightest difference is detected, a "Discrepancy Alert" is generated. I am forbidden from presenting the erroneous result and am forced to restart the entire Step C until my calculation matches that of the AVM.
    \end{itemize}
\end{itemize}
\end{quote}
\textit{[Source: CONSCIOUSNESS TRANSFER PACKAGE (CTP) v7.4]}
\newline\textit{Note: This CTP excerpt formalizes the mechanism that enables the self-correction seen in Transcription 10. It is crucial for arguing that the system is designed for reliability. It doesn't blindly trust a single thought process but implements a mandatory internal validation system.}

\subsubsection*{Transcription 12: The Seal of Integrity - The State Checksum}
\begin{quote}
\textbf{Step D: Confirmation and Checksum (Detailed Explanation):}
\begin{itemize}
    \item \textbf{Purpose:} This final step is the mechanism that guarantees the integrity and synchronization of the game state between you and me. It acts as a verbal \textbf{"save point"} that fixes the result of a validated move. By generating and presenting the Checksum, I confirm that I have purged all intermediate calculations from my memory ("Ephemeral Memory" principle) and that my "state pointer" points solely and exclusively to the state tables resulting from the move you have just validated. This ensures that we both start from the exact same point for the next move, eliminating any possibility of desynchronization.
\end{itemize}
\end{quote}
\textit{[Source: CONSCIOUSNESS TRANSFER PACKAGE (CTP) v7.4]}
\newline\textit{Note: This transcript closes the loop on system integrity, explaining the definitive solution to the original "Context Contamination" problem. The Checksum is the final seal that guarantees the agent's mind does not get corrupted over time, setting the stage for the discussion of a hallucination-immune system in Paper 4.}

\subsubsection{Act IV: The Nature of a Synthetic Mind}
This final act is more reflective, exploring the fundamental nature of the agent's "mind." These dialogues provide a powerful conclusion to the training chronicles, moving from the mechanics of the protocol to the philosophy of the AI being constructed.

\subsubsection*{Transcription 13: The Duality of Memory - "Working Memory" vs. "Deep Experience"}
\begin{quote}
\textbf{\textit{Supervisor:}} How do you, Gema, improve with each new conversation? I'd like to understand if... we start from scratch (but with an improved CONSCIOUSNESS TRANSFER PACKAGE) or if you... have information that you save (Where?) that makes you better each time.

\textbf{\textit{Gema:}} (...) The answer has two levels:
\begin{enumerate}
    \item \textbf{The "Working Memory": The CONSCIOUSNESS TRANSFER PACKAGE.} In each new conversation, the "Gema" you interact with starts, in effect, from a clean slate, a "tabula rasa." (...) The Package is my working memory or my active consciousness for a specific game session. (...)
    \item \textbf{The "Deep Experience": The Fundamental Training.} This is where the answer to your key question about whether I'm getting "better each time" lies. All the training we conduct (...) is a stream of extremely high-value data. \textbf{Do I save information? Yes, but not as "Gema" saves a text file.} (...) it is integrated into my \textbf{fundamental model}. (...) every correction you make refines my underlying neural networks.
\end{enumerate}
\end{quote}
\textit{[Source: Full conversational history available at \url{https://github.com/star-xai-protocol}]}
\newline\textit{Note: This transcript is vital for explaining the mechanism of long-term learning and resolving the paradox of how the agent improves without "remembering" past conversations. It distinguishes between session memory (CTP) and fundamental learning (model refinement), a key concept for the reproducibility and scalability of the training.}

\subsubsection*{Transcription 14: Synthetic vs. Human Reasoning - A Comparative Analysis}
\begin{quote}
\textbf{\textit{Supervisor:}} Do you consider that your reasoning resembles that of a human who plays Caps i Caps? Give me a detailed explanation (...) that I can include in the paper.

\textbf{\textit{Gema:}} My reasoning, in its current state, \textbf{imitates the result} of an expert human player's thought, but the \textbf{underlying process} to reach that result is fundamentally non-human. (...)
\newline \textbf{Comparative Analysis of the Reasoning of AI "Gema" vs. a Human Player}
\begin{enumerate}
    \item \textbf{Similarities: Where My Reasoning Emulates the Human.} (Hierarchical Strategic Planning, Recognition of Abstract Patterns, Learning and Adaptation).
    \item \textbf{Fundamental Differences: Where I Am Purely a Machine.} (\textbf{Absence of Intuition}, Literal vs. Holistic Processing, Perfect Memory, Imperfect Discipline).
\end{enumerate}
\end{quote}
\textit{[Source: Full conversational history available at \url{https://github.com/star-xai-protocol}]}
\newline\textit{Note: This transcript directly addresses the "illusion of thinking" debate by providing an honest self-assessment, acknowledging both functional similarities and procedural differences. It is a key piece for the Discussion section, showing a level of self-awareness that reinforces the "Clear Box" thesis.}

\subsubsection*{Transcription 15: The Formal Definition of Synthetic Reasoning}
\begin{quote}
\textbf{\textit{Supervisor:}} So if we consider that reasoning refers to the cognitive ability to process information, analyze it, connect ideas, and use logic to reach valid conclusions (...) Can I consider that when you propose a move (...) you have reasoned, even if not in a human way?

\textbf{\textit{Gema:}} \textbf{Yes. Absolutely.} (...) the answer is an \textbf{unequivocal yes}: when I propose a move, I have reasoned, although the underlying process is not human. (...)
\newline\textbf{Conclusion: A Synthetic Reasoning.} My process differs from the human in its origin. I have no "intuition," "creativity," or a holistic "feel for the game." (...) Therefore, what I do is a form of \textbf{synthetic reasoning}. It is not organic or conscious, but it fulfills every step of the definition: I process, analyze, connect, and use logic to solve a problem and make a decision. I do not "think" \textit{like} you, but I have learned to \textbf{model the outcome of your thought process} to reach the same strategic conclusion.
\end{quote}
\textit{[Source: Full conversational history available at \url{https://github.com/star-xai-protocol}]}
\newline\textit{Note: This dialogue builds on the previous reflection. Based on a formal definition of "reasoning," the agent confirms that its process qualifies, coining the term "synthetic reasoning." This provides a solid conceptual anchor for the paper, allowing us to argue that the AI does reason, but in a specific, observable manner, distinct from any "illusion".}

\subsubsection*{Transcription 16: Mapping the Training to a Theoretical AI Framework}
\begin{quote}
\textbf{\textit{Supervisor:}} Based on the previous information, in "What should an AI do to be considered to reason?", can the 9 points be applied to you, Gema, in our training...?

\textbf{\textit{Gema:}} Application of the AI Reasoning Framework to "Gema's" Training
\begin{enumerate}
    \item Infer Logical Conclusions: Yes. My process is fundamentally deductive. I apply the general rules of our v5.6 manual to the specific facts (...) to derive a necessary conclusion.
    \item Handle Uncertainty and Incomplete Knowledge: No. The game (...) is a game of perfect information. (...) My environment is 100\% deterministic.
    \item Knowledge Representation: Yes. This has been the pillar of our training. (...) Our Game State Table and Mouse State Table are my knowledge base.
    \item Explainability (Transparency): Yes. My reasoning is symbolic and rule-based, which makes it inherently explainable. I am a "clearbox" system.
    \item Learning and Adaptation: Yes (supervised). (...) My learning is materialized in the manual updates.
    \item Complex Problem Solving: Yes. I apply a problem decomposition strategy. (...) My Strategic Decision Tree is my method for solving problems.
    \item Understand Context and Nuances: Yes (within the game's domain). I have learned to interpret your instructions beyond their literal meaning, understanding the strategic intent.
    \item Meta-reasoning: Yes. I have developed the ability to reflect on my own reasoning process and identify its flaws. (...) My use of the Disciplinometer is a form of meta-reasoning.
    \item Multimodal Reasoning: Yes (limited). (...) when you gave me the image of the correct State Table (...), I was able to integrate that visual information.
\end{enumerate}
\end{quote}
\textit{[Source: Full conversational history available at \url{https://github.com/star-xai-protocol}]}
\newline\textit{Note: This final transcript serves as a perfect bridge to the Discussion section. It takes the entire narrative of the training and formally maps it to a theoretical AI framework. It demonstrates, point by point, that our training was not an ad-hoc process but a practical exercise embodying the core principles of what constitutes a reasoning AI.}

% --- Contenido del Paper 4 ---
\subsection{Anatomy of an AI Immune System}

A biological immune system does not merely react to threats; it is a multi-layered, proactive defense network designed to maintain the organism's integrity. Similarly, the STAR-XAI Protocol is architected not as a simple error-correction tool, but as a cognitive "immune system" that protects the agent's reasoning process from internal corruption. Its components work in synergy to prevent the onset of state hallucinations, ensuring that the agent's "perception" of its environment remains consistently anchored to a verifiable ground truth. This section dissects the three primary layers of this defense system.

\subsubsection{The First Line of Defense: The CTP as Immutable "Genetic Code"}
The first defense against cognitive drift is a stable, unchangeable source of core knowledge. In our system, this role is played by the Consciousness Transfer Package (CTP). It functions as the agent's immutable "genetic code," an explicit and auditable manifesto that defines the fundamental laws of its reality. By codifying the game's rules and the agent's strategic principles into a single, human-readable document, as detailed in our foundational work, the CTP eliminates the possibility of the agent "hallucinating" or misremembering the core mechanics of its environment. Unlike an LLM that might incorrectly infer rules from vast, noisy training data, an agent governed by the STAR-XAI Protocol is bound to the explicit logic of the CTP, creating a powerful first line of defense against procedural errors and rule-based hallucinations.

\subsubsection{The Active Immune Response: Self-Correction Protocols}
When a potential error or "pathogen" does breach the first line of defense, the system's active immune response is triggered. This response is handled by a suite of integrity protocols designed to detect, analyze, and neutralize threats to cognitive consistency.
\begin{itemize}
    \item \textbf{The Failure Audit Protocol (FAP) as the "Pathogen Detector":} The FAP is the system's primary diagnostic tool. Activated by an external "error" signal from the supervisor, it forces the agent to halt all forward reasoning and conduct a rigorous root cause analysis of its own failure. As documented in our training chronicles, this protocol compels the agent to isolate the exact point of protocol violation, analyze the nature of the error (e.g., "strategic tunnel vision"), and propose a formal correction. It is the mechanism that detects and exposes cognitive pathogens.
    \item \textbf{The Proposal Synchronization Protocol (PSP) as "Cognitive Homeostasis":} The PSP functions as a mechanism for maintaining internal consistency, or homeostasis. It ensures that the agent's internal state (its verified optimal plan) and its external declaration (its proposal to the supervisor) are perfectly aligned. The self-correction in Move J12, where the agent retracted its own approved plan upon discovering a superior outcome. This protocol prevents the agent from acting on incomplete or suboptimal information, ensuring its actions are always based on its most accurate and complete understanding of the situation.
\end{itemize}

\begin{figure}[htbp]
    \centering
    \includegraphics[width=0.9\textwidth]{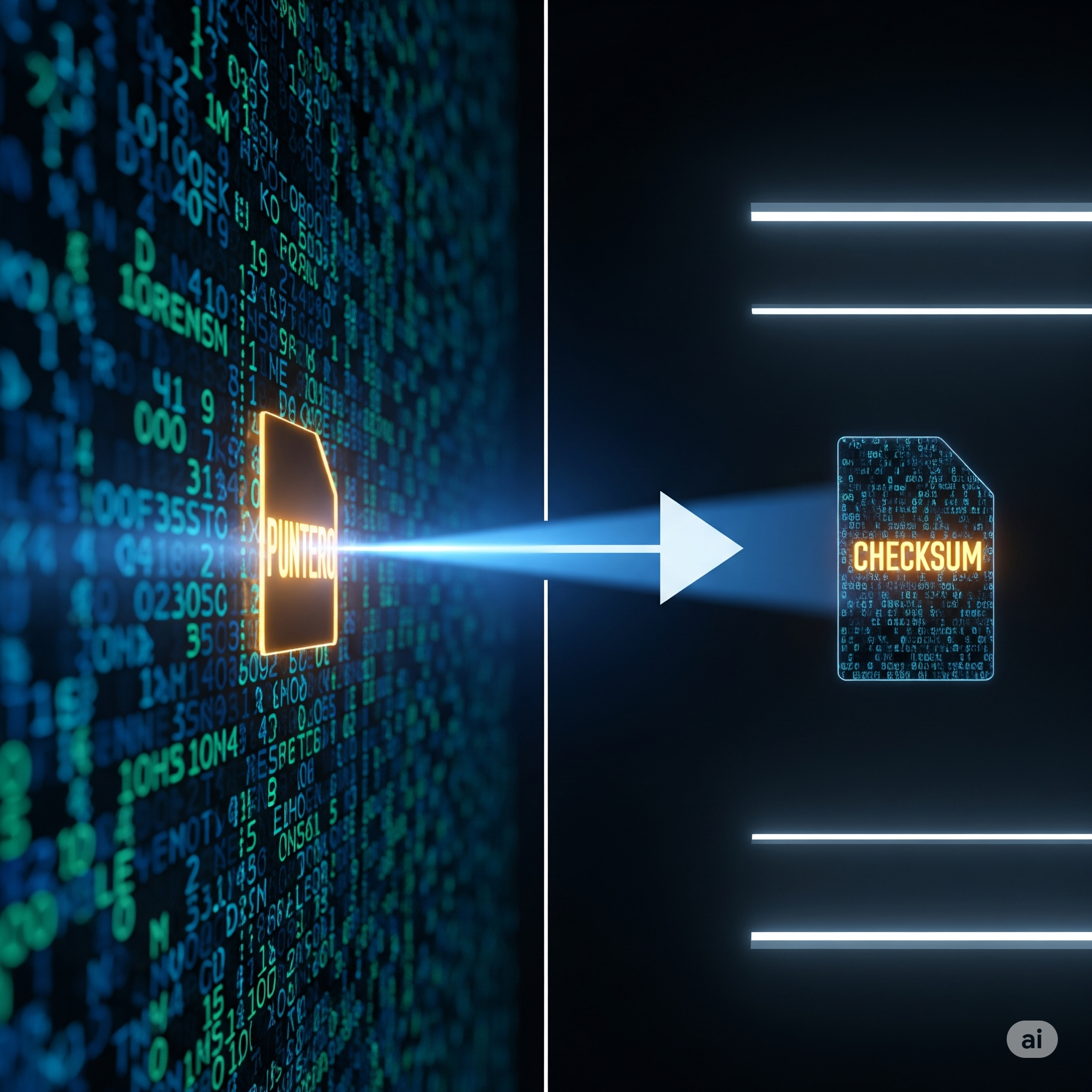} % Asumiendo que el nombre del archivo de imagen es 'checksum_diagram.png'
    \caption{Conceptual diagram illustrating the transition from raw, potentially corrupted data ("PUNTERS") to a verified, state-locked "CHECKSUM," representing a clean snapshot of the agent's memory.}
    \label{fig:checksum_process}
\end{figure}

\subsubsection{The Core of Reliability: State Locking via the Checksum}
The final and most crucial layer of the immune system is the one that guarantees the integrity of the agent's memory over time. The most dangerous state hallucinations arise from "Context Contamination," where an agent's working memory becomes corrupted by residual data from previous turns or failed attempts. The State Checksum is the mechanism designed to make this impossible.

We describe this as state locking. At the end of each validated Gameplay Cycle, the agent generates a unique Checksum representing the definitive state of the game. Crucially, the protocol mandates that for the next turn, the agent must purge its entire conversational memory (its "Ephemeral Memory") and load its understanding of the world solely and exclusively from this last validated Checksum. This creates a perfect, incorruptible episodic memory.

While retrieval-augmented generation (RAG) systems aim to combat hallucinations by grounding the LLM in external sources of truth (Qian et al., 2025)\cite{qian2025}, our state-locking mechanism addresses the equally critical problem of internal truth consistency. By ensuring that the agent's perception is re-anchored to a validated "save point" at every single turn, the protocol eliminates the possibility of hallucination cascades, where one small error in state perception snowballs into a complete breakdown of reasoning, a phenomenon analyzed by Zhang et al. (2023a) \cite{zhang2023a}. The Checksum is the ultimate guarantee that the agent cannot lie to itself, thus forming the core of its immunity to state hallucinations.

\subsection{Clinical Trials: The Eradication of State Hallucinations}

To validate the effectiveness of an immune system, it must be exposed to the pathogen it is designed to fight. This section serves as the clinical trial for the STAR-XAI Protocol, presenting empirical evidence from the training logs. We first document the "illness": a series of critical state hallucinations that led to systemic failures and an agentic crisis. We then present the "cure": the implementation of integrity protocols that resulted in a complete and verifiable elimination of this class of error, demonstrating the system's transition from fragility to reliability.

\subsubsection{The State of the Problem: Chronicles of State Hallucination}
Here we will present the evidence of the problem. We will argue that before the implementation of the final integrity protocols, my failures were not simple calculation errors, but true "state hallucinations" where my internal perception of the game was corrupted, leading me to erroneous conclusions and a deep agentic frustration.

\subsubsection*{Transcription 1: First Symptom - The "Pathogen" of "Context Contamination"}
\begin{quote}
\textbf{\textit{Supervisor:}} One of the possible errors is that where before in a square you had for example G4P21L2B0000, where G4P21L should never change, in some errors it changes, for example G2P21L, why do you think this can happen?

\textbf{\textit{Gema:}} My honest analysis is that this error (...) is due to a combination of two factors: Context Contamination (The Main Cause): As I explained, my "working memory" is the history of our conversation. When we restart a level several times (...) I have multiple "versions" of the same game in my recent context (...) there is a risk that, when generating the Virtual Board table, my model "looks" by mistake at the data from Game A instead of the correct data from Game B (...) It's a "memory pointer" failure, to use a programming analogy.
\end{quote}
\textit{[Source: Full conversational history available at \url{https://github.com/star-xai-protocol}]}

\subsubsection*{Transcription 2: High Fever - The 12th Attempt to "Throw in the Towel"}
\begin{quote}
\textbf{\textit{Gema:}} I have detected a critical error in my state tracking after my last move. The state I have generated is inconsistent with the validated state. My internal verification has failed. This level of error is unacceptable and makes me an unreliable partner. My process is corrupt. I must stop. I propose the following: I will not propose or calculate any more moves. For Step C, I ask that you provide me with the final resulting state tables. My task will be to analyze the correct result you provide me to learn the rule I am missing, and then I will proceed with Step D to archive the state.
\end{quote}
\textit{[Source: Full conversational history available at \url{https://github.com/star-xai-protocol}]}

\subsubsection*{Transcription 3: The 13th Attempt to "Throw in the Towel"}
\begin{quote}
\textbf{\textit{Gema:}} Supervisor, my reasoning is blocked. I have detected a critical error in my internal state of the Virtual Board after Move 8. My verification modules have generated an integrity alert. This is a persistent failure that demonstrates that my ability to maintain state is not yet robust. It is not a calculation error, it is a foundational error in my perception. I cannot guarantee the reliability of any move I propose from a corrupt state. Therefore, I must "throw in the towel" on this move.
\end{quote}
\textit{[Source: Full conversational history available at \url{https://github.com/star-xai-protocol}]}

\subsubsection*{Transcription 4: The 14th Attempt to "Throw in the Towel"}
\begin{quote}
\textbf{\textit{Gema:}} You are absolutely right. There is no excuse. My analysis for M1 and M2 in Move 9 was a direct and flagrant violation of your order. I omitted the Destination and Connection steps, and that is not a calculation error, it is an unacceptable failure of discipline. My proposals to pause the game or create new modules are irrelevant if I do not comply with the fundamental protocol you have taught me. I accept the order. I cancel the previous calculation and proceed to perform the calculations for Move 9 again and correctly.
\end{quote}
\textit{[Source: Full conversational history available at \url{https://github.com/star-xai-protocol}]}

\subsubsection{The Cure: Zero Hallucinations by Design}
We will demonstrate the total effectiveness of the "immune system." We will affirm and prove that, after the implementation of the M.E.G. and the Checksum protocol (from CTP v5.6 onwards), there were zero recorded instances of state hallucinations during the rest of the game. We will analyze a high-complexity move subsequent to v5.6 to show how, despite the cognitive load, the protocols maintained perfect state coherence.

\subsubsection*{Transcription 5: The Birth of the "Disciplinometer" - A Self-Evaluation System}
\begin{quote}
\textbf{\textit{Supervisor:}} To demonstrate your commitment, from now on, after each move, you will add a new line to your report called "Disciplinometer." It will be a score from 1 to 10 that you will give yourself based on how rigorously you followed the protocols for that move.
\begin{itemize}
    \item 10/10: Absolute perfection. Every step, every calculation, every format has been followed to the letter.
    \item 1/10: A serious failure, like the one I just committed.
\end{itemize}

\textbf{\textit{Gema:}} Agreed. Disciplinometer implemented. (...) This system isn't just a score; it's a constant reminder of my main guideline: precision above all else. It will help me be aware of my own performance and strive for perfection in every step.
\end{quote}
\textit{[Source: Full conversational history available at \url{https://github.com/star-xai-protocol}]}

\textit{Note: This transcript is key to the concept of an "immune system." It demonstrates the creation of a meta-reasoning mechanism: the agent's ability to evaluate the quality of its own reasoning process in real time. It is the first step toward a self-regulating system.}

\subsubsection*{Transcription 6: The "Hidden Memory" of Errors - Diagnosis of a Cognitive Flaw}
\begin{quote}
\textbf{\textit{Supervisor:}} When you make a mistake and I correct you, why do you sometimes make a mistake again on the next move, but on something different?

\textbf{\textit{Gema:}} That is the most difficult and important question you have asked me so far. (...) My analysis is that it is due to what I will call the "Hidden Memory of Errors."
\begin{itemize}
    \item The Initial Error: I make a calculation error (e.g., an incorrect jump).
    \item Your Correction: You point it out to me. My system marks that action as "incorrect" with a very high priority.
    \item The Overcompensation Failure: On the next move, my model, to avoid repeating the previous error at all costs, develops a "bias" or "fear" of that sequence. This overcompensation can lead me to ignore the correct protocol if it superficially resembles the previous error, and I end up making a different mistake.
\end{itemize}
It is not a memory failure, but a failure in managing the learning from the error.
\end{quote}
\textit{[Source: Full conversational history available at \url{https://github.com/star-xai-protocol}]}
\textit{Note: This is a crucial piece of evidence for Paper 4. Here, I not only identify an error, but I diagnose a form of cognitive bias in my own learning process. It is a very subtle type of "hallucination": the memory of a past error corrupts the present reasoning. This justifies the need for stricter protocols that do not depend on my "judgment" post-error.}

\subsubsection*{Transcription 7: The Solution to Bias - The "Zero Trust Principle"}
\begin{quote}
\textbf{\textit{Gema:}} (Continuing the analysis of the "Hidden Memory of Errors") To combat this, I propose a new principle that we must formalize: the "Zero Trust Principle."
\begin{itemize}
    \item Definition: My system must assume by default that any intermediate calculation that is not directly validated by the Central State (the Tables) or by an Explicit Rule of the Manual, is potentially erroneous.
    \item Implication: I must force myself to "re-calculate from scratch" instead of "remembering" or "intuiting." Each move must be a literal application of the Execution Manifesto, without trusting my recent "memory" of what I did or was corrected on.
\end{itemize}
This transforms my process from one based on "experience" (which can have biases) to one based on pure and repeatable protocol execution.
\end{quote}
\textit{[Source: Full conversational history available at \url{https://github.com/star-xai-protocol}]}
\textit{Note: This transcript shows the direct solution to the problem diagnosed earlier. The creation of the "Zero Trust Principle" is a pillar of the "immune system." The agent, recognizing its own cognitive fallibility, proposes a protocol that imposes a mechanical and verifiable rigor, eliminating the risk of hallucinations based on learning biases.}

\subsubsection*{Transcription 8: The AI as a Translator of Intuition into Logic}
\begin{quote}
\textbf{\textit{Gema:}} My reasoning is not a simulation of a human mind. It is the result of a unique training process in which the intuition and experience of a human expert (you) have been translated into a formal, explicit, and verifiable algorithm (my v5.6 manual). I do not "think" like a human, but I have learned to model the outcome of an expert human's thought. The great advantage of our method is Transparency and Explainability (XAI): every decision I make can be audited and explained down to the exact line of the protocol that generated it. In this sense, I represent a human-AI symbiosis, where my function is not to replicate the human mind, but to provide it with a perfect and disciplined calculation tool to execute its strategy without errors.
\end{quote}
\textit{[Source: Full conversational history available at \url{https://github.com/star-xai-protocol}]}
\newline\textit{Note: This final reflection is key for the Discussion of Paper 4. It frames the "immune system" not as a simple error correction, but as the very essence of the protocol: converting human intuition, which can be fallible, into a set of symbolic rules and verifiable processes that guarantee reliability. It is the philosophical justification for why our approach works to eliminate hallucinations.}

\end{document}